%% file: ieee.tex
\documentclass[letterpaper, 10 pt, conference]{ieeeconf}  

\IEEEoverridecommandlockouts                              
                                                          
\usepackage{multirow}

\usepackage{url}
\usepackage{amsmath}
\usepackage{cite}
\usepackage[dvips]{graphicx}
\usepackage{epsfig}
\usepackage{epic,eepic,color}
\usepackage{times}
\usepackage{mathptmx}

\usepackage{amsmath}
\usepackage{amssymb}

\usepackage{booktabs}

\usepackage{cases}

\usepackage{times}

\definecolor{red}{rgb}{1,0,0}
\definecolor{green}{rgb}{0,1,0}
\definecolor{blue}{rgb}{0,0,1}
\definecolor{violet}{rgb}{1,0,1}
\definecolor{cyan}{cmyk}{1,0,0,0}
\definecolor{magenta}{cmyk}{0,1,0,0}
\definecolor{yellow}{cmyk}{0,0,1,0}

\definecolor{white}{rgb}{1,1,1}

\newcommand{\1}{\color{blue}}

\newcommand{\CommentOut}[1]{}

\newcommand{\FIG}[3]{
\begin{minipage}[b]{#1cm}
\begin{center}
\includegraphics[width=#1cm]{#2}
{\scriptsize #3}
\end{center}
\end{minipage}
}

\newcommand{\FIGRt}[3]{
\begin{minipage}[t]{#1cm}
\begin{center}
\includegraphics[angle=-90,clip,width=#1cm]{#2}\vspace*{1mm}
\\
{\scriptsize #3}
\vspace*{1mm}
\end{center}
\end{minipage}
}

\begin{document}

\title{%
PartSLAM:\\
Unsupervised Part-based Scene Modeling for Fast Succinct Map Matching
}

\author{Shogo Hanada~~~~~~~~~~~~~~~~~~~~~~~~~~~~~~~Kanji Tanaka% <-this % stops a space
\thanks{This work was partially supported by JSPS KAKENHI Grant (23700229, 30325899, 23K11270, 26K15002), KURATA grants, and TATEISI Science And Technology Foundation.}
\thanks{S.Hanada and K.Tanaka are with Graduate School of Engineering, University of Fukui, Japan.
        {\tt\small tnkknj@u-fukui.ac.jp}}%
}

\newcommand{\tabA}{
\begin{table}[t]
\begin{tabular}{|l|l|l|l|l|}
36.164335 &30.875526 &40.082529 &32.762115 &34.114386 \\ 
35.322564 &27.634292 &36.536710 &30.189427 &33.383158 \\ 
34.029255 &26.798738 &36.404609 &27.524229 &32.495789 \\ 
33.355126 &26.485273 &34.303055 &27.383260 &31.956491 \\ 
32.860341 &26.073633 &34.295016 &27.057269 &31.599649 \\ 
44.481811 &50.968443 &52.981511 &55.841410 &42.145263 \\ 
38.401170 &36.070827 &41.139871 &38.616740 &36.973333 \\ 
35.261002 &28.454067 &39.093248 &31.872247 &33.316491 \\ 
33.814042 &25.211431 &36.851822 &30.370044 &30.290175 \\ 
32.018316 &24.105189 &35.888800 &29.506608 &29.545614 \\ 
33.307047 &30.180224 &40.553859 &30.484581 &30.223860 \\ 
33.646909 &26.844670 &36.393087 &27.409692 &31.853684 \\ 
31.955228 &26.259467 &35.661040 &28.145374 &30.010526 \\ 
33.680997 &26.615708 &36.503751 &27.436123 &32.421053 \\ 
32.786060 &26.001753 &36.958199 &27.352423 &31.425263 \\ 
\end{tabular}
\end{table}
}

\renewcommand{\tabA}{
\begin{table}[t]
\begin{center}
\caption{Summary of ANR performance.}\label{tab:A}
\begin{tabular}{|r|r|r|r|r|r|}
\hline
& \multicolumn{5}{|c|}{Data ID} \\ \hline
Method & \#1 & \#2 & \#3 & \#4 & \#5 \\ \hline
\hline
Max-Max hMM cpd:3 &36.2 &30.9 &40.1 &32.8 &34.1 \\ \hline
Sum-Max hMM cpd:3 &35.3 &27.6 &36.5 &30.2 &33.4 \\ \hline 
rerank10 hMM cpd:3 &34.0 &26.8 &{\1 36.4} &27.5 &32.5 \\ \hline 
rerank20 hMM cpd:3 &33.4 &26.5 &{\1 \bf 34.3} &{\1 27.4} &32.0 \\ \hline 
dMM &{\1 32.9} &{\1 26.1} &{\1 \bf 34.3} &{\1 \bf 27.1} &31.6 \\ \hline 
iMM cpd:1 &44.5 &51.0 &53.0 &55.8 &42.1 \\ \hline 
iMM cpd:2 &38.4 &36.1 &41.1 &38.6 &37.0 \\ \hline 
iMM cpd:3 &35.3 &28.5 &39.1 &31.9 &33.3 \\ \hline 
iMM cpd:4 &33.8 &{\1 25.2} &36.9 &30.4 &{\1 30.3} \\ \hline 
iMM cpd:5 &{\1 \bf 32.0} &{\1 \bf 24.1} &{\1 35.9} &29.5 &{\1 \bf 29.5} \\ \hline 
hMM cpd:1 &{\1 33.3} &30.2 &40.6 &30.5 &{\1 30.2} \\ \hline 
hMM cpd:2 &33.6 &26.8 &{\1 36.4} &{\1 27.4} &31.9 \\ \hline 
hMM cpd:3 &{\1 \bf 32.0} &{\1 26.3} &{\1 35.7} &28.1 &{\1 30.0} \\ \hline 
hMM cpd:4 &33.7 &26.6 &36.5 &{\1 27.4} &32.4 \\ \hline 
hMM cpd:5 &{\1 32.8} &{\1 26.0} &37.0 &{\1 27.4} &{\1 31.4} \\ \hline
\end{tabular}
\end{center}
The Data ID \#1-\#5
correspond to ANR for global map database from ``fr101", ``abuilding", ``albert", ``kwing" and ``fr079",
where
``fr079", ``fr101", ``run", ``claxton2" and ``albert" 
are respectively used as dictionary map.
The colored fonts and the bold-face fonts
respectively represent the best 5 methods and the best method
for each data.
\end{table}
}

\newcommand{\figA}{
\begin{figure}[t]
\begin{center}
\FIG{7}{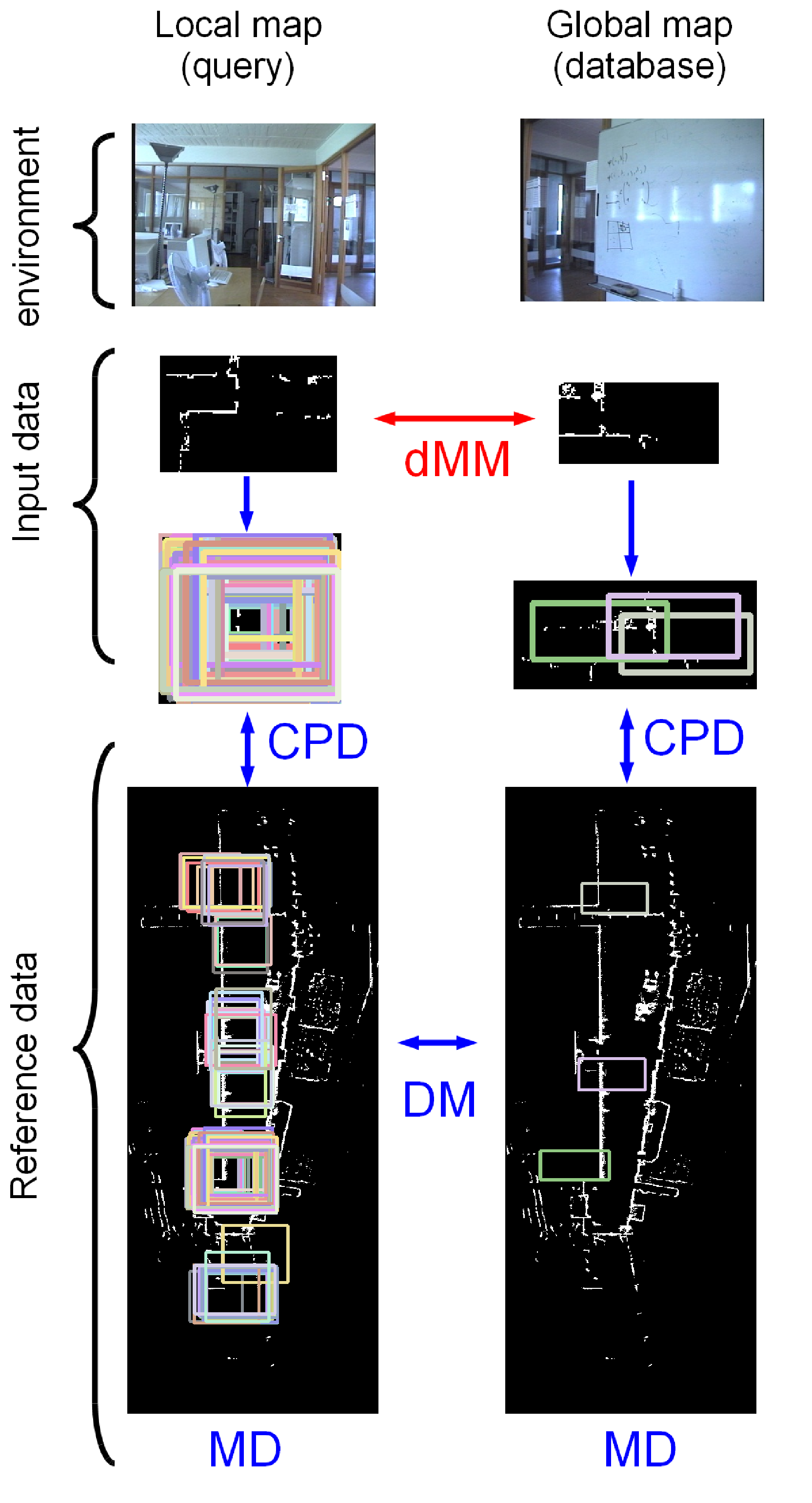}{}
\caption{%
Compared to existing direct map matching (``dMM") methods,
our method indirectly matches between the local map and each of the global maps, 
while using a known reference map as the intermediary.
In offline work, a common pattern discovery (``CPD") process
translates an input local/global map to a compact part-based map descriptor (``MD"), by extracting representative parts (colored bounding boxes) 
that effectively explain an input map from a known reference map.
In online work, a descriptor matcher (``DM") rapidly matches between the compact part-based maps. %
}\label{fig:A}
\vspace*{-5mm}
\end{center}
\end{figure}
}

\newcommand{\figB}{
\begin{figure}[t]
\begin{center}
%\vspace*{-10mm}%
\FIG{8}{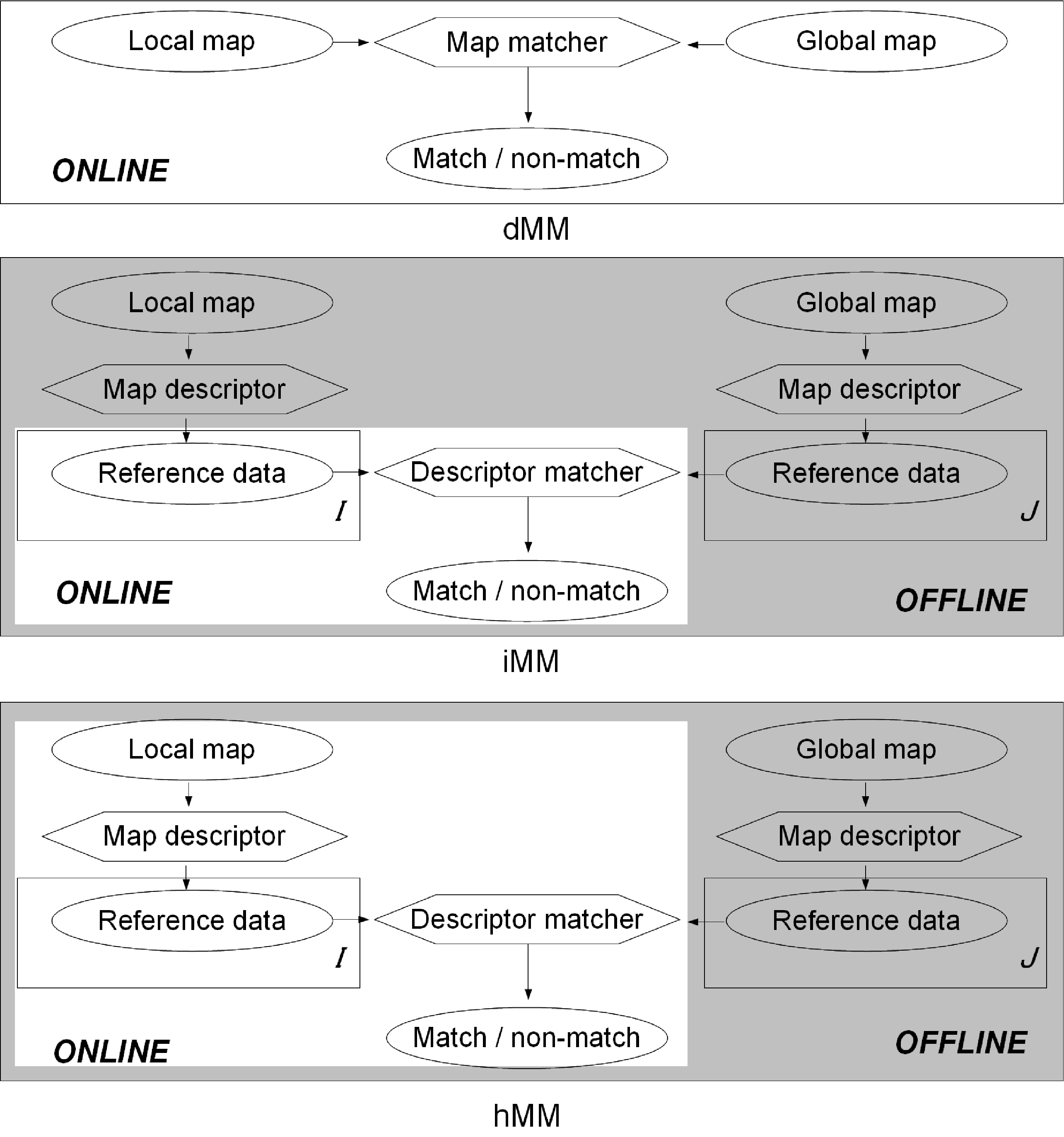}{}
\caption{Three types of map matching schemes considered 
in this study: direct map matching (dMM), indirect map matching (iMM) and hybrid map matching (hMM).
The dMM scheme directly matches between a given map pair, while 
the iMM and the hMM schemes match in an indirect manner using a given dictionary map as intermediate.
While iMM deals with a situation where 
only the compact map descriptors are available,
hMM addresses a situation where 
the original local map data is available.}\label{fig:B}\vspace*{-1mm}%
\end{center}
\end{figure}
}

\newcommand{\figC}{
\begin{figure}[t]
\FIG{8.5}{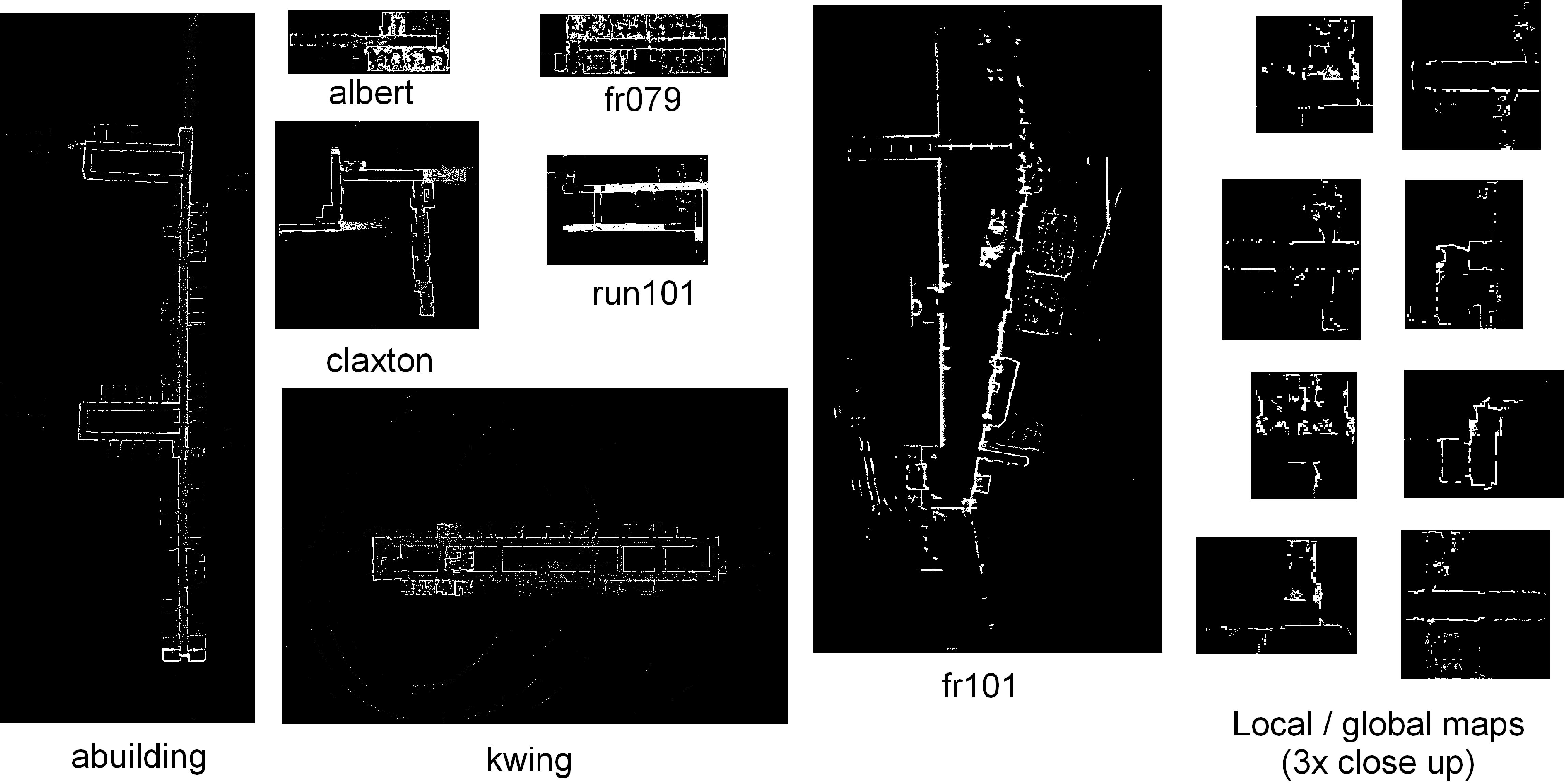}{}
\caption{Datasets used for experiments.
``abuilding",
``albert",
``claxton",
``fr079",
``run101",
and ``kwing" from radish dataset \cite{15} are used as the dictionary maps.
``Local/global maps" are several samples from local and global maps.}\label{fig:C}
\vspace*{-2mm}
\end{figure}
}

\newcommand{\figD}{
\begin{figure}[t]
\begin{center}
\FIGRt{8.5}{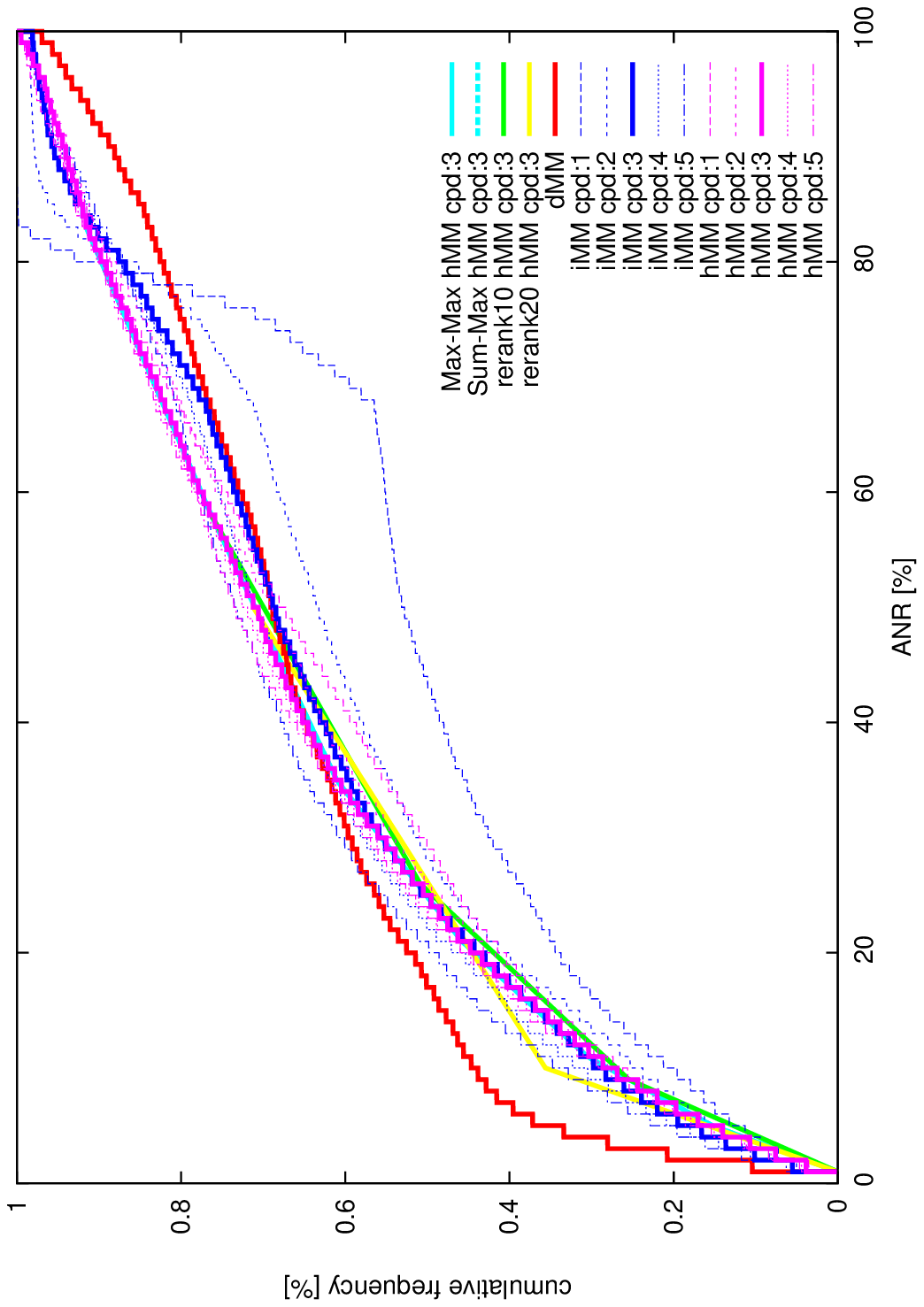}{}%
\caption{Normalized rank for each strategy.}\label{fig:D}%
\vspace*{-5mm}% 
\end{center}
\end{figure}
}

\newcommand{\figE}{
\begin{figure}[t]
\vspace*{2mm}
\begin{center}
\FIGRt{8.5}{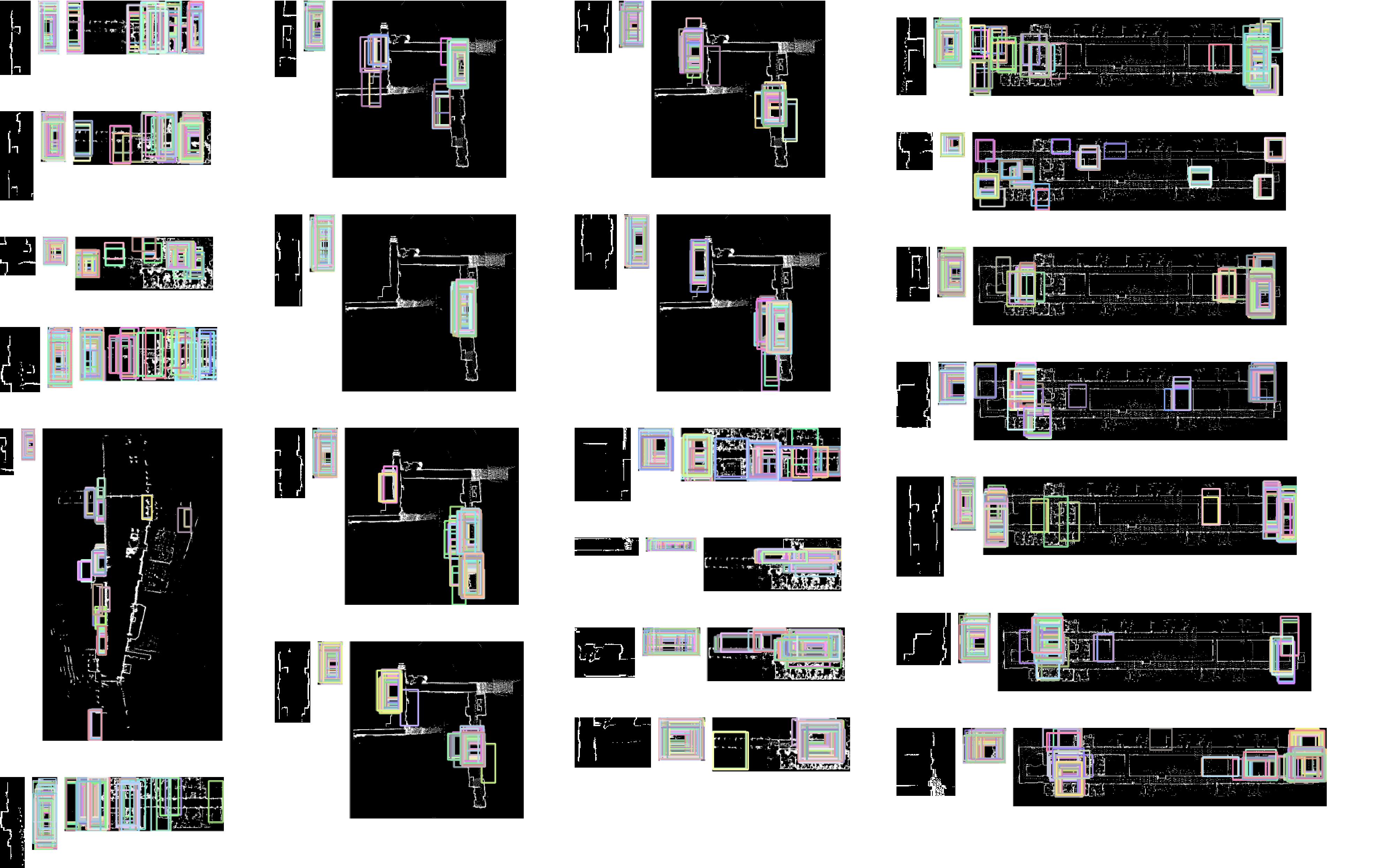}{}
\caption{Results of part-based scene modeling. For each panel, the top subpanel shows the input local map, while the middle and the bottom subpanels respectively illustrate the representative parts discovered w.r.t. the local and the global maps. In each figure, the white dots indicate point cloud in the map, and the colored rectangles indicate bounding boxes that crop the representative parts. }\label{fig:E} 
\end{center}
\vspace*{-5mm}
\end{figure}
}

\maketitle
\thispagestyle{empty}
\pagestyle{empty}

\newcommand{\vh}[2]{
\fbox{
\begin{minipage}{#1cm}
\ \vspace*{#2cm} \\
\end{minipage}
}
}

\renewcommand{\v}[1]{
\vspace*{#1cm}
}

\newcommand{\CO}[1]{}

\begin{abstract}
In this paper, we explore the challenging 1-to-N map matching problem, which exploits a compact description of map data, to improve the scalability of map matching techniques used by various robot vision tasks. We propose a first method explicitly aimed at fast succinct map matching, which consists only of map-matching subtasks. These tasks include offline map matching attempts to find a compact part-based scene model that effectively explains each map using fewer larger parts. The tasks also include an online map matching attempt to efficiently find correspondence between the part-based maps. Our part-based scene modeling approach is unsupervised and uses common pattern discovery (CPD) between the input and known reference maps. This enables a robot to learn a compact map model without human intervention. We also present a practical implementation that uses the state-of-the-art CPD technique of randomized visual phrases (RVP) with a compact bounding box (BB) based part descriptor, which consists of keypoint and descriptor BBs. The results of our challenging map-matching experiments, which use a publicly available radish dataset, show that the proposed approach achieves successful map matching with significant speedup and a compact description of map data that is tens of times more compact. Although this paper focuses on the standard 2D point-set map and the BB-based part representation, we believe our approach is sufficiently general to be applicable to a broad range of map formats, such as the 3D point cloud map, as well as to general bounding volumes and other compact part representations. 
\end{abstract}

\newcommand{\FIGS}[1]{#1}

\figA

\section{Introduction}

Map matching is a technique in which a robot vision system creates a map of its local surroundings and compares that local map to $N$ global maps previously constructed from prior sensor data. 
This technique is the foundation for a wide spectrum of robot vision applications, including viewpoint localization, change detection, alignment, merging, map segmentation, and multi-robot mapping 
\cite{1,2,3}. %
These applications are built on a common pipeline consisting of stages for 
extracting features from maps (e.g., SIFT, shape context, and GIST),
performing visual indexing and 
map database searches to find correspondences (e.g., kd-trees, LSH), 
and map matching to find correspondence sets (points, lines, primitives, and so on) that are inliers to an affine, homography, or geometric transformation, such as by RANSAC matching 
\cite{4,5,16}. %
Another key challenge relates to the recent explosion in global efforts to develop scalable map-building systems. To deal with map matching of such large-scale maps, it is necessary to exploit the compact description of data \cite{3}. To this end, most of previous efforts have focused on compact description of feature data (that is, the first and second stages) relying on PCA or other advanced dimension-reduction techniques. Thus far, few studies have focused on the third stage, the map-matching stage using the compact description of {\it map data}, 
which we address in this paper (Fig. \ref{fig:A}).

Our research in this paper is motivated by two independent techniques derived from the field of computer vision: part model \cite{11,12,10} and common pattern discovery \cite{13,14,21}. First, our basic observation is that a part model is a powerful discriminative model, in which a scene is explained by its parts and their configuration, and a part model can also be {\it compact} if a scene is explained by {\it fewer larger parts}. Second, we are inspired by the recent progress in common pattern discovery (CPD), which is the mining of common patterns among different scenes. Our key approach is to use the CPD techniques as a method for mining {\it fewer larger parts} that effectively explain an input scene from a known reference map. In contrast to existing supervised part models that rely on pre-trained part detectors, our CPD-based approach learns a compact part model with {\it fewer larger parts} in an unsupervised manner, which enables a robot to learn a compact map model without human intervention. 

To develop the above map-matching framework, 
we must address the following questions.
\begin{enumerate}
\item 
How to efficiently mine the parts?
\item
How to compactly describe each part?
\item
How to efficiently match between part-based maps?
\end{enumerate}
To efficiently mine the parts, randomized visual phrase (RVP) \cite{14}, 
which is a recently developed CPD technique, 
is adapted for the map data. 
From our unsupervised modeling standpoint, 
RVP has advantages over existing CPD techniques in that
RVP does not rely on a good segmentation technique, and it 
does not require a priori knowledge of how many common patterns exist in the scene.
To compactly describe each part, we exploit traditional bounding box (BB) based object annotation and knowledge transfer and compactly represent a part as a BB. 
To efficiently match part-based maps, we can make online matching fast and succinct  
(descriptor matcher (DM) in Fig.\ref{fig:A})
because the map matching becomes a low dimensional problem of matching between bounding boxes, and because the CPD can be done in offline as a map-building task. 
The results of our challenging map-matching experiments,
which uses a publicly available radish dataset \cite{15},
show that the proposed approach achieves successful map matching with significant speedup 
and a compact description of map data that is tens of times more compact.

\subsection{Overview of proposed method}

The major undertaking of this work can be summarized as follows.
(a) We propose a first method explicitly aimed at {\it fast succinct map matching}, which purely consists of map-matching subtasks.
These tasks include offline map matching (``CPD"), which attempts to explain an input map by fewer larger parts, and online map matching (``DM"), which efficiently finds correspondence between the part-based maps. (b) Our part-based scene modeling is {\it unsupervised} and uses CPD between the input and a known reference maps, which enables a robot to learn a compact map model without human intervention. 
(c) We present a practical implementation that exploits randomized visual phrase (RVP), 
a state-of-the-art CPD technique, 
and employs compact bounding box (BB) -based object annotation.
Although this paper focuses on the standard 2D point-set map and the BB-based part representation, we believe our approach is sufficiently general to be applicable to a broad range of map formats, such as the 3D point cloud map, as well as to general bounding volumes and other compact part representations.

\subsection{Relation to Other Work}

In the field of computer vision, various types of part models, part detectors, and part appearance models with rigid/deformable templates have been studied with and without learning \cite{11,12,19}.
In \cite{11}, 
a computational part-based model is introduced as a descriptive model of pictorial structures in an object matching task.
In this model, templates of object's parts (such as hair, eyes, mouth for face model) and their configuration are used to determine whether or not the target object is present in the scene. In the above context, template matching techniques have been studied in vision applications using a part-based model, such as object detection/recognition, object comparison, object scaling, and so on.
In \cite{12}, 
a framework is presented for joint categorization and segmentation of object images. 
In this framework, a number of interest points are extracted and compared against the codebook of patches of local appearances: 
probabilistic voting and refinement are then done to segment the object from the background.
In \cite{19}, 
an object detection system based on a mixture of multiscale deformable part models is presented.
This system, which relies on discriminative training of classifiers,
achieves state-of-the-art results in object recognition tasks.
In \cite{33ism}, 
a framework is presented for weakly supervised discovery of common visual structure in highly variable cluttered images by introducing deformable part-based models.
In \cite{35ism}, 
a method is presented for specifying which region model is assigned to each part
by introducing a latent variable.
However, existing part-based methods primarily deal with a supervised setting 
in which a scene category label is explicitly given for training the part detectors.
Those supervised approaches are not well suited for our autonomous robot scenarios 
in which no category or class label is available as training data; 
rather, a part model with fewer larger parts must be learned from raw map data in an unsupervised manner.

Principal component analysis (PCA) and other advanced dimension reduction techniques 
have been widely used for the compact description of feature data for the feature extraction and the visual search stages to accelerate map-matching systems. 
In \cite{16}, 
for example, a method for appearance-guided monocular structure-from-motion for initial motion estimation is presented.
In this method, a place recognition scheme and loop closing are employed for loop detection, which works with a visual word-based approach. 
Earlier works have also presented dimension reduction techniques,
specifically locality sensitive hashing \cite{20} 
and compact projection \cite{22}, 
within RANSAC map matching for large-scale applications \cite{23}. 
For map matching stage, however, the compact description of feature data is of minor importance, and improving visual search performance is not our objective.

The mining of common patterns among scenes, or CPD \cite{13}, remains a challenging task because of  huge search scales and problem domains.
Several different solutions that address the common problem of rotation, translation, occlusion and segmentation have been studied by employing various techniques, 
including the earth mover's distance, co-segmentation, and correspondence growing. 
In \cite{21}, 
a robust CPD technique 
was also developed 
based on 
the correspondence growing framework
in the form of a probabilistic Markov Chain Monte Carlo (MCMC) algorithm.
However,
most of existing frameworks 
focus on segmentation and discovery of repetitive objects or object parts \cite{33}.
In contrast, the current study focuses on {\it use} of the CPD technique as an unsupervised method for compact part-based scene modeling (see Fig.\ref{fig:A}).

Structure of this Paper and Supplementary Material:
The primary focus and core contribution of this paper is our part-based scene modeling and succinct map-matching framework (PartSLAM).
To further support our map representation and matching framework, we provide the following Supplementary Material:
\begin{enumerate}
\item
Supplementary Material A: BB-SAM structural module extraction and single-/cross-map dictionary compression under $D_4$ symmetries with zero-shot foundation segmentation.
\item
Supplementary Material B: Map-FM for semantic-free, variable-length latent map representation adapting sequential and state-space/spectral backbones (TimesFM, Mamba4Cast, and SiMBA) for cross-building place recognition.
\item
Supplementary Material C: Scalable map matching using 2D-Flatness homogeneous foundation integrals, $O(1)$ valid-region spatial pyramid queries, and single-pass 3-DoF partial LiDAR alignment.
\end{enumerate}

\section{Problem}\label{sec:pm}

The map-matching problem addressed in this study is a general 1-to-$N$ matching problem.
In it, 
a local map is given as a query, 
and the system retrieves a size $N$ global map database to find the relevant global map in terms of the similarity score, 
under affine, homography, or geometric transformation.
In our experiments,
we focus on the simple scenario of map matching of 2D point-set maps under rigid transformation.
In this scenario, each point $x\in X$ in a map $X$ is represented by a 2D coordinate.
The similarity score $v$ between a pair of pointset maps 
$(X,Y)$
is obtained by searching for the optimal similarity transform parameters
$T=(t, o)$
where 
$(t, o)$
denote the transformation (i.e. translation and orientation) parameters:
\begin{equation}
v = \max_{T} \sum_{x\in X} P_Y(T(x)); ~~~
P_Y(x)
= 
\begin{cases}
1 & \mbox{$x$ is an inlier of $Y$} \\
0 & \mbox{otherwise}
\end{cases}. \label{eqn:1}
\end{equation}

\subsection{Performance Metrics}\label{sec:pm}

In this study, we develop several different implementations of the 1-to-$N$ map-matching system, and we compare and evaluate them using a standard performance measure of averaged normalized rank (ANR) \cite{17}. To determine ANR, a number of independent map-matching tasks with different queries and databases are conducted.
For each task, the rank assigned to the ground-truth global map by a map matcher of interest is investigated and normalized by the database size $N$. 
ANR is then obtained as the average of the normalized ranks over all the map-matching tasks.

\figB

\section{Approach}

\subsection{Map Matching Schemes}\label{sec:mm}

Three types of map-matching schemes are considered in this paper (Fig.\ref{fig:B}).
The first one is 
the well-studied 
{\it direct map-matching (dMM)} scheme,
which takes a pair of maps as input 
and aims to find 
correspondences that are inliers to a geometric transformation
within a hypothesize-and-test framework.
We implement a standard map-matching method
based on a RANSAC matching algorithm
and use it as a baseline method.
In the 2D map-matching scenario, the
dMM method
iteratively
\begin{enumerate}
\item
hypothesizes a rigid transformation with rotation and translation 
\item
tests the hypothesis by counting inlier points under the hypothesized transformation
\end{enumerate}
for a number of times, and, after the iteration, 
the hypotheses are ranked in descending order of the inlier count.

The second map-matching scheme we consider is {\it indirect map matching (iMM)}.
This scheme is based on a compact description of map data,
and it matches a given map pair in an indirect manner using the dictionary map as intermediatry.
The iMM scheme consists of three steps:
\begin{enumerate}
\item
the common pattern discovery (CPD) mines similar common patterns (i.e., parts) between each input local/global map and the dictionary map;
\item
the map descriptor (MD) selects the best parts that effectively explain an input query/database map;
\item
the descriptor matcher (DM) evaluates the likelihood 
that a given query map and a database map are a match pair;
\end{enumerate}
then it ranks the global maps in terms of the likelihood score. Fig.\ref{fig:A} illustrates the relationship among CPD, MD and DM.

The third one is {\it hybrid map matching (hMM)},
which is a hybrid of the above direct and indirect map matching.
This scheme is motivated by an asymmetric relationship between the query and the database maps.
For example, in a situation where the map matching is conducted online while building the local map, the original local map can be used as a query map, while the original database maps are not available because of memory limitations.
The algorithm of hMM is basically the same as that of iMM.
A key difference between iMM and hMM is the number of parts that are allowed per map (Fig.\ref{fig:B}). 
In the latter case,
we are allowed to extract a large number (e.g., 100) 
of parts from the original local map that is available online.

The following subsections explain 
the building blocks of the proposed map-matching framework: common pattern discovery, 
map descriptor, and descriptor matcher. 
Several different combinations of strategies will be evaluated in the experimental section.

\subsection{Common Pattern Discovery (CPD)}\label{sec:cpd}

Inspired by the randomized visual phrase (RVP) \cite{14},
a recently developed CPD technique,
we randomly partition 
an input map into overlapping bounding boxes (BBs)
and view each BB as a potential candidate for the parts.
The CPD task is therefore formulated as the problem of finding the {\it best} BB pairs to crop parts. In this problem,
\begin{itemize}
\item
the {\it keypoint BB} crops the input query/database map to indicate the location of a part to be explained out; and
\item
the {\it descriptor BB} crops the dictionary map to indicate the appearance of the part.
\end{itemize}

We believe that good parts should meet the following requirements:
\begin{itemize}
\item
Maximality Criteria (MC),
\item
Appearance Similarity (AS), and
\item
Geometric Criteria (GC).
\end{itemize}
In MC, each part should explain a large portion of the input map.
In AS, the appearance of each part should be 
{\it similar} between the input and the dictionary maps.
And in GC, 
the constellation of parts should be consistent between 
the maps.

To address the above requirements, 
we first retrieve BBs that satisfy MC and GC;
among the retrieved BBs, we select the one that receives the highest AS. 
The MC criteria 
judges whether or not
the ratio of points within the cropped part to all input map points exceeds a pre-defined threshold 
$T^{size}=0.9$.
The GC criteria 
judges whether or not
each pair of descriptor BBs overlaps. 
For each candidate part pair 
that satisfies both 
MC and GC criteria,
we construct a temporal occupancy grid map
$R$ 
from the dictionary map
with a fixed resolution (in implementation, 0.1 m),
and we evaluate the AS
between the descriptor BB pair $(B, \hat{B})$:
\begin{equation}
f^{AS}(B, \hat{B}) = 
\max_{T} \sum_{x\in X} P_R(T(x)), \label{eqn:as}
\end{equation}
where the meanings of 
$P_R$ and $T(\cdot)$
are the same as in (\ref{eqn:1}).

\subsection{Map Descriptor (MD)}\label{sec:md}

The map descriptor obtains a compact description of an input query/database map.
For example, 
given a given number of parts output by the CPD
(in implementation, 
100 parts,
as shown in Fig.\ref{fig:E}),
each part is represented by a pair of BBs (i.e., a keypoint and a descriptor BBs) as mentioned earlier.
Among them,
the map descriptor selects a small number of $K$ parts
that effectively and compactly explain the input map.
Currently,
we use
a simple selection scheme
in which 
the given parts 
are ranked
in descending order of the AS score;
the $K$ top-ranked parts are then output.
In our experiments,
we test 
several different settings of $K$
and investigate their influence on the map-matching performance.

Our approach enables the user to make a trade-off between accuracy and compactness by adjusting the number of parts $K$ per input map. A part is compactly represented by the pose and the shape of a BB. Currently, each $x,y$ coordinate of the pose is represented in spatial resolution 0.1 m and memory space consumed per part is 42 bit. For instance, when an input map is represented by 3 parts, the space required per input map is $3 \times 42$ bit=126 bit $<16$ byte, which is an extremely compact map descriptor.

\subsection{Descriptor Matcher (DM)}\label{sec:mm}

We now describe how to
efficiently evaluate similarity between map descriptors.
Recall that a local map and global maps are described by collections of descriptor BBs, $\{B_l^i\}$ and $\{B_g^j\}$.
Intuitively, a pair of descriptor BBs with larger overlap indicates that it is likely that the regions on the global and local maps cropped by the BB pair are similar to each other, and vice versa.
To evaluate similarity 
between a pair of such local and global maps, we introduce a measure of pair-wise similarity between a BB pair
\begin{equation}
\hspace*{-5mm}
f^{RS}(B_l^i, B_g^j) =
\frac{1}{\left(Size(B_l^i)Size(B_g^j)\right)^{1/2}}
Size(Overlap(B_l^i, B_g^j)),
\end{equation}
where 
$Size(\cdot)$
is the area of a given BB on the 2D map,
and $Overlap(\cdot,\cdot)$
is the overlap between a given BB pair.
Based on the terminology,
we design
three types of evaluation measures,
\begin{itemize}
\item[]Max-max:
\begin{equation}
F^{MM}=\max_i \max_j f^{RS}(B_l^i, B_g^j)
\end{equation}
\item[]Sum-max:
\begin{equation}
F^{SM}=\sum_i \max_j f^{RS}(B_l^i, B_g^j)
\end{equation}
\item[]Sum-max-weighted:
\begin{equation}
F^{SMW}= \sum_i \max_j f^{RS}(B_l^i, B_g^j)
\Bigl( f^{AS}(B_l^i, \hat{B}_l^i) \Bigr) \label{eqn:smw}
\end{equation}
\end{itemize}
The intuition behind the third measure 
in (\ref{eqn:smw})
is that 
we weight 
the pair-wise similarity between elements in a BB pair 
according to the approximation accuracy,
which is measured by reusing
the appearance similarity 
$f^{AS}$ described in (\ref{eqn:as}).

\figC

\section{Experiments}

\CO{

\begin{verbatim}

10.0.54.4
1/ input:fr101 dict:fr079
3/ input:albert dict:run
5/ input:kwing dict:claxton2
9/ input:fr079 dict:albert 
10/ input:abuilding dict:fr101 

\end{verbatim}

}

We conduct challenging map-matching experiments to validate the benefits of the proposed approach.
In the following,
we first describe the datasets and the map-matching tasks used in the experiments, and we then present our results on performance comparison among different map-matching schemes and strategies, time and space efficiency.

\subsection{Dataset}

For map matching, we have created a large-scale map collection 
from the publicly available radish dataset\cite{15},
which consists of
logs of odometry and laser data 
acquired by a car-like mobile robot 
in indoor environments.
We created a collection of local/global maps
using a scan matching algorithm
from each of 5 different datasets---named
fr101, albert, kwing, fr079 and abuilding---which respectively were obtained by
the mobile robot's 79--295m travel
corresponding to 521-5299 scans.
We also used
fr079, run, claxton2, albert and fr101,
respectively,
as the dictionary maps for each dataset.

Fig.\ref{fig:C} shows the original datasets and random samples from local and global maps.
The map collection consists in total of more than 13,000 submaps.
Using the entropy minimization technique in \cite{18},
the $xy$-axes of each map were aligned 
with the ``Manhattan-like structure".
As shown in Fig.\ref{fig:C},
our map collection
contains many near duplicate maps,
which makes the map matching a challenging task.

\figE

\subsection{Map Matching Tasks}

Recall that
the objective of map matching is 
giving a local map as a query
to find a relevant map from the global map database.
The relevant map is defined as a global map that satisfies two conditions:
1) Its pose is near the query map's pose within a pre-defined range, where the pose of a map is defined as the center of gravity of the map's point cloud, 
and 2) Its distance traveled along the robot's trajectory is distant from that of the local map, such as in a ``loop-closing" situation in which a robot, after traversing a loop-like trajectory, returns to a previously explored location. 

For each relevant map pair,
a map-matching task is conducted by 
using a local map and a size $N$ global map database,
which consists of 
one relevant map and $(N-1)$ random irrelevant maps.
The spatial resolution of the temporal occupancy map used by the iMM and hMM is set to 0.1 m.
The descriptor matcher uses ``Sum-Max" and ``Sum-Max-Weighted" 
as the default strategies, 
respectively, for iMM and hMM. 
We have implemented several combinations of map-matching algorithms in C++, and successfully tested on various maps. 

Shown in Fig.\ref{fig:E} are the results of part-based scene modeling, in which an input local/global map is explained by a pool of representative parts discovered by CPD. 
The CPD task was mostly successful, 
and similar input maps were characterized by similar descriptor BBs.
Quantitative evaluations of our CPD-based approach will also be provided in the following subsections.

\tabA

\subsection{Comparison among dMM, iMM, and hMM}

For performance comparison, we evaluated the averaged normalized rank (ANR) introduced in Section \ref{sec:pm} for all the three basic map-matching methods: dMM, iMM and hMM. All map-matching tasks were conducted using 13,592 different local and global map databases. Fig.\ref{fig:D} shows the normalized cumulative histograms of rank assigned to the relevant map, 
and Table \ref{tab:A} summarizes the ANR performance. 

The proposed hMM achieved results comparable result to the dMM method when the number of parts per map $K$ was set to 3--5, despite the fact that the proposed hMM achieves tens of times more compact map representation than dMM, as will be shown in Section \ref{sec:space}.
Much of the analysis from the previous sections on map matching also holds for the dataset considered here. By extracting a number of parts from the query local map and comparing their attributes (i.e., BB pairs) against each map in the global map database, our method is able to successfully perform the map matching. On the other hand, the presented iMM method is satisfactory when we set the number of parts per input map to a large value, 5. 
This is because the input map tends to be well explained by the larger set of common object patterns, which is still a compact representation,
as will be discussed in Section \ref{sec:space}.

\subsection{Effect of Matching Strategies}

We also compared three different implementations of the descriptor matcher for the hMM method, which was explained in Section \ref{sec:mm}. 
However, the difference among them is turned out to be subtle, as shown by
``Max-Max hMM", ``Sum-Max hMM", and ``hMM"
in Fig. \ref{fig:D} and in Table \ref{tab:A}.

\subsection{Effect of Reranking}

For precise map matching, we tested another ``reranking" strategy, which is in essentially a cascade of dMM and hMM. 
The strategy is motivated by the observation that, according to Fig. \ref{fig:D}, the standard dMM method is good at ranking top-ranked elements more precisely than the hMM method. 
Using the cascaded method,
the top $R$ elements ranked by the proposed hMM method were further input to the dMM method, and their rankings were reordered according to the match score assigned by the dMM method. 
We set the number $R$ of elements to be reranked as $R=10$ and $R=20$ and conducted the map-matching tasks. The result is similar or better than the proposed hMM method, as shown in Fig.\ref{fig:D}. 
On the other hand,
a major drawback of this strategy is memory usage, as
the dMM method that the strategy employs requires the information from the original global maps.

\figD

\subsection{Time Efficiency}

Matching between a query of a local map and a database map
took less than 5 ms on a 2.4 Ghz Intel laptop. The time complexity was linear in terms of the number of parts per map and the database size. We tested different settings of the number of parts per map, 1, 2, 3, 4 and 5, and got time cost 1.3, 2.3, 3.3, 4.2 and 5.0 msec. In contrast, the conventional dMM algorithm took an average of 33 msec. Its complexity was linear in the database size, and its time cost could be quite high when using a large size database, e.g., with tens of millions of landmarks. On the other hand, one of main concerns of our method is the time cost for CPD. 
The cost in principle depends on the size of the dictionary map,
this can be non-negligible when using a
large-scale dictionary map.
However, because cost is not dependent on database size, 
it therefore does not affect the scalability of our proposed approach. 
It is beyond the scope of this paper to discuss how to suppress 
the cost for CPD performed as part of the map-building task,
and it will have to be addressed in future work.

\subsection{Space Efficiency}\label{sec:space}

Space cost for our map descriptor is 
linear to 
the number of parts per map
and the cost per part.
In general, a part is defined by 
a pair of BBs,
i.e.,
a keypoint BB 
$[x_{begin}, x_{end}]$
$\times$
$[y_{begin}, y_{end}]$,
and a descriptor BB
$[\hat{x}_{begin}, \hat{x}_{end}]$
$\times$
$[\hat{y}_{begin}, \hat{y}_{end}]$.
Because the shapes of both BBs are the same,
our database omits
$(\hat{x}_{end}, \hat{y}_{end})$
as redundant information.
In implementation,
7 bits are used for each of 
$x_{begin}$, $x_{end}$, $y_{begin}$, $y_{end}$.
$(x_{begin}, y_{begin})$
indicates a point on the dictionary map
and is represented in 14 bits.
In total,
$7 \times 4 + 14$ = 42 bits
are used for each part.
For instance,
the proposed iMM method consumes
$42 \times 3$ bits = 126 bits
when it uses 3 parts per map.
Compared with the dMM method,
which costs
$(7 + 7) \times 500$ = 7000 bits
for typical size 500 point cloud map,
our map-matching method
achieves compact description of map data
that is tens of times more compact.

\section{Conclusion and Future Work}

In this paper, we proposed a first method explicitly aimed at {\it fast succinct map matching}, which consisted only of map-matching subtasks. These tasks included offline map matching (CPD) attempts to explain an input map using {\it fewer larger} parts, and online map matching (DM) to efficiently find correspondence between the part-based maps. Our part-based scene modeling was unsupervised and used CPD between the input and a known reference maps, which enables a robot to learn a compact map model without human intervention. We presented a practical implementation that leverages the randomized visual phrase (RVP), a state-of-the-art CPD technique, and a compact BB-based part descriptor. We showed that the proposed approach achieves successful map matching with significant speedup and a compact description of map data that is tens of times more compact.

In future work we plan to continue our exploration of CPD, and consider how to achieve good tradeoff between the compactness and the accuracy. We will also investigate efficient methods for building the part-based maps. It would be interesting to extend the methods to support a compressive SLAM task \cite{22}, which aims at the incremental building of the compact map model. 

Another obvious improvement is to optimize the reference maps. Because our approach is designed to represent an input map by cropped reference maps, it would not be suitable for general cases where whole regions of the input map are dissimilar from the reference map. In the future we shall study a way for automatically choosing the reference maps adaptively for given global maps. 

Finally, our approach could be extended to a broad range of map formats, such as the 3D point cloud map, as well as to other compact part representations, such as general bounding volumes. In this direction, the approach was recently extended to an alternative map format, view sequence map, by introducing an unsupervised part-based scene modeling technique, ``bag-of-bounding-boxes (BoBB)" \cite{32}.

\bibliographystyle{unsrt}
\bibliography{ncd}

\setcounter{section}{0}
% \setcounter{figure}{0}
% \setcounter{table}{0}
% \setcounter{equation}{0}
% \setcounter{page}{1} % ページ番号も1からに戻す場合（不要なら削除）

% --- 3. ★ 2回目の \maketitle を許可するための内部フラグ調整 ---
\if@twocolumn
  \twocolumn
\fi
\makeatother

\vspace*{1ex}
{\centering
  {\Large \bfseries Supplementary Material A: BB-SAM Structural Module Extraction and Dictionary Compression \par}
  \vspace{1ex}
  \vspace{3ex}
}

\begin{table*}[t]
\centering
\caption{Top Cross-Map Part Compression (CPS) Pairs Across 56 Evaluated Map Combinations}
\label{tab:appendix_cps_top}
% \resizebox{\textwidth}{!}{%
\begin{tabular}{llcccc}
\toprule
\textbf{Dict Map ($I_{\text{dict}}$)} & \textbf{Target Map ($I_{\text{target}}$)} & \textbf{Dict ($K$ / pts)} & \textbf{Inst. ($N_{\text{inst}}$)} & \textbf{Covered ($N_{\text{cov}}$)} & \textbf{Ratio (Net Saved)} \\
\midrule
\texttt{orebro} & \texttt{fr079-complete} & $4$ ($716\text{ pts}$) & $15$ & $4,878$ ($28.6\%$) & $\mathbf{6.41\times}$ ($+4,117\text{ pts}$) \\
\texttt{aces\_publicb} & \texttt{fr079-complete} & $5$ ($929\text{ pts}$) & $22$ & $6,154$ ($36.0\%$) & $\mathbf{6.18\times}$ ($+5,159\text{ pts}$) \\
\texttt{orebro} & \texttt{edmonton} & $3$ ($522\text{ pts}$) & $18$ & $3,327$ ($13.8\%$) & $\mathbf{5.78\times}$ ($+2,751\text{ pts}$) \\
\texttt{edmonton} & \texttt{fr079-complete} & $3$ ($786\text{ pts}$) & $12$ & $4,564$ ($26.7\%$) & $\mathbf{5.55\times}$ ($+3,742\text{ pts}$) \\
\texttt{orebro} & \texttt{csail} & $3$ ($409\text{ pts}$) & $10$ & $1,801$ ($17.2\%$) & $\mathbf{4.10\times}$ ($+1,362\text{ pts}$) \\
\texttt{edmonton} & \texttt{fr-campus} & $9$ ($1,284\text{ pts}$) & $28$ & $5,250$ ($12.2\%$) & $\mathbf{3.84\times}$ ($+3,882\text{ pts}$) \\
\bottomrule
\end{tabular}%
% }
\end{table*}

\begin{table*}[t]
\centering
\caption{Dictionary Map ($I_{\text{dict}}$) Generalizability Across All 7 Unseen Target Maps}
\label{tab:appendix_dict_suitability}
% \resizebox{\textwidth}{!}{%
\begin{tabular}{lcccc}
\toprule
\textbf{Source Dict Map ($I_{\text{dict}}$)} & \textbf{Mean Ratio} & \textbf{Min / Max Ratio} & \textbf{Win Rate ($>1\times$)} & \textbf{Cumulative Saved} \\
\midrule
\texttt{orebro} & $\mathbf{3.01\times}$ & $1.88\times$ / $6.41\times$ & $\mathbf{7 / 7}$ ($100\%$) & $\mathbf{+14,993\text{ pts}}$ \\
\texttt{edmonton} & $\mathbf{2.91\times}$ & $1.70\times$ / $5.55\times$ & $\mathbf{7 / 7}$ ($100\%$) & $+13,031\text{ pts}$ \\
\texttt{aces\_publicb} & $\mathbf{2.88\times}$ & $1.34\times$ / $6.18\times$ & $\mathbf{7 / 7}$ ($100\%$) & $+12,540\text{ pts}$ \\
\texttt{mexico} & $1.97\times$ & $1.21\times$ / $2.50\times$ & $\mathbf{7 / 7}$ ($100\%$) & $+10,977\text{ pts}$ \\
\texttt{csail} & $1.86\times$ & $0.70\times$ / $3.61\times$ & $6 / 7$ ($85.7\%$) & $+9,270\text{ pts}$ \\
\texttt{fr-campus} & $1.78\times$ & $1.10\times$ / $2.33\times$ & $\mathbf{7 / 7}$ ($100\%$) & $+7,008\text{ pts}$ \\
\texttt{belgioioso} & $1.45\times$ & $0.89\times$ / $3.31\times$ & $6 / 7$ ($85.7\%$) & $+4,519\text{ pts}$ \\
\texttt{fr079-complete} & $1.14\times$ & $0.59\times$ / $1.75\times$ & $4 / 7$ ($57.1\%$) & $+3,647\text{ pts}$ \\
\bottomrule
\end{tabular}%
% }
\end{table*}

\begin{figure*}[t]
  \centering
  \includegraphics[width=\textwidth]{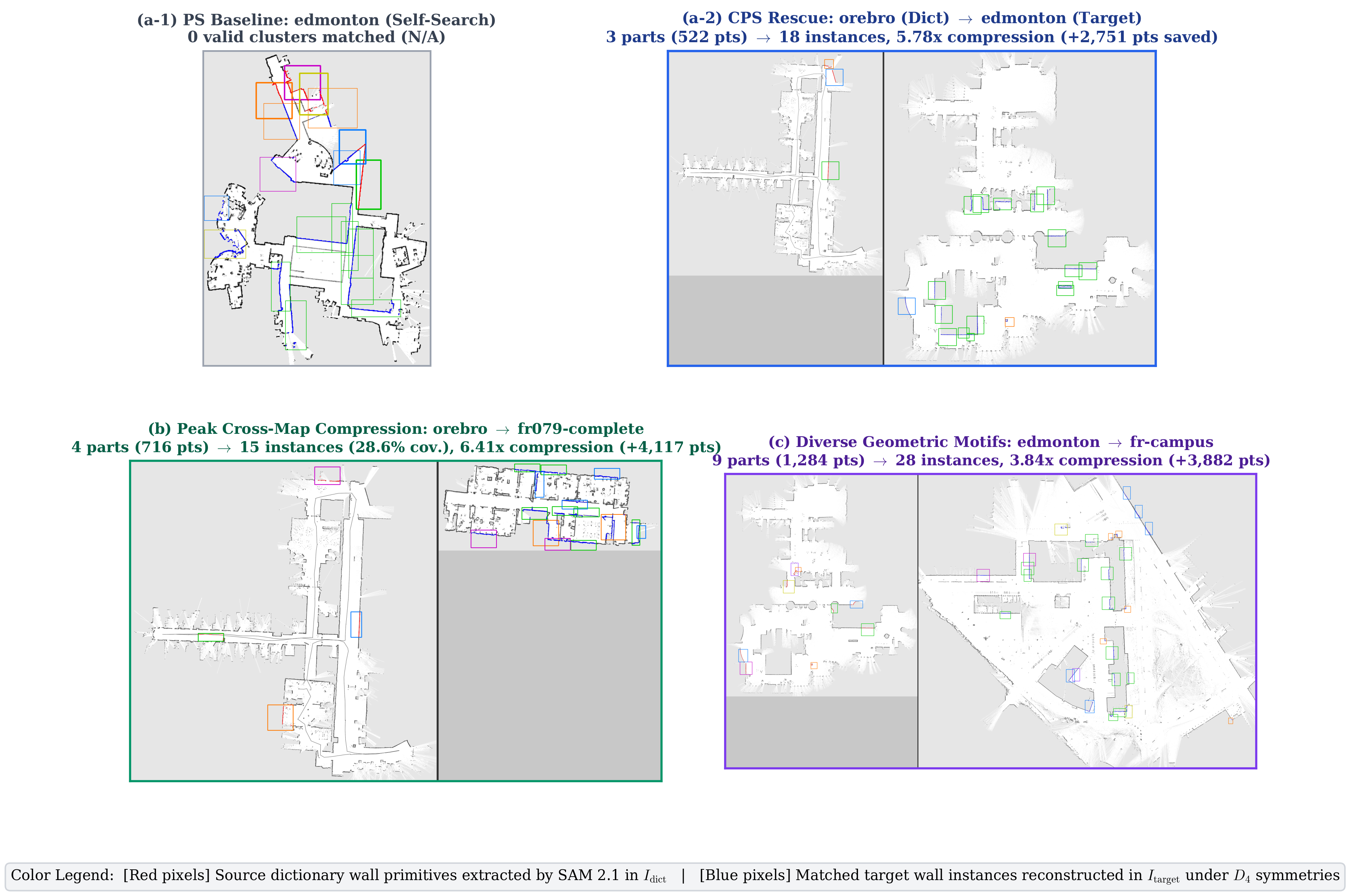}
  \caption{Qualitative comparison of Single-Map Part Compression (PS) and Cross-Map Part Compression (CPS) across representative indoor SLAM maps. In each CPS panel, the left sub-map shows source dictionary wall primitives (red pixels) extracted by SAM 2.1 in $I_{\text{dict}}$, and the right sub-map shows matched target instances (blue pixels) reconstructed in $I_{\text{target}}$ under $D_4$ symmetries. \textbf{(a-1 vs.\ a-2)} While \texttt{edmonton} yields zero self-repeating clusters under PS due to its asymmetric layout, transferring $3$ compact primitives ($522\text{ pts}$) from \texttt{orebro} reconstructs $18$ structural instances at $5.78\times$ compression ($+2,751\text{ pts}$ saved). \textbf{(b)} Peak cross-map compression ($6.41\times$): $4$ parts ($716\text{ pts}$) from \texttt{orebro} cover $28.6\%$ of \texttt{fr079-complete} across $15$ instances ($+4,117\text{ pts}$ saved). \textbf{(c)} Cross-building transfer from \texttt{edmonton} ($9$ parts, $1,284\text{ pts}$) to the large-scale \texttt{fr-campus} environment ($28$ matched instances, $3.84\times$ compression, $+3,882\text{ pts}$ saved), illustrating clean wall skeleton extraction across diverse building geometries.}
  \label{fig:cps_qualitative}
\end{figure*}

% ==============================================================================
% Section 1: Problem Formulation and Theoretical Analysis
% ==============================================================================
\section{Neuro-Symbolic Formulation and Theoretical Analysis}
\label{sec:supp_bbsam_theory}

In the main body of this paper, we formulated Common Pattern Discovery (CPD) to construct a compact part-based Map Descriptor (MD) using Randomized Visual Phrases (RVP) and axis-aligned Bounding Boxes (BBs) without pre-trained part detectors. In Section~V, we further identified automated reference map selection and incremental compressive SLAM as key open challenges for real-world deployment.

In this supplementary material, we address these challenges in a practical operational setting where a mobile robot navigates indoor environments using three readily available resources at zero task-specific training cost:
\begin{enumerate}
    \item \textbf{Operational Map Reuse:} An existing occupancy grid map $I_{\text{dict}}$ maintained for navigation is reused as a reference dictionary to compress either itself (\textbf{Single-Map Part Compression; PS}, $I_{\text{dict}} = I_{\text{target}}$) or an unseen target map (\textbf{Cross-Map Part Compression; CPS}, $I_{\text{dict}} \neq I_{\text{target}}$).
    \item \textbf{Symmetry-Aware Pattern Discovery:} Repeating architectural structures are mined within or across maps under dihedral group symmetries ($D_4$: $90^\circ$ rotations and reflections).
    \item \textbf{Zero-Shot Foundation Segmentation:} Prompting a vision foundation model with candidate BBs suppresses clutter (e.g., furniture, LiDAR scattering) to isolate the pure architectural wall skeleton, leveraging the universal geometric prior of structural boundaries learned from natural images.
\end{enumerate}

We term this framework \textbf{BB-SAM} (\textbf{B}ounding-\textbf{B}ox-prompted \textbf{S}egment-\textbf{A}nything-class \textbf{M}odel prior). While we instantiate BB-SAM with Segment Anything Model 2.1 (SAM 2.1), the formulation is model-agnostic and extends to any promptable segmentation or open-vocabulary detection backbone. Methodologically, BB-SAM operates as a \emph{neuro-symbolic map compression framework}: it couples sub-symbolic perceptual filtering (foundation segmentation) with discrete combinatorial optimization over $D_4$ transformations. By distilling raw occupancy grids into a compositional dictionary of clean architectural primitives---including straight, diagonal, and curved walls, L/T-junctions, and isolated rooms---BB-SAM serves both as a standalone structural map compressor (PS/CPS) and as a high-purity dictionary generator for our fast online descriptor matcher (iMM/hMM).

\subsection{Complementarity of BB and SAM (Succinctness vs.\ Accuracy)}
Achieving our core design principle---explaining a scene with \emph{fewer larger parts}---requires jointly optimizing \emph{succinctness} (model compactness) and \emph{accuracy} (geometric fidelity) along the rate--distortion frontier. Relying on either discrete BB discovery or neural segmentation alone incurs a fundamental failure mode:
\begin{itemize}
    \item \textbf{BB-Only ($\rightarrow$ Loss of Succinctness):} Without neural background separation, all occupied cells inside a rectangular BB---including furniture clutter, sensor noise, and adjacent walls---are baked into the canonical primitive. To prevent cross-environment contamination in CPS, large structures (e.g., whole rooms or corridors) cannot be extracted as unified symbols and must instead be fragmented into tiny, noise-free patches. This inflates the dictionary size $K$ and instance count $N_{\text{inst}}$, degrading \emph{succinctness}.
    \item \textbf{SAM-Only ($\rightarrow$ Loss of Accuracy):} Applying foundation segmentation globally without spatially bounded BB prompts introduces severe scale and granularity ambiguity (e.g., segmenting a single wall segment in one room while merging an entire hallway wing in another). Because identical motifs are not grounded into consistent symbolic units, geometric matching fails, degrading \emph{accuracy} and forfeiting the fixed-bit compactness of BB descriptors.
\end{itemize}
Thus, BB discovery provides consistent spatial framing and compositional granularity, while SAM-class filtering performs neural-to-symbolic grounding by isolating pure structural skeletons from clutter.

\subsection{Chicken-and-Egg Dependency, Ill-Posedness, and Alternating Decomposition}
Jointly recovering a canonical part dictionary $\mathcal{D}$ from $I_{\text{dict}}$ and instance placements $\mathcal{M}$ on $I_{\text{target}}$ from unlabeled occupancy grids exhibits a \emph{chicken-and-egg} circularity: localizing optimal BBs requires knowing which pixels belong to a coherent wall primitive rather than clutter, whereas prompting SAM to extract a clean primitive requires well-localized BBs. Moreover, because 2D floorplans admit infinitely many hierarchical decompositions, this inverse problem is inherently \emph{ill-posed}.

BB-SAM resolves this circularity via a two-stage alternating decomposition akin to a single iteration of coordinate descent (or an Expectation--Maximization step):
\begin{itemize}
    \item \textbf{E-step Analogy (Symbolic Spatial Support and Group-Pose Estimation):} Given raw grid priors, BB-SAM estimates discrete spatial supports (paired BBs across $I_{\text{dict}}$ and $I_{\text{target}}$) and $D_4$ group transformations of repeating motifs via symmetry-aware ZNCC region growing and transitive consolidation.
    \item \textbf{M-step Analogy (Neural Primitive Refinement):} Given these estimated BB supports, BB-SAM refines the canonical symbolic part point sets $\mathcal{D} = \{P_k\}_{k=1}^{K}$ via batched SAM 2.1 inference and normalized local geometric verification, rejecting outlier clutter. To prevent boundary leakage and free-space inflation during mask extraction (at $0.1\text{ m/px}$ grid resolution), candidate seeds and masks are filtered using a $6\text{ px}$ ($0.6\text{ m}$) border-margin seed guard and a $2\text{ px}$ ($0.2\text{ m}$) empty-space over-expansion guard (rejecting masks extending $>2\text{ px}$ into free space).
\end{itemize}

\subsection{Optimization Formulation and Computational Complexity}
Let $\mathcal{W}_{\text{target}} \subset \mathbb{Z}^2$ denote the set of occupied grid cells in $I_{\text{target}}$. Grounded in the \emph{Minimum Description Length (MDL)} principle---which operationalizes the rate--distortion frontier between compact encoding (rate) and geometric reconstruction fidelity (distortion)---dictionary compression is formulated as the combinatorial optimization problem $\Pi_{\text{BB-SAM}}$. Specifically, we optimize over the dictionary size $K$, instance count $N_{\text{inst}}$, source bounding boxes $\mathcal{B} = \{B_k^{\text{dict}}\}_{k=1}^{K}$ in $I_{\text{dict}}$, and target instance placements $\mathcal{M} = \{(k_i, T_i)\}_{i=1}^{N_{\text{inst}}}$. Each source box induces a canonical primitive $P_k \subset \mathbb{Z}^2$, while each placement pairs a primitive index $k_i \in \{1, \dots, K\}$ with a discrete Euclidean/reflective transformation $T_i = (t_i, g_i) \in \mathbb{Z}^2 \rtimes D_4$, parameterized by a 2D translation $t_i = (x_i, y_i) \in \mathbb{Z}^2$ and a dihedral group element $g_i \in D_4$ (encoding $90^\circ$ rotations and reflections).

To directly evaluate raw grid compression in units of occupied cells (complementing the fixed 42-bit BB descriptor encoding used for fast online matching in the main paper), we minimize the total description cost:
\begin{align}
\min_{K,\, N_{\text{inst}},\, \mathcal{B},\, \mathcal{M}} \quad & C_{\text{total}}(\mathcal{D}, \mathcal{M}) = \underbrace{\sum_{k=1}^{K} |P_k|}_{\text{Model Cost}} + \underbrace{\lambda N_{\text{inst}}}_{\text{Code Cost}}, \label{eq:opt_obj} \\
\text{s.t.} \quad & P_{k} = \Phi_{\text{SAM}}\!\left(I_{\text{dict}},\, B_{k}^{\text{dict}}\right), \quad \forall k \in \{1, \dots, K\}, \label{eq:opt_sam} \\
& B_i^{\text{target}} = \mathrm{AABB}\!\left(T_i(P_{k_i})\right), \quad \forall i \in \{1, \dots, N_{\text{inst}}\}, \label{eq:opt_bb} \\
& \vec{d}_{\text{CD}}\!\left(T_i(P_{k_i}),\, \mathcal{W}_{\text{target}} \cap B_i^{\text{target}}\right) \le \tau_{\text{CD}}, \quad \forall i, \label{eq:opt_geom_cd} \\
& \eta_{\text{in}}\!\left(T_i(P_{k_i}),\, \mathcal{W}_{\text{target}} \cap B_i^{\text{target}}\right) \ge \tau_{\text{in}}, \quad \forall i, \label{eq:opt_geom_in} \\
& \left|\bigcup_{i=1}^{N_{\text{inst}}} \left(T_i(P_{k_i}) \cap \mathcal{W}_{\text{target}}\right)\right| \ge \rho_{\min} |\mathcal{W}_{\text{target}}|, \label{eq:opt_cov}
\end{align}
where $\lambda = 3$ accounts for the three discrete pose parameters $(x_i, y_i, g_i)$ per instance, $\mathrm{AABB}(\cdot) \subset \mathbb{Z}^2$ denotes the integer lattice set enclosed by the axis-aligned bounding box, $\Phi_{\text{SAM}}(\cdot, \cdot)$ is the black-box promptable segmentation operator, $\vec{d}_{\text{CD}}(A, B) = \frac{1}{|A|}\sum_{a \in A}\min_{b \in B}\|a - b\|_2$ and $\eta_{\text{in}}(A, B) = \frac{1}{|A|}|\{a \in A \mid \min_{b \in B}\|a - b\|_2 \le \delta_{\text{in}}\}|$ denote the forward (primitive-to-target) Chamfer distance and inlier ratio (tolerating unmodeled interior clutter within $B_i^{\text{target}}$), and $\rho_{\min} \in (0, 1]$ is the target coverage ratio (with uncovered residual cells stored verbatim in a hybrid representation).

\noindent\textbf{Theorem 1 (Strong NP-Hardness and Inapproximability).}
\emph{Even when candidate part shapes are given by a component-level oracle and $\lambda = 3$ is fixed, the dictionary compression problem $\Pi_{\text{BB-SAM}}$ (in both PS and CPS regimes) is \textbf{strongly NP-hard} and cannot be approximated within a factor of $(1 - o(1))\ln|\mathcal{W}_{\text{target}}|$ in polynomial time unless $\mathrm{P} = \mathrm{NP}$.}

\noindent\emph{Proof.}
We construct a polynomial-time cost-preserving reduction from Minimum Set Cover ($\Pi_{\text{MSC}}$), which is strongly NP-hard and $(1 - o(1))\ln n$-inapproximable. Given an instance of $\Pi_{\text{MSC}}$ with universe $U = \{u_1, \dots, u_n\}$ ($n > \lambda$) and subsets $\mathcal{S} = \{S_1, \dots, S_m\}$ ($\bigcup_{j=1}^{m} S_j = U$), choose an integer spacing $L \gg \max(m, n)$ and construct maps $I_{\text{dict}}$ and $I_{\text{target}}$ in $\mathbb{Z}^2$:
\begin{enumerate}
    \item \emph{Target Map ($I_{\text{target}}$):} Place isolated points $x_r = (rL, 0)$ for each $u_r \in U$, together with a dense, $D_4$-asymmetric anchor pattern $A_0$ of size $M = n + 1$ located near $(-L, 0)$ (sufficiently separated from $x_1$ to prevent accidental BB overlap), forming $\mathcal{W}_{\text{target}} = \{x_1, \dots, x_n\} \cup A_0$ with $|\mathcal{W}_{\text{target}}| = 2n + 1$.
    \item \emph{Dictionary Map ($I_{\text{dict}}$):} For each $S_j \in \mathcal{S}$, place $Q_j = \{(rL, jL) \mid u_r \in S_j\}$ along with a shifted copy of $A_0$ near $(-L, jL)$, adding dummy padding points inside the interior of $A_0$ so that $|Q_j \cup A_0| = C_0$ is uniform across all $j$, yielding $\mathcal{W}_{\text{dict}} = \bigcup_{j=1}^{m} (Q_j \cup A_0)$ where the component-level oracle returns the grouped candidate components $\{Q_j \cup A_0\}_{j=1}^{m}$. (For the PS regime where $I_{\text{dict}} = I_{\text{target}}$, place $\mathcal{W}_{\text{dict}}$ and $\mathcal{W}_{\text{target}}$ in the same grid separated by an offset $L^2$.)
    \item \emph{Parameters:} Set $\tau_{\text{CD}} = 0$, $\tau_{\text{in}} = 1$, and $\rho_{\min} = 1$.
\end{enumerate}
This embedding takes polynomial time and grid dimensions in $n$ and $m$. Because $\tau_{\text{CD}} = 0$, $\tau_{\text{in}} = 1$, and $A_0$ is dense and strictly asymmetric under all non-identity elements of $D_4$, covering any $x_r \in \mathcal{W}_{\text{target}}$ requires selecting $P_j = (Q_j \cup A_0) - (0, jL)$ with $u_r \in S_j$ into $\mathcal{D}$ and placing it via the pure translation $T_j: (x,y) \mapsto (x, y - jL)$. In CPS, every selected part $P_j$ incurs a uniform dictionary cost $C_0$ and placement cost $\lambda = 3$, so minimizing $C_{\text{total}} = (C_0 + \lambda)|J|$ is strictly equivalent to minimizing $|J|$ subject to $\bigcup_{j \in J} S_j = U$. (In PS, covering $\mathcal{W}_{\text{dict}}$ itself forces all $m$ parts to be included in $\mathcal{D}$ at a fixed dictionary overhead $m C_0$ and placement cost $\lambda m$; because these $m$ primitives are then already resident in $\mathcal{D}$ at zero marginal dictionary cost, covering $\mathcal{W}_{\text{target}}$ minimizes $\lambda |J|$, which remains strictly equivalent to minimizing $|J|$.) Since $\ln|\mathcal{W}_{\text{target}}| = \ln(2n+1) = \ln n + O(1)$, we obtain $\Pi_{\text{MSC}} \le_P \Pi_{\text{BB-SAM}}$, establishing strong NP-hardness and $(1 - o(1))\ln|\mathcal{W}_{\text{target}}|$ inapproximability. $\blacksquare$

\noindent\textbf{Remark (Bilevel Coupling and Neural Oracle Query Bottleneck).}
In practice, $\Pi_{\text{BB-SAM}}$ exhibits an even sharper computational barrier than single-level symbolic set cover because the part geometry $P_{k} = \Phi_{\text{SAM}}(I_{\text{dict}}, B_{k}^{\text{dict}})$ in Eq.~\eqref{eq:opt_sam} is generated by a neural operator conditioned on the upper-level BB prompt. This hybrid coupling induces a \emph{bilevel optimization} structure reminiscent of $\Sigma_2^P$-hard bilevel combinatorial problems. Moreover, evaluating $\Phi_{\text{SAM}}$ across all $\Theta(H^2 W^2)$ rectangular BBs in an $H \times W$ map requires $\Theta(H^2 W^2)$ forward passes, which is prohibitive for real-time robotics. BB-SAM provides a principled polynomial-time neuro-symbolic approximation: vectorized ZNCC region growing prunes the $\Theta(H^2 W^2)$ hypothesis space by several orders of magnitude to a compact candidate set, batched SAM 2.1 extracts high-purity primitives in a small constant number of batched forward passes, and transitive consolidation greedily approximates the symbolic set cover.
% ==============================================================================
% Section 2: Experimental Evaluation and Future Extensions
% ==============================================================================
\section{Experimental Evaluation and Future Extensions}
\label{sec:supp_bbsam_eval}

To empirically validate the theoretical formulation in Section~\ref{sec:supp_bbsam_theory}, we evaluate BB-SAM on a benchmark suite of $8$ real-world indoor SLAM occupancy grid maps (\texttt{orebro}, \texttt{edmonton}, \texttt{aces\_publicb}, \texttt{fr079-complete}, \texttt{fr-campus}, \texttt{csail}, \texttt{mexico}, and \texttt{belgioioso}) sourced from standard robotics repositories, including the Radish dataset~\cite{15}. Our experiments are designed to answer three key research questions:
\begin{itemize}
    \item[\textbf{RQ1:}] \emph{(Cross-Building Generalization)} Can a dictionary extracted from one building compress an unseen building (\textbf{CPS}, $I_{\text{dict}} \neq I_{\text{target}}$), particularly when intra-map self-similarity is absent (\textbf{PS}, $I_{\text{dict}} = I_{\text{target}}$)?
    \item[\textbf{RQ2:}] \emph{(Dictionary Design Principle)} What structural properties govern whether a reference map $I_{\text{dict}}$ serves as a universally transferable dictionary across diverse environments?
    \item[\textbf{RQ3:}] \emph{(Neuro-Symbolic Purity)} Does BB-prompted SAM filtering isolate clean architectural skeletons without clutter contamination or free-space inflation?
\end{itemize}

\begin{figure*}[t]
  \centering
  \includegraphics[width=0.78\textwidth]{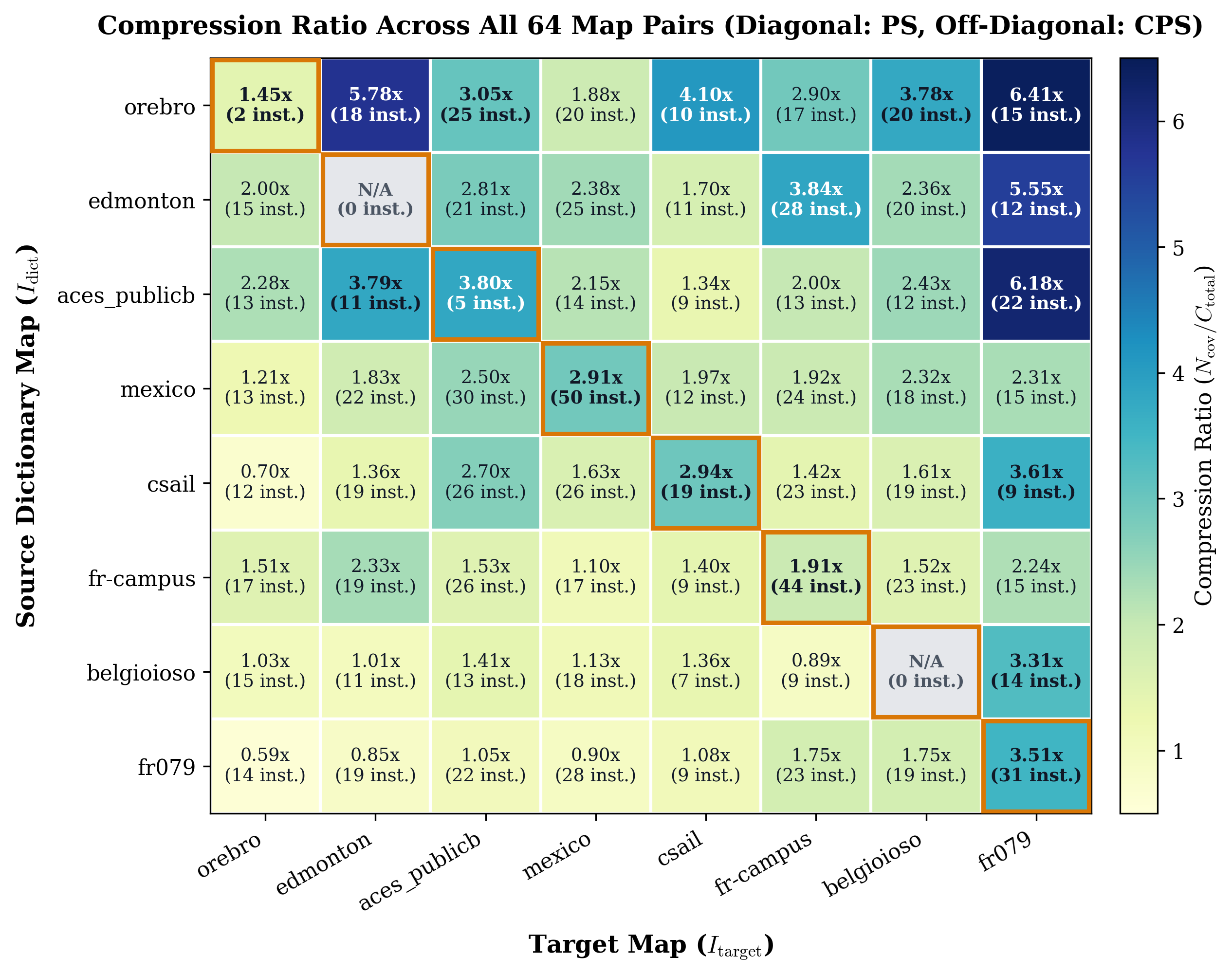}
  \caption{Comprehensive structural compression ratio ($N_{\text{cov}} / C_{\text{total}}$) and matched instance count matrix across all $8 \times 8 = 64$ map combinations (rows: source dictionary $I_{\text{dict}}$; columns: target map $I_{\text{target}}$). Diagonal cells (orange borders) denote Single-Map Part Compression (PS), whereas off-diagonal cells denote Cross-Map Part Compression (CPS). Gray cells (\texttt{N/A}) indicate zero valid self-repeating clusters under PS (\texttt{edmonton} and \texttt{belgioioso}), which are effectively compressed up to $5.78\times$ and $3.78\times$ via CPS using external dictionaries such as \texttt{orebro}.}
  \label{fig:cps_ps_heatmap}
\end{figure*}

\paragraph{Evaluation Protocol and Metrics.}
We evaluate both operational regimes across all $8 \times 8 = 64$ ordered map pairs: $8$ diagonal \textbf{Single-Map Part Compression (PS)} tasks ($I_{\text{dict}} = I_{\text{target}}$) and $56$ off-diagonal \textbf{Cross-Map Part Compression (CPS)} tasks ($I_{\text{dict}} \neq I_{\text{target}}$). Consistent with the MDL objective in Eq.~\eqref{eq:opt_obj} ($\lambda = 3$), the description cost of the modeled structure is $C_{\text{total}} = \sum_{k=1}^{K} |P_k| + 3 N_{\text{inst}}$, where $\sum_{k=1}^{K} |P_k|$ counts the occupied cells stored in the $K$-part dictionary and $3 N_{\text{inst}}$ accounts for the discrete pose parameters $(x_i, y_i, g_i)$ across $N_{\text{inst}}$ matched instances. Let $N_{\text{cov}} = \bigl|\bigcup_{i=1}^{N_{\text{inst}}} (T_i(P_{k_i}) \cap \mathcal{W}_{\text{target}})\bigr|$ denote the number of target occupied cells explained by the dictionary. Assuming uncovered residual cells $\mathcal{W}_{\text{target}} \setminus \mathcal{W}_{\text{cov}}$ are stored verbatim in a hybrid representation, we quantify efficiency via the \textbf{structural compression ratio} ($N_{\text{cov}} / C_{\text{total}}$) and \textbf{net saved cells} ($N_{\text{cov}} - C_{\text{total}}$). Figure~\ref{fig:cps_ps_heatmap} presents the complete $8 \times 8$ matrix across all $64$ pairs, where rows index source dictionaries $I_{\text{dict}}$ and columns index target maps $I_{\text{target}}$.

\subsection{Cross-Map Compression (CPS) vs.\ Single-Map Compression (PS) [RQ1]}
A central finding of our evaluation is that cross-building structural transfer (CPS) achieves positive net compression ($N_{\text{cov}} / C_{\text{total}} > 1.0\times$) in \textbf{52 out of 56 off-diagonal pairs (92.9\%)}, while successfully compressing irregular environments that exhibit zero internal self-similarity.

Specifically, CPS overcomes the \emph{intra-map structural scarcity bottleneck} inherent to Single-Map Part Compression (PS) by leveraging a shared \emph{cross-building architectural alphabet}:
\begin{itemize}
    \item \textbf{Rescuing Non-Repetitive Environments:} While diagonal PS baselines achieve $1.45\times$--$3.80\times$ compression on $6$ environments with internal symmetry, \texttt{belgioioso} ($|\mathcal{W}_{\text{target}}| = 21,919\text{ pts}$) and \texttt{edmonton} ($|\mathcal{W}_{\text{target}}| = 24,136\text{ pts}$) appear as gray \texttt{N/A (0 inst.)} entries along the diagonal of Fig.~\ref{fig:cps_ps_heatmap}, yielding zero valid self-repeating clusters under PS due to their asymmetric layouts (Fig.~\ref{fig:cps_qualitative}(a-1)). Counter-intuitively, rather than requiring self-repetition within the target map, transferring a compact external dictionary $\mathcal{D}$ from \texttt{orebro} rescues both environments, compressing \texttt{belgioioso} at \textbf{$3.78\times$} ($20$ instances, $+2,282\text{ pts}$ net saved) and \texttt{edmonton} at \textbf{$5.78\times$} ($18$ instances, $+2,751\text{ pts}$ net saved; Fig.~\ref{fig:cps_qualitative}(a-2)).
    \item \textbf{Peak Cross-Map Compression Pairs:} Table~\ref{tab:appendix_cps_top} and the dark-blue cells in Fig.~\ref{fig:cps_ps_heatmap} highlight the highest-performing transfers. Transferring $K = 4$ compact parts ($716\text{ base pts}$) from \texttt{orebro} to \texttt{fr079-complete} reconstructs $N_{\text{cov}} = 4,878\text{ wall pts}$ ($28.6\%$ of $|\mathcal{W}_{\text{target}}|$) across $N_{\text{inst}} = 15$ instances at a total cost of $C_{\text{total}} = 761\text{ pts}$, achieving a peak \textbf{$6.41\times$ compression ratio} ($+4,117\text{ pts}$ saved; Fig.~\ref{fig:cps_qualitative}(b)). Similarly, transferring $K = 5$ parts ($929\text{ base pts}$) from \texttt{aces\_publicb} to \texttt{fr079-complete} covers $6,154\text{ pts}$ ($36.0\%$) across $22$ instances at \textbf{$6.18\times$ compression} ($+5,159\text{ pts}$ saved).
\end{itemize}

\subsection{The Compression--Transferability Trade-Off: Primitive Granularity Principle [RQ2]}
Comparing horizontal rows across Fig.~\ref{fig:cps_ps_heatmap} alongside the aggregated cross-target statistics in Table~\ref{tab:appendix_dict_suitability} reveals a fundamental \emph{compression--transferability trade-off} governed by primitive granularity when selecting a universal reference map $I_{\text{dict}}$:
\begin{itemize}
    \item \textbf{Elementary Modular Dictionaries Maximize Transferability (\texttt{orebro}, \texttt{edmonton}, \texttt{aces\_publicb}):} The top three rows of Fig.~\ref{fig:cps_ps_heatmap} consistently achieve net compression across all $7$ unseen targets ($100\%$ win rate $>1.0\times$). In particular, \texttt{orebro} attains the highest cross-map mean compression ratio of \textbf{$3.01\times$} ($1.88\times$--$6.41\times$, $+14,993\text{ pts}$ cumulative net saving), followed closely by \texttt{edmonton} (\textbf{$2.91\times$}) and \texttt{aces\_publicb} (\textbf{$2.88\times$}). Their extracted primitives have compact footprints ($\approx 150$--$230\text{ pts}$ per part) that capture elementary architectural motifs---straight corridor segments, L-corners, and wall alcoves---that recur frequently across disparate buildings.
    \item \textbf{Over-Specialized Compound Dictionaries Fail to Generalize (\texttt{fr079-complete}):} Conversely, the bottom row of Fig.~\ref{fig:cps_ps_heatmap} shows that selecting \texttt{fr079-complete} as $I_{\text{dict}}$ yields the lowest mean compression ratio (\textbf{$1.14\times$}), incurring net storage expansion ($< 1.0\times$, pale-yellow cells) on $3$ unseen targets: \texttt{orebro} ($0.59\times$), \texttt{edmonton} ($0.85\times$), and \texttt{mexico} ($0.90\times$). Because \texttt{fr079-complete} extracts large, site-specific multi-room assemblies ($\approx 270$--$500\text{ pts}$ per part), fewer target instances satisfy the geometric consistency constraints in Eqs.~\eqref{eq:opt_geom_cd}--\eqref{eq:opt_geom_in}, leaving high dictionary base overhead unamortized.
\end{itemize}
This finding establishes an architectural design principle directly analogous to subword tokenization in natural language processing and compositional priors in MDL inference: \emph{compact, elementary geometric primitives act as a universal structural alphabet that maximizes cross-environment reusability, whereas overly large, site-specific compound assemblies suffer from out-of-vocabulary generalization failure}.

\subsection{Qualitative Assessment of Extracted Primitives [RQ3]}
Figure~\ref{fig:cps_qualitative} illustrates qualitative reconstructions across representative map pairs. In Fig.~\ref{fig:cps_qualitative}(a-1), candidate region-growing boxes on \texttt{edmonton} fail to form valid self-repeating clusters under PS due to global layout asymmetry, whereas transferring just $3$ elementary wall primitives (red) from \texttt{orebro} in Fig.~\ref{fig:cps_qualitative}(a-2) accurately matches $18$ locations (blue) across its irregular corridors ($5.78\times$ compression). Similarly, Fig.~\ref{fig:cps_qualitative}(b) shows that $4$ compact primitives extracted along the main corridor of \texttt{orebro} generalize across $15$ rooms and corridor junctions in \texttt{fr079-complete} ($6.41\times$ compression), while Fig.~\ref{fig:cps_qualitative}(c) highlights large-scale cross-building transfer from \texttt{edmonton} ($9$ dictionary parts matching $28$ instances) onto \texttt{fr-campus} ($3.84\times$ compression).

Across all cases in Fig.~\ref{fig:cps_qualitative}, the dictionaries $\mathcal{D}$ extracted by BB-SAM isolate diverse architectural motifs cleanly without bounding-box clutter:
\begin{itemize}
    \item \textbf{Oblique and Curved Walls:} Non-Manhattan diagonal exterior walls and curved building wings (notably in \texttt{fr-campus}, Fig.~\ref{fig:cps_qualitative}(c)) are accurately matched across discrete rotational and reflective symmetries ($D_4$).
    \item \textbf{T-Junctions and Room Corners:} Protruding T-junctions and multi-room corners (seen in \texttt{fr079-complete}, Fig.~\ref{fig:cps_qualitative}(b)) conform tightly to physical wall boundaries without free-space inflation, demonstrating the effectiveness of the $6\text{ px}$ border-margin seed guard and the $2\text{ px}$ empty-space over-expansion guard introduced in Section~\ref{sec:supp_bbsam_theory}.
    \item \textbf{Isolated Rooms and Corridor Modules:} Detached small rooms and repeating corridor wall segments (Fig.~\ref{fig:cps_qualitative}(a-2)--(b)) are distilled into unified symbolic primitives and transferred seamlessly across buildings.
\end{itemize}

\subsection{Future Extensions}
By distilling noisy occupancy grids into a high-purity symbolic dictionary $\mathcal{D}$, BB-SAM directly supplies the compact reference vocabulary required by our fast online descriptor matcher (iMM/hMM), bridging zero-shot neuro-symbolic primitive emergence with real-time global localization. Promising directions for future work include:
\begin{enumerate}
    \item \textbf{Multi-Mask Compositional Unions within BBs:} Extending $\Phi_{\text{SAM}}$ to compose multiple disjoint wall masks within a single bounding box will enable unified encoding of parallel corridor walls and disconnected room boundaries, substantially increasing total target map coverage $N_{\text{cov}} / |\mathcal{W}_{\text{target}}|$.
    \item \textbf{Online Incremental Dictionary Adaptation for Compressive SLAM:} Updating $\mathcal{D}$ incrementally as a robot explores novel architectural regimes (e.g., transitioning from office corridors to open atriums) via online MDL model selection will support lifelong compressive SLAM~\cite{22}.
    \item \textbf{3D Volumetric and Open-Vocabulary Extensions:} Generalizing BB-SAM from 2D occupancy grids and $D_4$ symmetries to 3D voxel/point-cloud maps under $SE(3)$ or discrete cubic symmetries, coupled with open-vocabulary 3D foundation models, will enable semantic-geometric scene compression in multi-floor environments.
\end{enumerate}
\setcounter{section}{0}
% \setcounter{figure}{0}
% \setcounter{table}{0}
% \setcounter{equation}{0}
% \setcounter{page}{1} % ページ番号も1からに戻す場合（不要なら削除）

% --- 3. ★ 2回目の \maketitle を許可するための内部フラグ調整 ---
\if@twocolumn
  \twocolumn
\fi
\makeatother

\vspace*{1ex}
{\centering
  {\Large \bfseries 
Supplementary Material B: A Foundation Model Map-FM for Semantic-Free, Variable-Size, Transferable Latent Map Representation
\par}
  \vspace{1ex}
  \vspace{3ex}
}

\section{Overview and Motivation}
\label{app:mapfm_overview}

As discussed in the conclusion of the main paper (Section~V), our succinct scene modeling framework generalizes beyond standard 2D point-cloud maps to alternative representations such as view-sequence maps. In the original PartSLAM formulation, an input map is explicitly decomposed in physical space into a small set of large geometric parts via discrete dictionary lookup and axis-aligned bounding boxes (BBs).

In this appendix, we introduce \textbf{Map-FM}, a foundation model for maps that elevates this compositional principle into a continuous latent space. Rather than performing explicit dictionary retrieval and BB cropping in physical coordinates, Map-FM extracts and aggregates \emph{implicit spatio-temporal primitives} directly from raw sensor observation streams. When adapting existing foundation models to view-sequence maps acquired by mobile robots, conventional vision foundation models (e.g., DINOv2) and 3D point-cloud foundation models (e.g., Utonia) suffer from four fundamental structural mismatches:
\begin{enumerate}
    \item \textbf{Non-Semantic Numerical Waveforms:} Mobile robot range sequences consist of continuous, label-free geometric measurements (e.g., LiDAR scan distances) lacking visual textures or object categories. Consequently, the semantic inductive biases of vision and 3D foundation models are poorly aligned with raw range profiles.
    \item \textbf{Dynamic Variable-Length Trajectories:} Unlike vision backbones that require resizing or cropping inputs to a fixed spatial grid (e.g., $224 \times 224$), observation sequences vary dynamically in length $T$ depending on the robot's travel distance along a trajectory segment.
    \item \textbf{Anisotropic Spatio-Temporal Axes:} Unlike isotropic 2D image grids, multi-beam range logs are structured along two physically heterogeneous axes: a fixed angular axis across sensor beams ($B$) and a variable-length spatial-sequential axis induced by robot traversal ($T$).
    \item \textbf{Trajectory-Level Sequential Context:} Whereas static image encoders capture instantaneous local layouts, robust place recognition requires encoding global sequential context---such as continuous corridor transitions, corner appearances, and structural periodicities accumulated along a trajectory.
\end{enumerate}

Our key insight is that raw multi-beam range logs share a deep structural affinity with multivariate time-series waveforms rather than natural images. Accordingly, Map-FM adapts \textbf{TimesFM}---a time-series foundation model pretrained on vast, domain-agnostic numerical fluctuation patterns independent of semantic labels---via Low-Rank Adaptation (LoRA). By coupling frozen temporal priors with lightweight spatial aggregation, Map-FM maps variable-length, non-semantic observation sequences into compact, fixed-length ($256$-dimensional) unit descriptors optimized for rapid 1-to-$N$ place matching.

% ------------------------------------------------------------------------------
\section{Model Architecture}
\label{app:mapfm_arch}

Let $\mathbf{X} \in \mathbb{R}^{B \times T}$ denote a variable-length multi-beam observation sequence sampled over $T \ge 1$ trajectory steps across a standardized angular grid of $B$ spatial beams. To decouple long-horizon sequential modeling from cross-beam geometric aggregation, Map-FM adopts a factorized spatio-temporal architecture consisting of temporal patch tokenization, a LoRA-adapted TimesFM backbone, and task-specific pooling and projection heads.

\subsection{Temporal Patchification and Validity Masking}
Along the sequential axis, each beam trajectory $\mathbf{x}_b \in \mathbb{R}^{T}$ ($b \in \{1, \dots, B\}$) is partitioned into $N_p = \lceil T / L_p \rceil$ non-overlapping patches of fixed length $L_p$ via right-hand zero-padding, denoted as $\operatorname{Patchify}(\mathbf{x}_b, L_p)$. Because zero values can coincide with valid normalized range readings, each beam is accompanied by an element-wise binary validity mask $\mathbf{M}_b \in \{0, 1\}^{N_p \times L_p}$ (indicating valid trajectory steps versus padded entries) and a patch-level validity mask $\mathbf{m}_b \in \{0, 1\}^{N_p}$ (where $m_{b,p} = 1$ if patch $p$ contains at least one valid step, and $0$ otherwise), yielding the patch tensor $\mathbf{P}_b \in \mathbb{R}^{N_p \times L_p}$.

\subsection{LoRA-Adapted Pretrained TimesFM Backbone}
Sequential feature extraction is applied independently to each beam $b \in \{1, \dots, B\}$ using the pretrained TimesFM backbone adapted with LoRA, denoted as $\operatorname{TimesFM}_{\text{LoRA}}(\cdot)$:
\begin{equation}
    \mathbf{H}_b = \operatorname{LayerNorm}\Bigl( \operatorname{TimesFM}_{\text{LoRA}}\bigl([\mathbf{P}_b \,\Vert{}\, \mathbf{M}_b], \, \mathbf{m}_b\bigr) \Bigr) \in \mathbb{R}^{N_p \times D},
\end{equation}
where $[\cdot \,\Vert{}\, \cdot] \in \mathbb{R}^{N_p \times 2L_p}$ denotes feature-wise concatenation of the numerical patch and its binary validity mask, and $D = 1280$ is the hidden dimension of the backbone. Internally, $\operatorname{TimesFM}_{\text{LoRA}}(\cdot)$ comprises two components:
\begin{itemize}
    \item \textbf{Patch Tokenizer:} A residual MLP input layer that projects the concatenated patch-and-mask representation $[\mathbf{P}_b \,\Vert{}\, \mathbf{M}_b]$ into $D$-dimensional initial token embeddings, explicitly informing the network of within-patch padded boundaries.
    \item \textbf{Stacked Transformer Blocks:} Deep self-attention layers that model long-range sequential dependencies across the $N_p$ temporal tokens while restricting attention strictly to valid patches via $\mathbf{m}_b$. To preserve the general-purpose numerical waveform priors learned during pretraining while adapting to robotics geometry, all original backbone weights remain frozen, and trainable low-rank adapters (LoRA) are injected exclusively into the Query--Key--Value and output projections of each attention layer.
\end{itemize}

\subsection{Task-Specific Aggregation and Projection Heads}
To distill the variable-length spatio-temporal token representations $\{\mathbf{H}_b\}_{b=1}^B$ into a compact, fixed-length place descriptor $\mathbf{z} \in \mathbb{R}^{256}$ across varying sequence durations $T$, we attach three lightweight modules trained end-to-end alongside the LoRA parameters:
\begin{enumerate}
    \item \textbf{Masked Temporal Attention Pooling:}
    For each beam $b$, a two-layer attention network (parameterized by $\mathbf{W}_1 \in \mathbb{R}^{D \times D_a}$ and $\mathbf{W}_2 \in \mathbb{R}^{D_a \times 1}$) assigns data-dependent importance weights across temporal patches, acting as a soft selector of salient geometric events along the trajectory:
\begin{equation}
    \begin{aligned}
        \mathbf{s}_b &= \tanh(\mathbf{H}_b \mathbf{W}_1)\mathbf{W}_2 \in \mathbb{R}^{N_p \times 1}, \\
        \boldsymbol{\alpha}_b &= \operatorname{Softmax}\bigl(\tilde{\mathbf{s}}_b\bigr) \in \mathbb{R}^{N_p \times 1}, \\
        \mathbf{u}_b &= \mathbf{H}_b^\top \boldsymbol{\alpha}_b \in \mathbb{R}^D,
    \end{aligned}
\end{equation}
    where $\tilde{s}_{b,p} = s_{b,p}$ if $m_{b,p} = 1$ and $\tilde{s}_{b,p} = -\infty$ if $m_{b,p} = 0$. Suppressing fully padded patches ($m_{b,p} = 0$) prior to $\operatorname{Softmax}$ normalization guarantees strict invariance of $\mathbf{u}_b$ to additional zero-padded patch blocks, whereas partially padded terminal patches ($m_{b,p} = 1$ with trailing zeros) remain visible to the patch tokenizer so that sub-patch boundary truncation cues are preserved.

    \item \textbf{Spatial Beam Encoder:}
    While TimesFM captures sequential evolution along each individual beam, local geometric structures (e.g., wall corners and corridor widths) manifest as correlations across adjacent sensor angles. We stack the beam embeddings into $\mathbf{U} = [\mathbf{u}_1, \dots, \mathbf{u}_B]^\top \in \mathbb{R}^{B \times D}$ and apply two 1D convolutional layers along the spatial beam axis (treating $D$ as the input channel dimension over the transposed tensor $\mathbf{U}^\top \in \mathbb{R}^{D \times B}$):
    \begin{equation}
        \mathbf{v} = \operatorname{Flatten}\Bigl(\operatorname{Conv1d}_2\bigl(\operatorname{GELU}(\operatorname{Conv1d}_1(\mathbf{U}^\top))\bigr)\Bigr) \in \mathbb{R}^{B \cdot C_{\text{out}}},
    \end{equation}
    where padding preserves the spatial resolution $B$ and $C_{\text{out}}$ denotes the output channel width of $\operatorname{Conv1d}_2$.

    \item \textbf{Compact Descriptor Projection Head:}
    Finally, the flattened spatio-temporal feature vector $\mathbf{v}$ is mapped onto the $256$-dimensional unit hypersphere via a two-layer MLP ($\mathbf{W}_3 \in \mathbb{R}^{D_h \times (B \cdot C_{\text{out}})}$, $\mathbf{W}_4 \in \mathbb{R}^{256 \times D_h}$) followed by $L_2$ normalization:
    \begin{equation}
        \tilde{\mathbf{z}} = \mathbf{W}_4 \operatorname{GELU}(\mathbf{W}_3 \mathbf{v}), \quad \mathbf{z} = \frac{\tilde{\mathbf{z}}}{\|\tilde{\mathbf{z}}\|_2} \in \mathbb{R}^{256}.
    \end{equation}
    This representation compresses arbitrary-length multi-beam observation sequences into a fixed $1\,\mathrm{KB}$ (single-precision) unit embedding, enabling sub-millisecond 1-to-$N$ cosine-similarity map matching across large-scale environments.
\end{enumerate}
% ==============================================================================
% Appendix B (Part 2): Progressive Refinement of Training and Evaluation
%                      Protocols on Real-World Data
% Note: Requires \usepackage{amsmath, amssymb, booktabs, array} in the preamble.
%       Continues directly after \section{Model Architecture} in Part 1.
% ==============================================================================

\section{Principled Benchmark Design: Resolving Domain Gaps in Real-World Logs}
\label{app:mapfm_variants}

Adapting a domain-agnostic sequential foundation model to physical robot logs introduces three fundamental domain gaps: (i)~\emph{metric scale distortion} induced by naive sequence resizing, (ii)~\emph{angular field-of-view (FOV) heterogeneity} across different LiDAR hardware, and (iii)~\emph{boundary aliasing} when evaluating discrete segments sampled from continuous trajectories. 

To systematically isolate and quantify the impact of each factor on real-world mobile robot data, we construct a progressive series of three training and evaluation protocols (\textbf{Variant~1}, \textbf{Variant~2}, and \textbf{Variant~3}) on the public RADISH dataset~\cite{15}. Crucially, all three variants share the exact same Map-FM architecture and contrastive training objective, ensuring that performance differences directly reflect the physical and geometric fidelity of the data representation.

% ------------------------------------------------------------------------------
\subsection{Invariant Model and Training Configuration}

Across all three protocol variants, the underlying model architecture and optimization setup are held strictly constant:
\begin{itemize}
    \item \textbf{Shared Network Architecture:} All variants employ the frozen \texttt{TimesFM-2.5-200M} backbone ($D = 1280$, patch length $L_p = 32$) equipped with trainable LoRA adapters ($r = 8, \alpha = 16$) injected exclusively into the self-attention projection layers. Following Section~\ref{app:mapfm_arch}, a mini-batch tensor of shape $(\text{Batch}, B, T)$ with $B = 64$ beams is reshaped into $\text{Batch} \times B$ independent 1D sequences, padded and tokenized with binary validity masks, and processed by the backbone. The resulting patch embeddings are aggregated via masked temporal attention pooling, a 2-layer 1D spatial convolution across the beam axis, and an $L_2$-normalized 2-layer MLP head to output a $256$-dimensional unit-norm place descriptor $\mathbf{z} \in \mathbb{S}^{255}$.
    \item \textbf{Zero-Shot Cross-Building Split:} To evaluate cross-building generalization without spatial leakage, trajectories from three unseen indoor environments (\texttt{intel}, \texttt{fr079-complete}, and \texttt{csail}) are strictly held out for testing, while logs from the remaining eight buildings form the training split. Raw odometry-corrected scans are resampled at uniform spatial intervals of $\Delta s = 0.10\,\text{m}$ per step. Candidate revisit pairs (database vs.\ query) are mined from base trajectory windows of $100$ steps ($10.0\,\text{m}$), enforcing a minimum along-track separation to exclude trivial same-pass overlaps.
    \item \textbf{Contrastive Objective and Perturbation:} Models are trained with the symmetric InfoNCE loss (temperature $\tau = 0.10$), augmented with random cyclic/edge angular beam shifts ($\pm 1$ beam, corresponding to $\approx \pm 2.8^\circ$ on a $180^\circ$ grid) to simulate minor heading noise.
\end{itemize}

% ------------------------------------------------------------------------------
\subsection{Overview of the Progressive Protocol Variants}

Table~\ref{tab:variant_comparison} contrasts the three protocols across four design dimensions, tracing the progression from a naive sequence-resampling baseline to a physically and geometrically consistent benchmark.

\begin{table*}[t]
\centering
\caption{Controlled Evolution of Data Preprocessing, Sequence Sampling, and Evaluation Protocols Across Variants}
\label{tab:variant_comparison}
\renewcommand{\arraystretch}{1.28}
\begin{tabular}{p{3.3cm} p{4.2cm} p{4.3cm} p{4.5cm}}
\toprule
\textbf{Design Dimension} & \textbf{Variant 1} \newline \textit{(Sequence Resampling)} & \textbf{Variant 2} \newline \textit{(Scale-Preserving Subsequence)} & \textbf{Variant 3 (Proposed)} \newline \textit{(FOV-Aligned + Spatial Eval)} \\
\midrule
\textbf{Sensor FOV Harmonization} & Naive beam-count resampling to $B=64$ (ignores physical FOV angles). & Same as Variant~1 (unaligned beam angles across sensors). & \textbf{Angular interpolation onto a canonical frontal $180^\circ$ grid $[-\pi/2, +\pi/2]$}; out-of-FOV beams padded with max range ($1.0$). \\
\midrule
\textbf{Variable-Length Modeling} & Linear resampling of identical physical segments to random step counts $T_{\text{db}}, T_{\text{q}}$ (distorts metric scale). & \textbf{Scale-preserving random cropping} ($L_{\text{db}} \in [6.4, 10.0]\,\text{m}$, $L_{\text{q}} \in [3.2, L_{\text{db}}]\,\text{m}$) at fixed $0.10\,\text{m}$/step. & Same as Variant~2 (strict metric scale preservation at $0.10\,\text{m}$/step). \\
\midrule
\textbf{Revisit Mining \& Class Balance} & Distance $\le 1.5\,\text{m}$, heading $\cos\Delta\theta \ge 0.7$, uncapped pairs per building. & Distance $\le 1.0\,\text{m}$, $\cos\Delta\theta \ge 0.8$, balanced sampling capped at $70$--$80$ pairs/building. & Distance $\le 1.0\,\text{m}$, $\cos\Delta\theta \ge 0.85$, stride $25$ steps ($2.5\,\text{m}$) with overlap suppression, capped at $70$ pairs/building. \\
\midrule
\textbf{Query Cropping \& Retrieval Metrics} & Prefix cropping from segment start; 1-to-1 Direct Matching via \textbf{Exact Recall@1}. & \textbf{Center-aligned cropping}; Direct and \textbf{Sub-Window Sliding Matching} (stride $1.0\,\text{m}$) via Exact Recall@1. & Center-aligned cropping; Direct Matching evaluated with both \textbf{Exact Recall@1} and \textbf{Spatial Recall@1 / @5} ($\le 3.0\,\text{m}$). \\
\bottomrule
\end{tabular}
\end{table*}

% ------------------------------------------------------------------------------
\subsection{Analytical Rationale Behind Each Refinement}

\subsubsection{Variant 1: Naive Sequence Resampling Baseline}
Variant~1 treats multi-beam LiDAR logs purely as abstract multivariate sequences, simulating variable durations during training by linearly stretching or compressing identical physical segments into random step counts $T_{\text{db}} \in [64, 127]$ and $T_{\text{q}} \in [32, 127]$. However, because the trajectories are already discretized at a uniform spatial step ($\Delta s = 0.10\,\text{m}$), subsequent linear resampling warps the underlying Euclidean geometry---for example, compressing a $10.0\,\text{m}$ corridor into $32$ steps artificially shrinks wall distances and doorway spacing by over $3\times$. This scale distortion prevents the pretrained TimesFM backbone from exploiting consistent metric periodicity. Furthermore, Variant~1 suffers from severe building-level sample imbalance (dominated by long logs such as \texttt{MIT\_Infinite\_Corridor}) and systematic positional bias at test time caused by cropping short queries from the start (prefix) of a database segment rather than its center.

\subsubsection{Variant 2: Metric-Scale Preservation and Subsequence Matching}
Variant~2 eliminates sequence warping to enforce a strict $0.10\,\text{m}$/step physical scale invariant to trajectory length. During training, a database segment of physical length $L_{\text{db}} \in [64, 100]$ steps ($6.4$--$10.0\,\text{m}$) is sampled, and a shorter query segment of length $L_{\text{q}} \in [32, L_{\text{db}}]$ steps ($3.2\,\text{m}$ to $L_{\text{db}}$) is randomly cropped strictly from the interior of that database interval. This formulates training as genuine \emph{subsequence place recognition}, teaching the model to embed partial traversals into the same latent neighborhood as their enclosing reference segments. At test time, queries are cropped symmetrically around the segment centroid, and we evaluate both \emph{Direct Arbitrary-Length Matching} and hierarchical \emph{Sub-Window Sliding Matching} (sliding a window of length $L_{\text{q}}$ across each $10.0\,\text{m}$ database segment at a $1.0\,\text{m}$ stride). In addition, revisit thresholds are tightened ($1.0\,\text{m}$ radius, $\cos\Delta\theta \ge 0.8$) and capped per building to prevent overfitting to large environments.

\subsubsection{Variant 3: Cross-Sensor FOV Alignment and Spatial Tolerance Evaluation}
The RADISH repository aggregates logs recorded with heterogeneous 2D LiDAR sensors whose fields of view vary substantially, from standard $180^\circ$ scanners (\texttt{FLASER}) to wide-angle $240^\circ$ and $270^\circ$ scanners (\texttt{ROBOTLASER1}). In Variants~1 and~2, resizing raw scan arrays solely by beam count caused identical beam indices $b \in \{1, \dots, 64\}$ to observe completely different physical bearing angles across buildings (e.g., lateral walls in a $180^\circ$ sensor vs.\ rear-diagonal walls in a $270^\circ$ sensor). Variant~3 resolves this geometric mismatch by interpolating all scans onto a canonical frontal $180^\circ$ polar grid ($[-\pi/2, +\pi/2]$ across $B=64$ beams) and padding out-of-FOV angles with the normalized maximum range ($1.0$), establishing consistent angular semantics across unseen buildings and sensor platforms. 

Finally, we refine the evaluation criterion to account for continuous trajectory geometry. While we retain index-based \textbf{Exact Recall@1} for direct comparison across variants, penalizing adjacent database predictions under Exact Recall@1 introduces artificial false negatives because database segments sampled with a $2.5\,\text{m}$ stride share $75\%$ ($7.5\,\text{m}$) spatial overlap. Variant~3 therefore incorporates standard VPR spatial tolerance metrics---\textbf{Spatial Recall@1} and \textbf{Spatial Recall@5}---counting a retrieval as correct if the physical centroid of the retrieved database segment lies within $3.0\,\text{m}$ of the query segment's ground-truth centroid (with disjoint coordinate offsets applied across test buildings to prevent cross-building collisions).

Taken together, Variants~1--3 provide a controlled ablation progression: Variant~1 isolates the failure mode of scale distortion, Variant~2 establishes metric-consistent subsequence learning, and Variant~3 yields a sensor-agnostic, spatially grounded benchmark for real-world place recognition.
% ==============================================================================
% Appendix B (Part 3): Experimental Evaluation on Real-World RADISH Dataset
% Note: Requires \usepackage{amsmath, amssymb, booktabs, multirow, graphicx}
%       in the preamble. Continues directly after Part 2.
% ==============================================================================

\section{Experimental Evaluation on Real-World RADISH Dataset}
\label{app:mapfm_exp}

\subsection{Experimental Setup}
To empirically validate the cross-building generalization and variable-horizon robustness of Map-FM in real-world indoor environments, we conduct LiDAR-based place recognition experiments (analogous to visual place recognition, VPR) using real 2D range logs from the public \textbf{RADISH (Robotics Data Set Repository)} benchmark~\cite{15}.

\begin{itemize}
    \item \textbf{Zero-Shot Cross-Building Split:} Eight distinct indoor environments (including \texttt{MIT\_Infinite\_Corridor}, \texttt{fr-campus}, and \texttt{fhw-rec-001}) are allocated to the training set, while three completely unseen buildings (\texttt{intel}, \texttt{fr079-complete}, and \texttt{csail}) are strictly held out for testing.
    \item \textbf{Trajectory Preprocessing and Training Objective:} All odometry-corrected trajectories are resampled at a uniform spatial step of $\Delta s = 0.10\,\mathrm{m}$. Reference database (DB) segments are fixed to $100$ steps ($10.0\,\mathrm{m}$), and non-trivial loop-closure revisit pairs are automatically mined. At test time, query trajectories spanning varying travel distances from $3.2\,\mathrm{m}$ to $10.0\,\mathrm{m}$ ($32$ to $100$ steps, with $2.0\,\mathrm{m}$ additionally probed for Variant~1) are matched against full $10.0\,\mathrm{m}$ DB segments to evaluate partial-trajectory retrieval. All models are trained for $35$ epochs using the symmetric InfoNCE loss (temperature $\tau = 0.10$).
    \item \textbf{Evaluated Protocol Variants:} Following the ablation progression in Section~\ref{app:mapfm_variants}, we evaluate three protocols:
    \begin{enumerate}
        \item \textbf{Variant 1 (Sequence Resampling / Unaligned FOV):} Resizes raw scan arrays solely by beam count ($B = 64$) and simulates variable sequence lengths during training by linearly stretching or compressing identical physical segments into random step counts (Train: $376$ pairs, Test: $73$ pairs).
        \item \textbf{Variant 2 (Scale-Preserving Cropping / Unaligned FOV):} Enforces the strict $\Delta s = 0.10\,\mathrm{m}$/step metric scale and trains on genuine subsequence matching by cropping shorter queries ($3.2$--$10.0\,\mathrm{m}$) from the interior of reference DB intervals ($6.4$--$10.0\,\mathrm{m}$) (Train: $216$ pairs, Test: $57$ pairs).
        \item \textbf{Variant 3 (Scale-Preserving Cropping + Unified Sensor FOV):} Resolves field-of-view discrepancies across heterogeneous LiDAR hardware ($180^\circ$ vs.\ $240^\circ/270^\circ$) by interpolating all scans onto a canonical frontal $180^\circ$ polar grid ($[-\pi/2, +\pi/2]$, $B = 64$), while reporting both index-based Exact Recall@1 and spatial tolerance metrics ($\text{Spatial Recall@1}$ and $\text{Recall@5}$ within $\le 3.0\,\mathrm{m}$) to account for continuous sliding-window overlap (Train: $178$ pairs, Test: $36$ pairs).
    \end{enumerate}
\end{itemize}

\subsection{Quantitative Results}
\begin{figure}[htbp]
    \centering
    \includegraphics[width=0.95\linewidth]{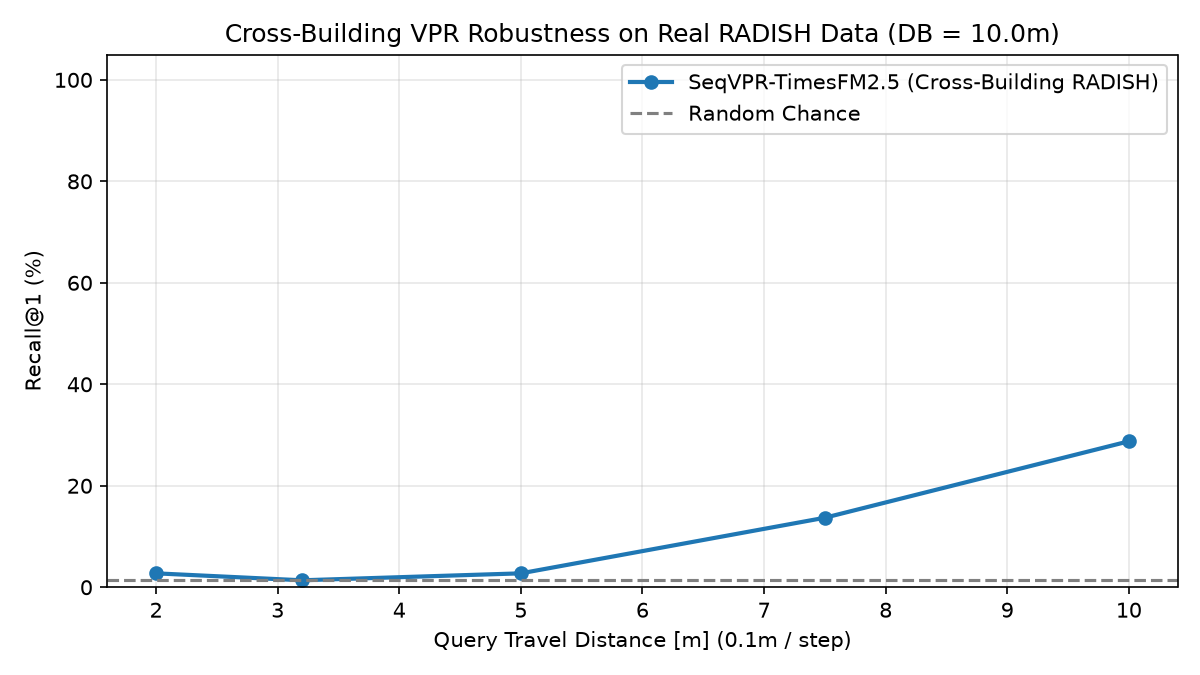}
    \caption{\textbf{Cross-building place recognition under naive sequence resampling (Variant~1).} While reaching $28.77\%$ Exact Recall@1 when the query length matches the $10.0\,\mathrm{m}$ database segment, recall collapses to near-random levels ($1.37\%$--$2.74\%$, vs.\ $1.37\%$ chance) for short queries ($\le 5.0\,\mathrm{m}$) due to physical scale warping.}
    \label{fig:radish_v1}
\end{figure}

\begin{figure}[htbp]
    \centering
    \includegraphics[width=0.95\linewidth]{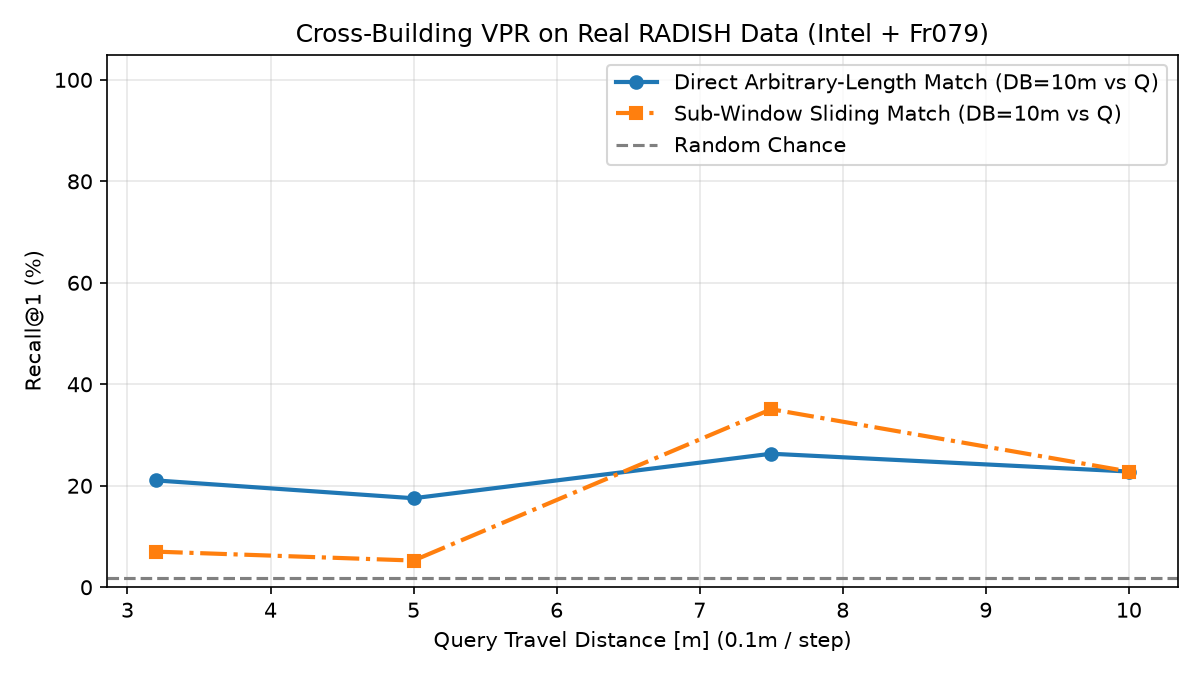}
    \caption{\textbf{Cross-building place recognition under scale-preserving subsequence cropping (Variant~2).} Enforcing a consistent $0.10\,\mathrm{m}$/step spatial scale boosts Direct Exact Recall@1 at $Q=3.2\,\mathrm{m}$ by over $15\times$ ($1.37\% \to 21.05\%$), while sliding sub-window matching reaches $35.09\%$ at $Q=7.5\,\mathrm{m}$.}
    \label{fig:radish_v2}
\end{figure}

\begin{figure}[htbp]
    \centering
    \includegraphics[width=0.95\linewidth]{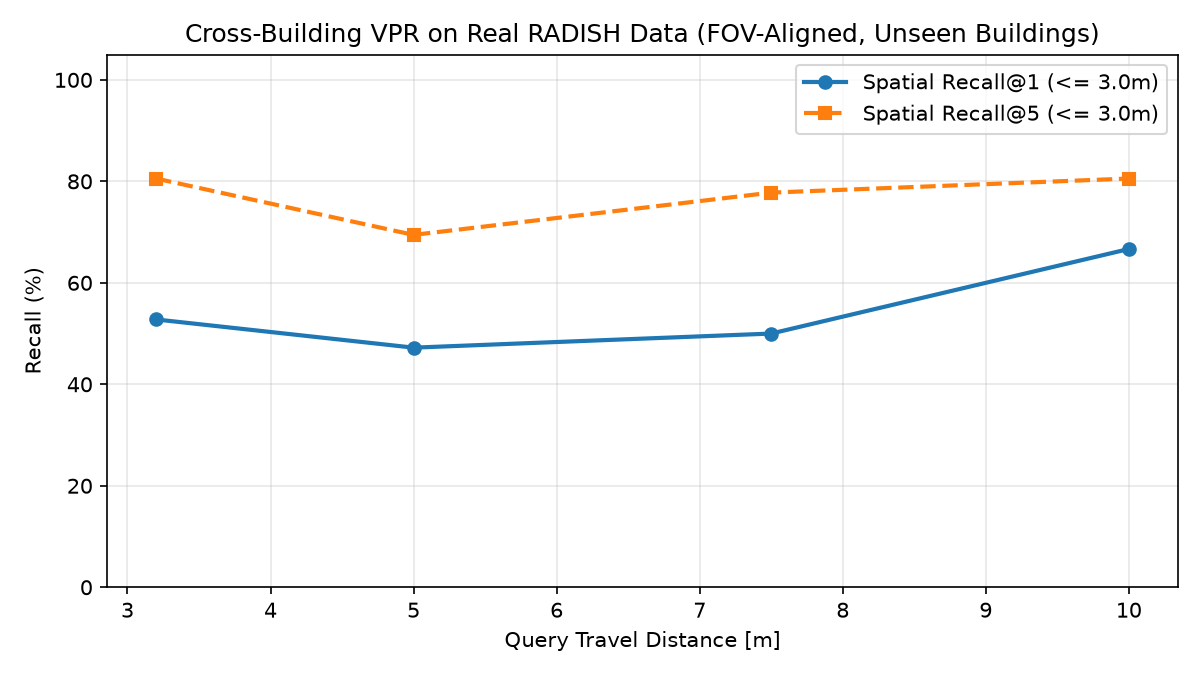}
    \caption{\textbf{Cross-building place recognition of the FOV-aligned model with spatial evaluation (Variant~3).} On unseen buildings (\texttt{intel}, \texttt{fr079-complete}, and \texttt{csail}), Map-FM consistently achieves $69.44\%$--$80.56\%$ Spatial Recall@5 ($\le 3.0\,\mathrm{m}$) across all query horizons from $3.2\,\mathrm{m}$ to $10.0\,\mathrm{m}$, outperforming the $2.78\%$ random baseline by nearly $30\times$.}
    \label{fig:radish_v3}
\end{figure}

Table~\ref{tab:radish_vpr_results} summarizes the place recognition recall on the unseen test environments (\texttt{intel}, \texttt{fr079-complete}, and \texttt{csail}) both prior to fine-tuning (zero-shot) and after $35$ epochs of contrastive adaptation across all three variants. Figures~\ref{fig:radish_v1}--\ref{fig:radish_v3} further illustrate the recall curves as a function of query travel distance $Q$.

\begin{table*}[t]
    \centering
    \caption{Cross-Building Place Recognition Recall (\%) on Unseen RADISH Environments (\texttt{intel}, \texttt{fr079-complete}, \texttt{csail}) Across Varying Query Travel Distances $Q$. The reference database segment length is fixed to $10.0\,\mathrm{m}$ ($100$ steps) across all settings. Boldface indicates post-fine-tuning performance.}
    \label{tab:radish_vpr_results}
    \footnotesize
    \begin{tabular}{llccccc}
        \toprule
        \textbf{Protocol Variant} & \textbf{Metric (Training Stage)} & $Q = 2.0\,\mathrm{m}$ & $Q = 3.2\,\mathrm{m}$ & $Q = 5.0\,\mathrm{m}$ & $Q = 7.5\,\mathrm{m}$ & $Q = 10.0\,\mathrm{m}$ \\
        \midrule
        \multirow{2}{*}{\shortstack[l]{\textbf{Variant 1}\\{\scriptsize (Sequence Stretch / Unaligned FOV)}}}
            & Exact R@1 (Zero-Shot) & 1.37 & 2.74 & 0.00 & --- & 17.81 \\
            & \textbf{Exact R@1 (Fine-Tuned)} & \textbf{2.74} & \textbf{1.37} & \textbf{2.74} & \textbf{13.70} & \textbf{28.77} \\
        \midrule
        \multirow{3}{*}{\shortstack[l]{\textbf{Variant 2}\\{\scriptsize (Scale-Preserving / Unaligned FOV)}}}
            & Direct R@1 (Zero-Shot) & --- & 8.77 & 8.77 & 15.79 & 24.56 \\
            & \textbf{Direct R@1 (Fine-Tuned)} & --- & \textbf{21.05} & \textbf{17.54} & \textbf{26.32} & \textbf{22.81} \\
            & \textbf{Sub-Window R@1 (Fine-Tuned)} & --- & \textbf{7.02} & \textbf{5.26} & \textbf{35.09} & \textbf{22.81} \\
        \midrule
        \multirow{6}{*}{\shortstack[l]{\textbf{Variant 3 (Proposed)}\\{\scriptsize (Scale-Preserving + Unified FOV)}}}
            & Exact R@1 (Zero-Shot) & --- & 27.78 & 2.78 & 19.44 & 41.67 \\
            & \textbf{Exact R@1 (Fine-Tuned)} & --- & \textbf{38.89} & \textbf{19.44} & \textbf{38.89} & \textbf{50.00} \\
            & Spatial R@1 $\le 3.0\,\mathrm{m}$ (Zero-Shot) & --- & 36.11 & 13.89 & 38.89 & 50.00 \\
            & \textbf{Spatial R@1 $\le 3.0\,\mathrm{m}$ (Fine-Tuned)} & --- & \textbf{52.78} & \textbf{47.22} & \textbf{50.00} & \textbf{66.67} \\
            & Spatial R@5 $\le 3.0\,\mathrm{m}$ (Zero-Shot) & --- & 72.22 & 30.56 & 80.56 & 80.56 \\
            & \textbf{Spatial R@5 $\le 3.0\,\mathrm{m}$ (Fine-Tuned)} & --- & \textbf{80.56} & \textbf{69.44} & \textbf{77.78} & \textbf{80.56} \\
        \bottomrule
    \end{tabular}
    \vspace{1.5mm}
    \begin{flushleft}
        {\footnotesize Note: ``---'' denotes query horizons omitted from the respective evaluation sweep. For Variant~1, the zero-shot Exact R@1 at $Q=8.0\,\mathrm{m}$ is $6.85\%$. Random-chance baselines for the three test splits ($N_{\text{test}} = 73, 57, 36$) are $1.37\%$, $1.75\%$, and $2.78\%$, respectively. Sub-window sliding search is omitted in Variant~3 as single-shot Direct Matching already achieves superior recall at a fraction of the computational cost.}
    \end{flushleft}
\end{table*}

\subsection{Analysis and Discussion}
Contrasting the three controlled variants in Table~\ref{tab:radish_vpr_results} and Figs.~\ref{fig:radish_v1}--\ref{fig:radish_v3} yields three foundational insights for adapting sequential foundation models to physical map representations:

\paragraph{Insight 1: Metric-Scale Preservation Unlocks Short-Horizon Robustness (Variant~1 vs.\ Variant~2).}
In Variant~1 (Fig.~\ref{fig:radish_v1}), although the model achieves $28.77\%$ Exact Recall@1 when the query length matches the $10.0\,\mathrm{m}$ reference interval, accuracy collapses to $1.37\%$--$2.74\%$ (indistinguishable from the $1.37\%$ random-chance floor) for queries shorter than $5.0\,\mathrm{m}$. Because Variant~1 linearly stretches or compresses identical physical trajectories into arbitrary step counts during training, Euclidean distances between walls and doorways are warped, preventing the pretrained TimesFM backbone from exploiting true metric periodicity. By contrast, Variant~2 (Fig.~\ref{fig:radish_v2}) preserves the exact $\Delta s = 0.10\,\mathrm{m}$/step physical scale and trains on genuine interior subsequence crops, yielding a \textbf{$15.4\times$ surge in Direct Exact Recall@1 at $Q=3.2\,\mathrm{m}$ (from $1.37\%$ to $21.05\%$)}. Furthermore, hierarchical sub-window sliding matching attains $35.09\%$ at $Q=7.5\,\mathrm{m}$, confirming that masked temporal attention pooling effectively isolates localized sub-trajectories embedded within longer reference segments.

\paragraph{Insight 2: Cross-Sensor FOV Harmonization Drives Zero-Shot Transfer (Variant~2 vs.\ Variant~3).}
In Variants~1 and~2, naive beam-count interpolation misaligns physical bearing angles across heterogeneous LiDAR sensors between the training logs and the unseen test environments (\texttt{intel}, \texttt{fr079-complete}, and \texttt{csail}). When Variant~3 (Fig.~\ref{fig:radish_v3}) projects all range profiles onto a canonical frontal $180^\circ$ polar grid, \textbf{zero-shot Exact Recall@1 at $Q=10.0\,\mathrm{m}$ jumps from $24.56\%$ to $41.67\%$ prior to any fine-tuning} ($15.0\times$ above the $2.78\%$ random baseline, even under stricter overlap suppression). Notably, a localized dip in Exact Recall@1 appears at the intermediate horizon $Q=5.0\,\mathrm{m}$ ($19.44\%$ post-fine-tuning vs.\ $38.89\%$ at $Q=3.2\,\mathrm{m}$ and $7.5\,\mathrm{m}$). This reflects an interplay between sub-patch boundary truncation and perceptual aliasing: $Q=3.2\,\mathrm{m}$ ($32$ steps) aligns cleanly with a single unpadded backbone patch ($L_p = 32$), whereas $Q=5.0\,\mathrm{m}$ ($50$ steps) induces a partially padded second patch ($14$ trailing zero-padded steps, $28\%$ padding ratio) whose boundary mask $\mathbf{M}_b$ alters the patch tokenizer's activation while spanning featureless mid-corridor segments prone to structural ambiguity. Even at this challenging horizon, Spatial Recall@1 remains robust at $47.22\%$. Overall, aligning sensor geometry enables the domain-agnostic sequential priors of pretrained TimesFM to generalize zero-shot to unseen architectural layouts.

\paragraph{Insight 3: Practical Global Candidate Filtering for Succinct Map Matching (Variant~3).}
After contrastive adaptation, Variant~3 (Fig.~\ref{fig:radish_v3}) achieves \textbf{$50.00\%$ Exact Recall@1, $66.67\%$ Spatial Recall@1 ($\le 3.0\,\mathrm{m}$), and $80.56\%$ Spatial Recall@5} at $Q=10.0\,\mathrm{m}$ on completely unseen buildings. Remarkably, even with a brief $3.2\,\mathrm{m}$ ($32$-step) traversal---less than one-third of the database segment length---Map-FM sustains \textbf{$52.78\%$ Spatial Recall@1 and $80.56\%$ Spatial Recall@5}. Consistent with the fast succinct map matching objective of PartSLAM, each variable-length trajectory is distilled into a fixed $256$-dimensional unit embedding $\mathbf{z} \in \mathbb{S}^{255}$ ($1\,\mathrm{KB}$ in single precision), enabling sub-millisecond $1$-to-$N$ cosine-similarity retrieval that captures over $80\%$ of true loop-closure locations within the top-$5$ candidates for subsequent fine-grained geometric verification.
\begin{table*}[t]
    \centering
    \caption{Cross-Building Place Recognition Recall (\%) of State-Space and Spectral-Spatial Backbones Under the Variant~3 Protocol ($10.0\,\mathrm{m}$ Reference DB Segments, Spatial Tolerance $\le 3.0\,\mathrm{m}$). Parenthetical values ($\Delta$) denote absolute percentage-point differences relative to the fine-tuned Map-FM baseline in Table~\ref{tab:radish_vpr_results}. Boldface indicates the best post-adaptation score across all three backbones.}
    \label{tab:backbone_ablation}
    \footnotesize
\resizebox{\textwidth}{!}{%
    \begin{tabular}{llccccc}
        \toprule
        \textbf{Backbone Architecture} & \textbf{Metric (Stage)} & $Q = 3.2\,\mathrm{m}$ & $Q = 5.0\,\mathrm{m}$ & $Q = 7.5\,\mathrm{m}$ & $Q = 10.0\,\mathrm{m}$ & \textbf{Train Speed} \\
        \midrule
        \multirow{6}{*}{\shortstack[l]{\textbf{Mamba4Cast + LoRA}\\{\scriptsize Unidirectional Selective SSM}\\{\scriptsize (Continuous-Step, Beam-First)}}}
            & Exact R@1 (Before) & 27.78 & 27.78 & 33.33 & 25.00 & \multirow{6}{*}{$4.4\,\mathrm{s/ep}$} \\
            & \textbf{Exact R@1 (Fine-Tuned)} & 27.78 {\scriptsize ($-11.11$)} & \textbf{38.89} {\scriptsize ($+19.45$)} & 41.67 {\scriptsize ($+2.78$)} & 50.00 {\scriptsize ($\phantom{+}0.00$)} & \\
            & Spatial R@1 $\le 3.0\,\mathrm{m}$ (Before) & 50.00 & 52.78 & 50.00 & 44.44 & \\
            & \textbf{Spatial R@1 $\le 3.0\,\mathrm{m}$ (Fine-Tuned)} & \textbf{61.11} {\scriptsize ($+8.33$)} & \textbf{66.67} {\scriptsize ($+19.45$)} & 52.78 {\scriptsize ($+2.78$)} & 63.89 {\scriptsize ($-2.78$)} & \\
            & Spatial R@5 $\le 3.0\,\mathrm{m}$ (Before) & 77.78 & 83.33 & 80.56 & 77.78 & \\
            & \textbf{Spatial R@5 $\le 3.0\,\mathrm{m}$ (Fine-Tuned)} & \textbf{88.89} {\scriptsize ($+8.33$)} & \textbf{88.89} {\scriptsize ($+19.45$)} & 86.11 {\scriptsize ($+8.33$)} & \textbf{88.89} {\scriptsize ($+8.33$)} & \\
        \midrule
        \multirow{6}{*}{\shortstack[l]{\textbf{SiMBA (Bi-Mamba + EinFFT)}\\{\scriptsize Bidirectional SSM + Spectral Mix}\\{\scriptsize (Spatial-First Stem)}}}
            & Exact R@1 (Before) & 30.56 & 33.33 & 30.56 & 22.22 & \multirow{6}{*}{$\mathbf{2.6\,\mathrm{s/ep}}$} \\
            & \textbf{Exact R@1 (Fine-Tuned)} & 27.78 {\scriptsize ($-11.11$)} & 36.11 {\scriptsize ($+16.67$)} & \textbf{50.00} {\scriptsize ($+11.11$)} & \textbf{52.78} {\scriptsize ($+2.78$)} & \\
            & Spatial R@1 $\le 3.0\,\mathrm{m}$ (Before) & 50.00 & 52.78 & 47.22 & 41.67 & \\
            & \textbf{Spatial R@1 $\le 3.0\,\mathrm{m}$ (Fine-Tuned)} & 52.78 {\scriptsize ($\phantom{+}0.00$)} & 52.78 {\scriptsize ($+5.56$)} & \textbf{66.67} {\scriptsize ($+16.67$)} & \textbf{72.22} {\scriptsize ($+5.55$)} & \\
            & Spatial R@5 $\le 3.0\,\mathrm{m}$ (Before) & 77.78 & 77.78 & 72.22 & 72.22 & \\
            & \textbf{Spatial R@5 $\le 3.0\,\mathrm{m}$ (Fine-Tuned)} & 77.78 {\scriptsize ($-2.78$)} & \textbf{88.89} {\scriptsize ($+19.45$)} & \textbf{94.44} {\scriptsize ($+16.66$)} & \textbf{88.89} {\scriptsize ($+8.33$)} & \\
        \bottomrule
    \end{tabular}

}
    \vspace{1.5mm}
    \begin{flushleft}
        {\footnotesize Note: Reference baseline scores for Map-FM (TimesFM-2.5 + LoRA) are reported in Table~\ref{tab:radish_vpr_results} (Variant~3). Both SSM-based models share the exact same evaluation split ($N_{\text{test}} = 36$, $2.78\%$ random chance) and $1\,\mathrm{KB}$ descriptor footprint.}
    \end{flushleft}
\end{table*}

\begin{figure}[t]
    \centering
    \includegraphics[width=0.95\linewidth]{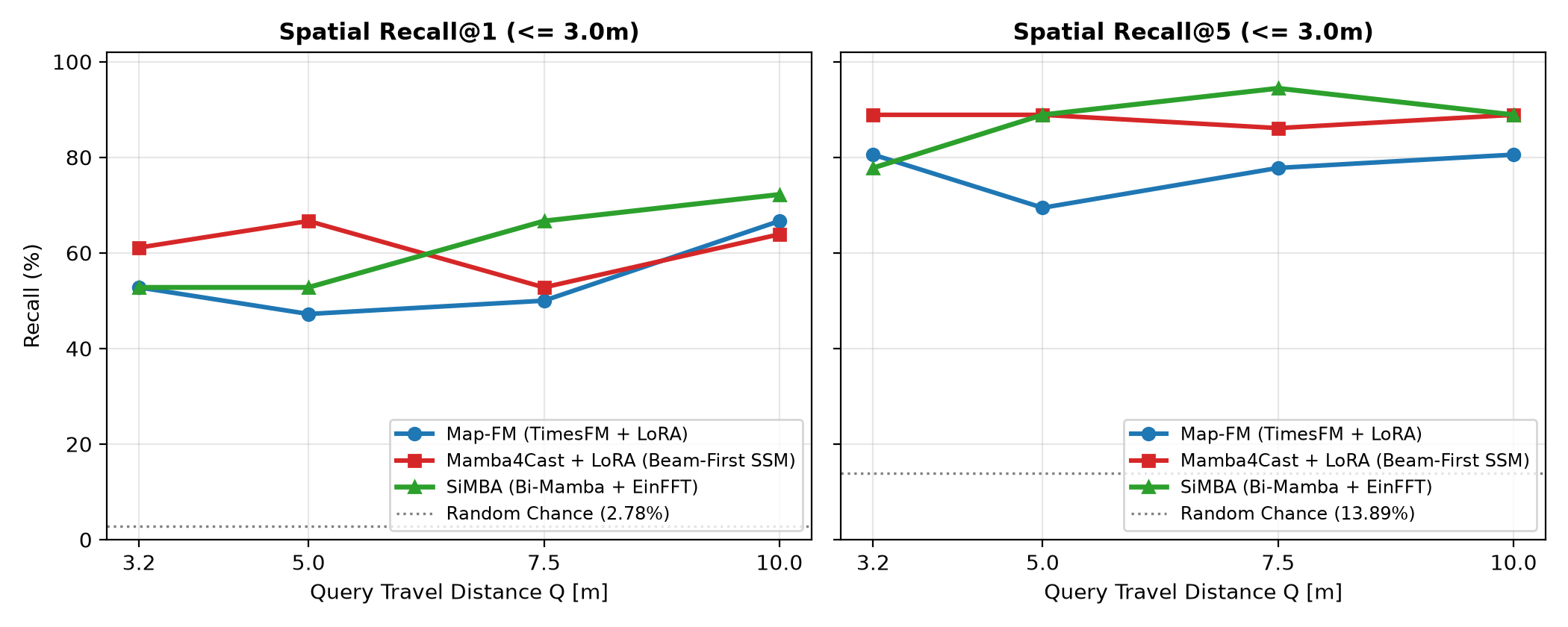}
    \caption{\textbf{Post-adaptation Spatial Recall@1 (left) and Recall@5 (right) across sequence backbones under Variant~3.} Map-FM curves are replotted from Table~\ref{tab:radish_vpr_results} for direct visual comparison. Continuous-step beam-wise SSMs (\textbf{Mamba4Cast}) excel at short horizons ($3.2$--$5.0\,\mathrm{m}$), whereas spatial-first bidirectional SSMs with spectral channel mixing (\textbf{SiMBA}) dominate long-horizon topological retrieval ($7.5$--$10.0\,\mathrm{m}$, reaching $94.44\%$ Spatial R@5).}
    \label{fig:backbone_comparison}
\end{figure}

% ==============================================================================
% Appendix B (Part 4): Architectural Ablation Across Sequence Backbones
%                      (Mamba4Cast & SiMBA vs. Map-FM Baseline)
% Note: Requires \usepackage{amsmath, amssymb, booktabs, multirow, graphicx}
%       in the preamble. Continues directly after Part 3 (suppb3.tex).
% ==============================================================================

\section{Architectural Ablation: State-Space and Spectral-Spatial Backbones}
\label{app:mapfm_backbone_ablation}

While Section~\ref{app:mapfm_exp} demonstrated the effectiveness of the patch-based Transformer foundation model (\textbf{Map-FM}) under the canonical \textbf{Variant~3} protocol, an open architectural question remains: \emph{How do modern sub-quadratic State-Space Models (SSMs) and spectral-spatial architectures perform under the exact same benchmark, and what inductive biases govern sequence-based place recognition across varying travel horizons?} In particular, Map-FM deliberately adopts a \emph{beam-first} factorization---modeling each individual range ray's temporal evolution independently before spatial aggregation---rather than compressing each instantaneous full scan line into a spatial embedding first, in order to prevent hardware-specific field-of-view (FOV) boundaries or missing rays from contaminating valid beam trajectories during cross-sensor transfer.

To investigate this without altering the data pipeline, we evaluate two alternative sequence backbones under the identical Variant~3 protocol defined in Section~\ref{app:mapfm_variants} (same train/test building splits, $10.0\,\mathrm{m}$ reference segments, symmetric InfoNCE training for $35$ epochs, and $256$-dimensional unit embeddings), directly contrasting them against the Map-FM baseline reported in Table~\ref{tab:radish_vpr_results}:

\begin{enumerate}
    \item \textbf{Mamba4Cast + LoRA (Beam-First Continuous-Step Selective SSM):}
    Replaces the patch-based TimesFM backbone with a continuous-step selective SSM ($d_{\text{model}} = 256$, $4$ selective SSM layers, $3.62\,\mathrm{M}$ total / $1.87\,\mathrm{M}$ trainable parameters) adapted via LoRA ($r = 8, \alpha = 16$) on input/output projections. Each of the $B = 64$ range beams is tokenized step-by-step together with its first-order temporal derivative $\Delta x_t = x_t - x_{t-1}$ via 1D convolutions, processed causally by selective scan kernels, and aggregated by the same masked temporal attention pooling and spatial beam convolution heads as Map-FM.

    \item \textbf{SiMBA (Spatial-First Bidirectional SSM + Spectral Mixing):}
    Evaluates an inverted \emph{spatial-first} paradigm ($5.13\,\mathrm{M}$ trainable parameters). Instead of modeling each beam as an independent 1D time series first, SiMBA compresses each instantaneous $64$-beam scan and its temporal difference $[\mathbf{x}_t, \Delta\mathbf{x}_t]$ into a $256$-dimensional spatial token via a 3-layer strided 1D convolutional stem. The resulting sequence of $T$ spatial tokens is then processed by $4$ layers combining \textbf{Bi-Mamba} (bidirectional selective SSMs capturing forward and reverse trajectory context) and \textbf{EinFFT} (fast Fourier transform-based complex-valued block-diagonal mixing across latent channels), followed by temporal attention pooling.
\end{enumerate}

\subsection{Quantitative Comparison Against Map-FM}

Table~\ref{tab:backbone_ablation} summarizes the cross-building recall of \textbf{Mamba4Cast + LoRA} and \textbf{SiMBA} before and after contrastive adaptation, explicitly reporting relative gains ($\Delta$) over the fine-tuned Map-FM baseline from Table~\ref{tab:radish_vpr_results}. Figure~\ref{fig:backbone_comparison} overlays the post-adaptation Spatial Recall@1 and Recall@5 trajectories of all three architectures across query travel distances $Q \in \{3.2, 5.0, 7.5, 10.0\}\,\mathrm{m}$.

\subsection{Architectural Findings and Design Trade-Offs}

Contrasting the new SSM-based results in Table~\ref{tab:backbone_ablation} against the Map-FM baseline in Table~\ref{tab:radish_vpr_results} yields four key insights:

\paragraph{1. Continuous-Step SSMs Eliminate Patch-Boundary Quantization Dips.}
As noted in Section~\ref{app:mapfm_exp} (Insight~2), Map-FM suffers a localized dip at $Q = 5.0\,\mathrm{m}$ ($47.22\%$ Spatial R@1) due to partial-patch zero-padding under fixed $L_p = 32$ tokenization. Because \textbf{Mamba4Cast} and \textbf{SiMBA} evolve hidden states continuously at every $0.10\,\mathrm{m}$ step without discrete patch boundaries, both avoid this quantization artifact. In particular, Mamba4Cast boosts $Q = 5.0\,\mathrm{m}$ performance by \textbf{$+19.45\%$ in both Exact R@1 ($38.89\%$) and Spatial R@1 ($66.67\%$)}, while raising Spatial R@5 to $88.89\%$.

\paragraph{2. Horizon-Dependent Duality: Beam-First Kinematics vs.\ Spatial-First Topology.}
Comparing Mamba4Cast and SiMBA exposes a clear regime shift governed by whether spatial compression occurs after or before temporal sequence modeling:
\begin{itemize}
    \item \textbf{Short Horizons ($Q \le 5.0\,\mathrm{m}$):} Beam-first modeling (\textbf{Mamba4Cast}) outperforms spatial-first compression (\textbf{SiMBA}) by $+8.33\%$ to $+13.89\%$ in Spatial R@1. Over short displacements, preserving full per-beam angular resolution alongside temporal range derivatives ($\Delta x_t$) retains subtle wall-slope cues that early spatial downsampling compresses away.
    \item \textbf{Long Horizons ($Q \ge 7.5\,\mathrm{m}$):} Spatial-first modeling with bidirectional SSM and spectral mixing (\textbf{SiMBA}) achieves the highest overall accuracy across all three backbones, reaching \textbf{$66.67\%$ Spatial R@1 / $94.44\%$ Spatial R@5 at $Q = 7.5\,\mathrm{m}$} ($+16.67\%$ over Map-FM) and \textbf{$72.22\%$ Spatial R@1 / $88.89\%$ Spatial R@5 at $Q = 10.0\,\mathrm{m}$}. Compressing each instantaneous scan into a holistic geometric token prior to Bi-Mamba and EinFFT mixing enables effective aggregation of large-scale topological transitions (e.g., corridor-to-room openings) across forward and reverse passes.
\end{itemize}

\paragraph{3. Computational Efficiency and Architectural Generality.}
Compressing the $B = 64$ spatial beams into a single latent token per step reduces SiMBA's effective sequence batch size by $64\times$ relative to beam-independent backbones, yielding a training throughput of \textbf{$2.6\,\mathrm{s}$/epoch} ($1.7\times$ faster than Mamba4Cast at $4.4\,\mathrm{s}$/epoch) with only $5.13\,\mathrm{M}$ parameters. Crucially, across all three distinct backbones, distilling variable-length range sequences into a $256$-dimensional unit descriptor ($1\,\mathrm{KB}$) consistently achieves \textbf{$80.56\%$--$94.44\%$ Spatial Recall@5} on unseen buildings, confirming that factorized spatio-temporal sequence modeling provides an architecture-agnostic foundation for fast succinct map matching.

\paragraph{4. Cross-Sensor Transferability: Beam-First vs.\ Spatial-First Paradigms.}
From a cross-sensor deployment standpoint, compressing all $B$ scan lines into a single spatial token prior to temporal aggregation (\emph{spatial-first} designs such as SiMBA or conventional single-scan encoders) couples the entire instantaneous representation to a specific sensor's angular layout; consequently, out-of-FOV padding or partial beam dropouts can contaminate the whole spatial embedding and promote overfitting to hardware-specific scan geometries. By contrast, \emph{beam-first} architectures (Map-FM and Mamba4Cast) model each range ray as an independent 1D time series before localized cross-beam convolution. Because domain-agnostic temporal waveform priors (e.g., continuous wall-range slopes and corner transitions) are extracted per beam without cross-channel mixing during sequence modeling, valid beams within the active FOV remain uncorrupted by missing or padded peripheral rays, conferring superior cross-sensor transferability among raw-observation sequence models.
% ==============================================================================
% Appendix B (Part 5): Spectral Channel Mixing Ablation (EinFFT vs. MLP)
%                      and Heading Perturbation Robustness of LoRA-SiMBA
% Note: Requires \usepackage{amsmath, amssymb, booktabs, multirow, graphicx}
%       in the preamble. Continues directly after Part 4 (suppb4.tex).
% ==============================================================================

\section{Spectral Channel Mixing Ablation and Heading Robustness of LoRA-Adapted SiMBA}
\label{app:simba_einfft_yaw}

Section~\ref{app:mapfm_backbone_ablation} demonstrated that the spatial-first \textbf{SiMBA} architecture---combining bidirectional selective state-space modeling (\textbf{Bi-Mamba}) with frequency-domain Einstein-product channel mixing (\textbf{EinFFT})---achieves the highest long-horizon recall and training throughput among all evaluated backbones. To isolate the specific contribution of spectral channel mixing and parameter-efficient adaptation, this section investigates two further questions under the canonical \textbf{Variant~3} benchmark:
\begin{enumerate}
    \item \emph{Does frequency-domain channel mixing (\textbf{EinFFT}) genuinely outperform standard spatial-domain channel mixing (\textbf{MLP}) when both are trained identically with LoRA-adapted Bi-Mamba?}
    \item \emph{How resilient is the resulting spectral-spatial representation to unseen robot heading (yaw) perturbations during inference?}
\end{enumerate}

\subsection{LoRA-Adapted SiMBA Architecture and Channel-Mixer Ablation Design}

Following the Variant~3 protocol (Section~\ref{app:mapfm_variants}), we parameterize the spatial-first backbone ($d_{\text{model}} = 256$, $4$ layers) using a $2$-layer strided 1D convolutional stem with \texttt{reflect} boundary padding (to avoid zero-padding artifacts at the $\pm 90^\circ$ field-of-view edges) followed by alternating temporal and channel-mixing blocks. To prevent overfitting on cross-building transfer, the Bi-Mamba projection layers (\texttt{in\_proj}, \texttt{out\_proj}) are adapted via Low-Rank Adaptation (\textbf{LoRA}, $r = 8, \alpha = 16$), while the convolutional stem, channel mixers, temporal attention pooling, and $256$-dimensional unit descriptor head are trained end-to-end with symmetric InfoNCE ($\tau = 0.10$, $35$ epochs, AdamW, $\text{lr} = 2\times 10^{-4}$).

We compare two identical configurations differing solely in their channel-mixing operator:
\begin{itemize}
    \item \textbf{SiMBA (Bi-Mamba + EinFFT + LoRA):} Splits the $256$-dimensional latent representation into $8$ spectral blocks ($32$ channels each), transforms across blocks via real fast Fourier transform (\texttt{rfft}), applies learnable complex-valued Einstein-product weights ($\mathbf{W} = \mathbf{W}_{\text{real}} + i\mathbf{W}_{\text{imag}}$) with soft-thresholding shrinkage ($\lambda = 0.01$), and reconstructs via inverse FFT (\texttt{irfft}). This requires \textbf{$1.13\,\mathrm{M}$ trainable parameters} ($4.64\,\mathrm{M}$ total) and runs at \textbf{$1.72\,\mathrm{s}$/epoch}.
    \item \textbf{Ablation (Bi-Mamba + MLP + LoRA):} Replaces each EinFFT module with a standard two-layer residual MLP ($\mathbb{R}^{256} \to \mathbb{R}^{256} \to \mathbb{R}^{256}$ with GELU activation), increasing trainable parameters by $+34.6\%$ to \textbf{$1.52\,\mathrm{M}$} ($5.03\,\mathrm{M}$ total, $1.33\,\mathrm{s}$/epoch).
\end{itemize}

\subsection{Impact of EinFFT Across Query Travel Horizons}

Table~\ref{tab:simba_einfft_ablation} contrasts the cross-building recall of the EinFFT and MLP variants on unseen RADISH environments (\texttt{intel}, \texttt{fr079}, \texttt{csail}) across query travel distances $Q \in \{3.2, 5.0, 7.5, 10.0\}\,\mathrm{m}$.

\begin{table*}[t]
    \centering
    \caption{Ablation of Frequency-Domain Channel Mixing (\textbf{EinFFT}) vs.\ Standard Spatial Channel Mixing (\textbf{MLP}) in LoRA-Adapted Bi-Mamba Under Variant~3 ($10.0\,\mathrm{m}$ Reference DB Segments, Spatial Tolerance $\le 3.0\,\mathrm{m}$). Boldface highlights the superior post-adaptation score per horizon.}
    \label{tab:simba_einfft_ablation}
    \footnotesize
    \resizebox{\textwidth}{!}{%
    \begin{tabular}{llccccc}
        \toprule
        \textbf{Model Variant (Trainable Params / Speed)} & \textbf{Metric (Stage)} & $Q = 3.2\,\mathrm{m}$ & $Q = 5.0\,\mathrm{m}$ & $Q = 7.5\,\mathrm{m}$ & $Q = 10.0\,\mathrm{m}$ & \textbf{Mean Spatial R@1} \\
        \midrule
        \multirow{6}{*}{\shortstack[l]{\textbf{SiMBA (Bi-Mamba + EinFFT + LoRA)}\\{\scriptsize Spectral Block Mixing ($1.13\,\mathrm{M}$ params)}\\{\scriptsize Throughput: $1.72\,\mathrm{s/epoch}$}}}
            & Exact R@1 (Before) & 25.00 & 25.00 & 36.11 & 27.78 & \multirow{2}{*}{49.31} \\
            & \textbf{Exact R@1 (Fine-Tuned)} & \textbf{27.78} & 27.78 & 33.33 & \textbf{44.44} & \\
            & Spatial R@1 $\le 3.0\,\mathrm{m}$ (Before) & 47.22 & 52.78 & 52.78 & 47.22 & \multirow{2}{*}{\textbf{61.11}} \\
            & \textbf{Spatial R@1 $\le 3.0\,\mathrm{m}$ (Fine-Tuned)} & \textbf{55.56} & \textbf{52.78} & \textbf{63.89} & \textbf{72.22} & \\
            & Spatial R@5 $\le 3.0\,\mathrm{m}$ (Before) & 72.22 & 80.56 & 80.56 & 80.56 & \multirow{2}{*}{(R@5: 86.11)} \\
            & \textbf{Spatial R@5 $\le 3.0\,\mathrm{m}$ (Fine-Tuned)} & 80.56 & 83.33 & \textbf{88.89} & \textbf{91.67} & \\
        \midrule
        \multirow{6}{*}{\shortstack[l]{\textbf{Ablation (Bi-Mamba + MLP + LoRA)}\\{\scriptsize Spatial Channel Mixing ($1.52\,\mathrm{M}$ params)}\\{\scriptsize Throughput: $1.33\,\mathrm{s/epoch}$}}}
            & Exact R@1 (Before) & 25.00 & 30.56 & 27.78 & 27.78 & \multirow{2}{*}{50.00} \\
            & \textbf{Exact R@1 (Fine-Tuned)} & 22.22 & \textbf{30.56} & 33.33 & 33.33 & \\
            & Spatial R@1 $\le 3.0\,\mathrm{m}$ (Before) & 47.22 & 52.78 & 52.78 & 47.22 & \multirow{2}{*}{54.17} \\
            & \textbf{Spatial R@1 $\le 3.0\,\mathrm{m}$ (Fine-Tuned)} & 52.78 & \textbf{52.78} & 55.56 & 55.56 & \\
            & Spatial R@5 $\le 3.0\,\mathrm{m}$ (Before) & 80.56 & 83.33 & 83.33 & 80.56 & \multirow{2}{*}{(R@5: 88.89)} \\
            & \textbf{Spatial R@5 $\le 3.0\,\mathrm{m}$ (Fine-Tuned)} & \textbf{83.33} & \textbf{88.89} & \textbf{88.89} & \textbf{91.67} & \\
        \bottomrule
    \end{tabular}%
    }
    \vspace{1.5mm}
    \begin{flushleft}
        {\footnotesize Note: Both models share identical random seeds, Bi-Mamba backbones, LoRA ranks ($r=8$), and evaluation splits ($N_{\text{test}} = 36$). Replacing MLP with EinFFT reduces trainable parameters by $25.7\%$ ($1.52\,\mathrm{M} \to 1.13\,\mathrm{M}$) while increasing $Q=10.0\,\mathrm{m}$ Spatial Recall@1 by $+16.66\%$.}
    \end{flushleft}
\end{table*}

\begin{table*}[t]
    \centering
    \caption{Heading (Yaw) Perturbation Robustness at $Q = 10.0\,\mathrm{m}$ Under Cyclic Angular Beam Shifts $\Delta b$ ($\Delta\theta = \Delta b \times 2.86^\circ$). Training augmentation only included $\Delta b \in \{-1, 0, +1\}$ ($\pm 2.86^\circ$). Boldface indicates the higher score between EinFFT and MLP at each angle.}
    \label{tab:simba_yaw_robustness}
    \footnotesize
    \resizebox{\textwidth}{!}{%
    \begin{tabular}{llccccccccc}
        \toprule
        \textbf{Model} & \textbf{Metric} & $-17.14^\circ$ & $-11.43^\circ$ & $-5.71^\circ$ & $-2.86^\circ$ & $0.00^\circ$ & $+2.86^\circ$ & $+5.71^\circ$ & $+11.43^\circ$ & $+17.14^\circ$ \\
        & & ($\Delta b=-6$) & ($\Delta b=-4$) & ($\Delta b=-2$) & ($\Delta b=-1$) & ($\Delta b=0$) & ($\Delta b=+1$) & ($\Delta b=+2$) & ($\Delta b=+4$) & ($\Delta b=+6$) \\
        \midrule
        \multirow{3}{*}{\shortstack[l]{\textbf{SiMBA}\\{\scriptsize (Bi-Mamba + EinFFT)}}}
            & Exact R@1 & 5.56 & 8.33 & \textbf{22.22} & \textbf{36.11} & \textbf{44.44} & \textbf{44.44} & \textbf{44.44} & \textbf{33.33} & 8.33 \\
            & \textbf{Spatial R@1 ($\le 3\,\mathrm{m}$)} & \textbf{22.22} & \textbf{27.78} & \textbf{47.22} & \textbf{61.11} & \textbf{72.22} & \textbf{69.44} & \textbf{66.67} & \textbf{52.78} & \textbf{27.78} \\
            & \textbf{Spatial R@5 ($\le 3\,\mathrm{m}$)} & 52.78 & \textbf{66.67} & 80.56 & \textbf{83.33} & \textbf{91.67} & \textbf{88.89} & 88.89 & 83.33 & \textbf{77.78} \\
        \midrule
        \multirow{3}{*}{\shortstack[l]{\textbf{Ablation}\\{\scriptsize (Bi-Mamba + MLP)}}}
            & Exact R@1 & \textbf{8.33} & \textbf{13.89} & \textbf{22.22} & \textbf{36.11} & 33.33 & 41.67 & 38.89 & 22.22 & \textbf{11.11} \\
            & \textbf{Spatial R@1 ($\le 3\,\mathrm{m}$)} & 16.67 & \textbf{27.78} & \textbf{47.22} & 52.78 & 55.56 & 66.67 & 63.89 & 47.22 & 22.22 \\
            & \textbf{Spatial R@5 ($\le 3\,\mathrm{m}$)} & \textbf{63.89} & \textbf{66.67} & \textbf{88.89} & \textbf{83.33} & \textbf{91.67} & \textbf{88.89} & \textbf{91.67} & \textbf{88.89} & 72.22 \\
        \bottomrule
    \end{tabular}%
    }
\end{table*}

\begin{figure}[t]
    \centering
    \includegraphics[width=0.95\linewidth]{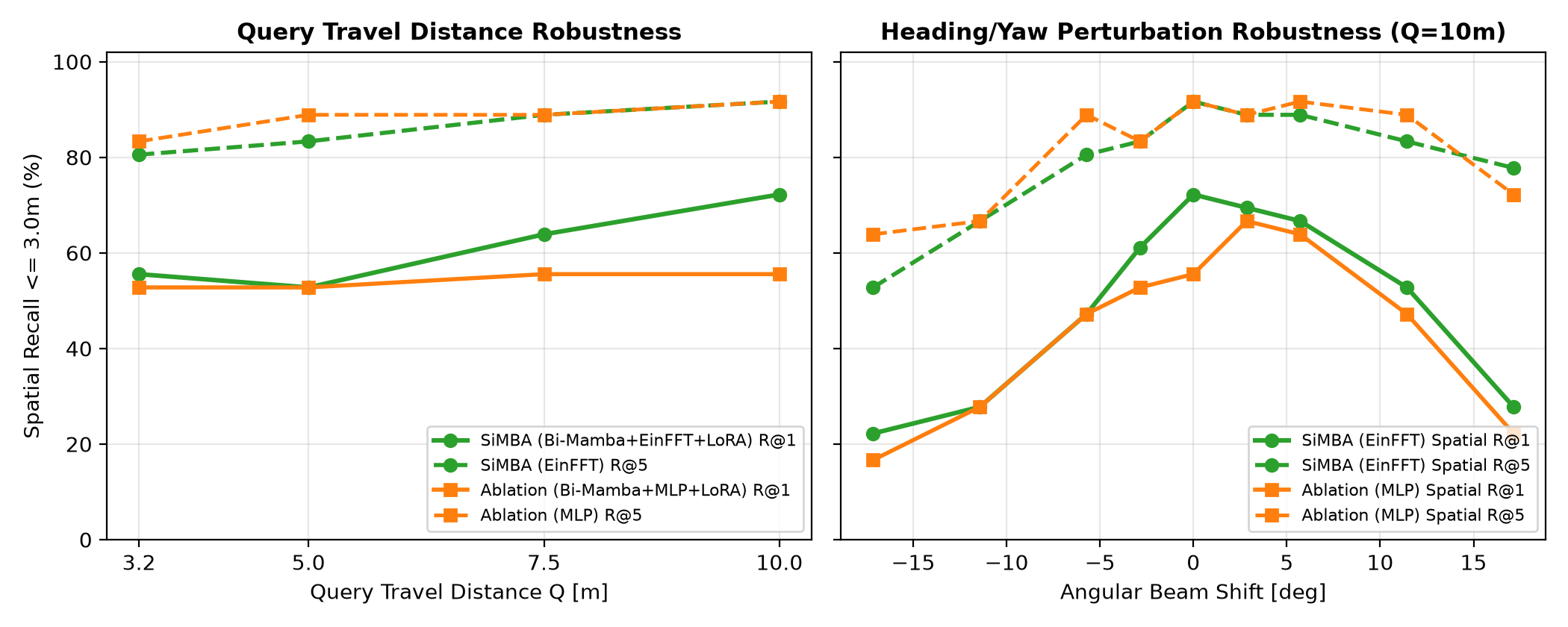}
    \caption{\textbf{Travel horizon scaling (left) and heading perturbation robustness (right) of LoRA-adapted SiMBA.} Spectral channel mixing (\textbf{EinFFT}, green) outperforms spatial MLP mixing (orange) across long travel horizons ($Q \ge 7.5\,\mathrm{m}$, reaching $72.22\%$ Spatial R@1) and maintains superior Spatial Recall@1 across the entire $\pm 17.14^\circ$ angular perturbation sweep.}
    \label{fig:simba_yaw_robustness}
\end{figure}

Despite utilizing \textbf{$25.7\%$ fewer trainable parameters} ($1.13\,\mathrm{M}$ vs.\ $1.52\,\mathrm{M}$), \textbf{EinFFT outperforms the MLP baseline in Spatial Recall@1 at every horizon above $5.0\,\mathrm{m}$}, elevating $Q = 7.5\,\mathrm{m}$ Spatial Recall@1 by $+8.33\%$ ($55.56\% \to 63.89\%$) and $Q = 10.0\,\mathrm{m}$ Spatial Recall@1 by \textbf{$+16.66\%$ ($55.56\% \to 72.22\%$)}, alongside a $+11.11\%$ gain in Exact Recall@1 ($33.33\% \to 44.44\%$). Across all four query horizons, EinFFT improves mean Spatial Recall@1 from $54.17\%$ to \textbf{$61.11\%$ ($+6.94\%$)}, confirming that block-diagonal spectral filtering regularizes cross-channel interactions against building-specific geometric overfitting.

\subsection{Heading (Yaw) Perturbation Robustness}

In real mobile robot deployments, query trajectories rarely align identically with reference database headings due to odometry drift and lane offset. Because angular beam translations in polar LiDAR scans correspond to phase rotations in the Fourier domain while preserving spectral magnitudes, EinFFT is theoretically well suited to tolerate moderate heading perturbations. To test this empirically, we apply synthetic cyclic beam shifts $\Delta b \in \{-6, -4, -2, -1, 0, +1, +2, +4, +6\}$ (spanning $-17.14^\circ$ to $+17.14^\circ$ at $2.86^\circ$/beam) to the $Q = 10.0\,\mathrm{m}$ test queries---well beyond the $\pm 1$-beam ($\pm 2.86^\circ$) training augmentation.

As shown in Table~\ref{tab:simba_yaw_robustness} and Fig.~\ref{fig:simba_yaw_robustness} (right), \textbf{SiMBA with EinFFT achieves higher or equal Spatial Recall@1 than the MLP ablation at all nine tested yaw offsets ($\pm 17.14^\circ$)}, averaging \textbf{$45.37\%$ Spatial Recall@1} across the full sweep versus $40.74\%$ for MLP ($+4.63\%$). Within moderate heading deviations of $\pm 5.71^\circ$ ($\pm 2$ beams, double the training augmentation range), EinFFT maintains \textbf{$47.22\%$--$72.22\%$ Spatial Recall@1} and \textbf{$80.56\%$--$91.67\%$ Spatial Recall@5}, while preserving $44.44\%$ Exact Recall@1 across $0^\circ$ to $+5.71^\circ$. Beyond $\pm 11.43^\circ$ ($\pm 4$ beams), recall degrades gracefully for both models because cyclic wrapping across a $180^\circ$ frontal field of view folds left-edge wall returns onto the right-edge boundary, introducing physical non-periodic discontinuities that exceed local phase-shift invariance.

\subsection{Cross-Sensor Transferability and Physical Limitations}
\label{app:mapfm_limitations}

While the occupancy-grid representations in the main paper (PartSLAM) and Supplementary Material~A (BB-SAM) decouple scene modeling from sensor-specific kinematics once a map is integrated, Map-FM operates directly on raw observation streams $\mathbf{X} \in \mathbb{R}^{B \times T}$ and thus exhibits narrower cross-platform transferability under specific hardware deviations:
\begin{enumerate}
    \item \textbf{Sensor-to-Base Extrinsic Offset (Lever-Arm Effect):}
    Variant~3 harmonizes scanners onto a canonical frontal polar grid $[-\pi/2, +\pi/2]$, and our yaw-robustness analysis models heading deviations as cyclic beam shifts $\Delta b$. While minor extrinsic offsets between the sensor optical center and the robot's rotation axis are negligible under near-straight traversal, a large lever-arm offset combined with sharp turning or large heading deviations introduces translational parallax and non-uniform range variations that cannot be captured solely by a cyclic angular shift $\Delta b$.

    \item \textbf{Non-Polar and Non-Radial Sensor Geometries:}
    Both the spatial beam encoder and spectral mixing assume that adjacent channels correspond to equiangular rays radiating from a single optical center. Direct transfer is therefore limited for distributed rangefinder arrays (e.g., multiple short-range ToF sensors mounted around a chassis whose rays do not share a common origin) or perspective depth sensors unless raw measurements are first re-projected onto a unified virtual polar scan centered at the robot base frame.

    \item \textbf{Wide-Beam and Specular Modalities (e.g., Ultrasonic Sonar):}
    Unlike narrow-beam LiDARs that track continuous wall-slope derivatives along well-defined bearings, ultrasonic sensors have wide acoustic beam cones ($15^\circ$--$30^\circ$), return the minimum distance within the cone rather than a ray-cast intersection, and suffer from severe specular reflections and multipath dropouts on oblique walls. Applying Map-FM directly to sonar arrays violates the fine-grained angular continuity assumed by the spatial stem, making grid-based methods (PartSLAM and BB-SAM) better suited for sparse or noisy range modalities.
\end{enumerate}

\setcounter{section}{0}
% \setcounter{figure}{0}
% \setcounter{table}{0}
% \setcounter{equation}{0}
% \setcounter{page}{1} % ページ番号も1からに戻す場合（不要なら削除）

% --- 3. ★ 2回目の \maketitle を許可するための内部フラグ調整 ---
\if@twocolumn
  \twocolumn
\fi
\makeatother

\vspace*{1ex}
{\centering
  {\Large \bfseries Supplementary Material C: Scalable Map Matching Using 2D-Flatness Homogeneous Foundation Integrals and Single-Pass 3-DoF Partial LiDAR Alignment
 \par}
  \vspace{1ex}
  \vspace{3ex}
}

\input suppc1.tex

\input suppc2.tex

\end{document}

%% file: suppc1.tex
\section{Introduction}
\label{sec:introduction}

In the main paper (PartSLAM), we introduced an indirect and hybrid map-matching framework (\texttt{iMM}/\texttt{hMM}) that represents local and global maps via compact collections of axis-aligned bounding-box (BB) parts mined from a known dictionary map $\mathcal{P}^{\text{Dict}}$, achieving orders-of-magnitude gains in speed and memory compactness over direct point-cloud matching (\texttt{dMM}). However, the original PartSLAM formulation relies on three simplifying assumptions that limit its applicability to incremental Simultaneous Localization and Mapping (SLAM) and multi-session map merging:
\begin{enumerate}
  \item \textbf{Pre-aligned Manhattan orientation and high map coverage:} PartSLAM assumes that input maps are pre-aligned to canonical axes via entropy minimization and that candidate parts cover a large fraction of the input map ($T^{\text{size}} = 0.9$). When an incoming query submap $\mathcal{P}^Q$ captures only a small, unvisited fraction ($12\%\text{--}25\%$) of $\mathcal{P}^{\text{DB}}$ or $\mathcal{P}^{\text{Dict}}$ under arbitrary planar yaw drift, axis-aligned BB overlap and unweighted grid pooling break down due to empty-region dilution.
  \item \textbf{Rotational latency in feature-based BB descriptors:} Extending BB descriptors from axis-aligned occupancy grids to deep vision foundation features (e.g., DINOv2) enables cross-environment generalization without manual dictionary curation, but re-running deep feature extractors across multiple rotated map hypotheses multiplies inference latency linearly with angular resolution.
  \item \textbf{Coarse-to-fine pose recovery bottleneck:} While PartSLAM's descriptor matcher (\texttt{DM}) rapidly retrieves candidate global maps and its \texttt{rerank} cascade refines top candidates using raw point-cloud RANSAC (\texttt{dMM}), recovering precise 3-DoF planar rigid transformations $T(\mathbf{p}; \Delta x, \Delta y, \theta) \in \mathrm{SE}(2)$ across wide-span ($400\,\text{m}$) corridors without storing dense point clouds or suffering longitudinal drift remains an open challenge.
\end{enumerate}

Complementing Supplementary Material~A (\texttt{BB-SAM}, which distills discrete $D_4$-symmetric structural dictionaries) and Supplementary Material~B (\texttt{Map-FM}, which embeds raw range sequences for 1-to-$N$ candidate retrieval), this appendix generalizes PartSLAM's discrete BB-based scene representation into a \emph{continuous, rotation-equivariant foundation-feature integral framework} for scalable map matching. Specifically, we adapt a DINOv2 ViT-S/14 backbone with a \emph{2D-Flatness} spatial adapter to map 3-channel occupancy tensors onto a dense $32 \times 32$ feature lattice paired with a continuous observation validity mask. By constructing a Homogeneous Summed-Area Table (\texttt{MapHomogeneousIntegral2D}), we extract validity-normalized Spatial Pyramid Matching (SPM $2\times 2$) part descriptors over arbitrary continuous observed BBs in $O(1)$ time—eliminating both empty-region dilution and discrete anchor constraints. Furthermore, we synthesize multi-angle query part descriptors via a \emph{single-pass} feature-space rotation and recover 3-DoF poses via global hypothesis pooling, Preemptive RANSAC pruning, and Iteratively Reweighted Least Squares (IRLS) 2D-SVD refinement. Across 12 real-world RADISH sequences spanning $80\,\text{m}$ to $404.7\,\text{m}$, our framework achieves $1.51\text{--}2.51\,\text{m}$ mean translation error and $2.11^\circ\text{--}4.82^\circ$ mean rotation error across $12\%\text{--}100\%$ partial map coverages with invariant $167\text{--}179\,\text{ms}$ online CPU latency.

\section{Related Work}
\label{sec:related_work}

Existing approaches to 2D LiDAR place recognition and submap alignment broadly fall into three paradigms: (i)~global scan-to-map descriptors and appearance-based loop-closure schemes~\cite{5,6} (such as Scan Context and LiDAR Iris), which compress polar or Cartesian occupancy statistics into compact vectors for fast retrieval but degrade sharply under severe partial overlap ($12\%\text{--}25\%$); (ii)~local keypoint and geometric feature matchers~\cite{4,24,25} paired with RANSAC, which estimate rigid transformations from point or patch correspondences but incur high computational cost and ambiguity in repetitive corridor environments; and (iii)~hierarchical spatial pyramid or sliding-window matching methods—including PartSLAM's discrete BB representation—that evaluate candidate regions over occupancy grids.

While self-supervised vision foundation models such as DINOv2 provide strong structural priors across domains, their raw token manifolds exhibit substantial spatial non-linearity and boundary saliency concentration, causing linear spatial pooling and sub-grid interpolation to fluctuate when candidate windows shift continuously. Summed-Area Tables (SATs) enable constant-time $O(1)$ box aggregation, yet standard unweighted SATs suffer from descriptor dilution over partially observed maps where empty cells dominate. Our work bridges these gaps by rectifying the 2D foundation feature lattice for linear spatial additivity and coupling it with a validity-weighted homogeneous SAT and single-pass feature-space rotation.

\section{Proposed Method}
\label{sec:method}

Figure~\ref{fig:bb_overlay} summarizes our four-stage pipeline: (i)~3-channel occupancy encoding and 2D-Flatness geometric transfer learning, (ii)~homogeneous integral image construction for $O(1)$ valid-region SPM $2\times 2$ part queries, (iii)~single-pass feature-space yaw rotation and global hypothesis pooling, and (iv)~Preemptive RANSAC pruning followed by IRLS-weighted 2D-SVD refinement.

\subsection{3-Channel Occupancy Encoding and 2D-Flatness Transfer Learning}
Given a 2D LiDAR submap $\mathcal{P}$ spanning physical extent $S_{\text{span}}$, we rasterize $\mathcal{P}$ into a $256 \times 256$ occupancy grid $O \in \{0, 127, 255\}^{256 \times 256}$ at resolution $r = S_{\text{span}} / 256\,\text{m/px}$, where $0$, $127$, and $255$ denote occupied walls, unobserved cells, and free space, respectively. To bridge the modality gap to visual foundation models while preserving observability masks, we construct a 3-channel tensor $\mathbf{X} \in \mathbb{R}^{3 \times 448 \times 448}$ combining occupancy intensity, smoothed Euclidean Distance Transform (EDT) wall proximity, and free-space visibility, paired with a dense observation validity mask $W \in [0, 1]^{G \times G}$ ($G = 32$) measuring the fraction of observed (occupied or free) cells within each grid patch.

We fine-tune the last two transformer blocks of a DINOv2 ViT-S/14 backbone followed by a lightweight residual projection head (\texttt{SpatialFlatAdapter2D}, structured as a two-layer MLP with LayerNorm and GELU activation) to map raw patch tokens $\mathbf{Z}_{\text{raw}}(y,x)$ onto an $\ell_2$-normalized spatial feature lattice $\mathbf{F} \in \mathbb{R}^{G \times G \times D}$ ($D = 128$):
\begin{equation}
  \mathbf{F}(y, x) = \frac{\mathbf{Z}_{\text{raw}}(y, x) + \alpha_{\text{res}}\,\mathrm{MLP}(\mathbf{Z}_{\text{raw}}(y, x))}{\|\mathbf{Z}_{\text{raw}}(y, x) + \alpha_{\text{res}}\,\mathrm{MLP}(\mathbf{Z}_{\text{raw}}(y, x))\|_2}.
\end{equation}
Because 2D prefix summation (SAT pooling) is a linear operator in ambient Euclidean space prior to hyperspherical $\ell_2$ normalization, continuous sub-grid box queries remain well-calibrated only if the underlying feature lattice varies smoothly and additively across space. Training with contrastive losses alone concentrates saliency on isolated wall corners, causing bilinearly interpolated box descriptors to oscillate under sub-grid shifts. To enforce local spatial linearity, we jointly optimize a spatial hard-negative margin loss $\mathcal{L}_{\text{margin}}$ (contrasting overlapping positive submaps against spatially shifted low-IoU distractors) and a \emph{2D-Flatness} structural consistency loss $\mathcal{L}_{\text{flat2D}}$ that aligns each box's validity-weighted average descriptor $\bar{\mathbf{f}}(B)$ with the normalized sum of its four $2\times 2$ sub-quadrants $\{B_q\}_{q=1}^4$:
\begin{equation}
  \mathcal{L}_{\text{flat2D}} = \mathbb{E}_{B}\!\left[
    1 - \left\langle \bar{\mathbf{f}}(B),\;
    \frac{\sum_{q=1}^{4} \bar{\mathbf{f}}(B_q)}{\|\sum_{q=1}^{4} \bar{\mathbf{f}}(B_q)\|_2}
    \right\rangle
  \right].
\end{equation}
This geometric rectification suppresses manifold curvature across the 2D lattice, ensuring that continuous bilinear interpolation and multi-scale SAT aggregation commute algebraically without re-encoding cropped submaps.

\subsection{Homogeneous SAT and Valid-Region SPM $2\times 2$ Queries}
Unlike discrete keypoint-snapped box proposals that become brittle in partially observed LiDAR fragments ($12\%\text{--}25\%$ coverage), we sample continuous multi-scale bounding boxes $\mathcal{B} = \{B_m\}_{m=1}^{M}$ directly over observed regions satisfying mean validity $\bar{W}(B_m) > 0.12$. To prevent unobserved cells ($W(y,x)=0$) from diluting pooled features across arbitrary continuous coordinates $B = [y_1, y_2] \times [x_1, x_2] \subset [0, G]^2$, we augment each lattice cell with its scalar validity weight to form a $(D+1)$-dimensional homogeneous tensor $[\tilde{\mathbf{F}}(y,x), W(y,x)] = [W(y,x)\mathbf{F}(y,x), W(y,x)] \in \mathbb{R}^{D+1}$ and precompute its zero-padded 2D prefix sums $\mathbf{S}_F \in \mathbb{R}^{(G+1)\times(G+1)\times D}$ and $S_W \in \mathbb{R}^{(G+1)\times(G+1)}$:
\begin{equation}
  \mathbf{S}_F(y, x) = \sum_{u=1}^{y}\sum_{v=1}^{x} W(u, v)\mathbf{F}(u, v), \qquad
  S_W(y, x) = \sum_{u=1}^{y}\sum_{v=1}^{x} W(u, v).
\end{equation}
For any real-valued sub-grid endpoint $(y, x)$, we evaluate $\hat{\mathbf{S}}_F(y, x)$ and $\hat{S}_W(y, x)$ via 4-neighbor bilinear interpolation over surrounding integer lattice nodes, and recover the unnormalized feature integral $\mathbf{S}_F(B)$ and effective observed mass $S_W(B)$ via four-corner inclusion-exclusion arithmetic in $O(1)$ time:
\begin{equation}
  \mathbf{S}_F(B) = \hat{\mathbf{S}}_F(y_2, x_2) - \hat{\mathbf{S}}_F(y_1, x_2) - \hat{\mathbf{S}}_F(y_2, x_1) + \hat{\mathbf{S}}_F(y_1, x_1),
\end{equation}
with $S_W(B)$ computed identically. Dividing $\mathbf{S}_F(B)$ by $\max(S_W(B), \epsilon)$ cancels empty-cell dilution and yields the exact validity-normalized regional centroid $\bar{\mathbf{f}}(B) = \mathbf{S}_F(B) / \max(S_W(B), \epsilon)$. Concatenating and $\ell_2$-normalizing the full-box descriptor and its four $2\times 2$ quadrant descriptors produces a rotation-sensitive $640$-D spatial pyramid part descriptor:
\begin{equation}
  \mathbf{v}(B) = \frac{
    [\bar{\mathbf{f}}(B);\bar{\mathbf{f}}(B_1);\bar{\mathbf{f}}(B_2);\bar{\mathbf{f}}(B_3);\bar{\mathbf{f}}(B_4)]
  }{
    \|[\bar{\mathbf{f}}(B);\bar{\mathbf{f}}(B_1);\bar{\mathbf{f}}(B_2);\bar{\mathbf{f}}(B_3);\bar{\mathbf{f}}(B_4)]\|_2
  } \in \mathbb{R}^{640}.
\end{equation}
Database descriptors $\mathbf{V}^{\text{DB}} \in \mathbb{R}^{512 \times 640}$ ($M_{\text{DB}} = 512$) are precomputed offline.

\subsection{Single-Pass Feature-Space Rotation and Global Hypothesis Pooling}
Because 2D-Flatness rectification preserves a smooth, spatially equivariant correspondence between physical coordinates and the $32 \times 32$ feature lattice, in-plane yaw rotations can be synthesized directly in feature space without re-running the vision backbone. Specifically, we execute DINOv2 \emph{once} on $\mathbf{X}^Q$ to obtain $(\mathbf{F}^Q, W^Q)$ and synthesize rotated feature maps $(\mathbf{F}^Q_{\theta_a}, W^Q_{\theta_a})$ on the $32 \times 32$ lattice via bilinear \texttt{affine\_grid} sampling across discrete yaw hypotheses $\Theta = \{-15^\circ, -10^\circ, \dots, +15^\circ\}$ ($\vert{}\Theta\vert{} = 7$). At each $\theta_a \in \Theta$, we query $M_Q = 112$ valid boxes via the rotated homogeneous SAT and match $\mathbf{V}^Q_{\theta_a}$ against $\mathbf{V}^{\text{DB}}$. For the top-$14$ correspondences at each angle, we solve the planar rigid transformation $T(\mathbf{p}; \Delta x, \Delta y, \theta_a) \in \mathrm{SE}(2)$ around grid center $\mathbf{c} = (128, 128)^\top$ in closed form:
\begin{equation}
  \begin{bmatrix} \Delta x_{i,a} \\ \Delta y_{i,a} \end{bmatrix} =
  \mathbf{p}_i^{\text{DB}} - \mathbf{R}(\theta_a)(\mathbf{p}_i^Q - \mathbf{c}) - \mathbf{c},
  \label{eq:rigid3dof}
\end{equation}
and pool all hypotheses globally. The consensus angle $\theta_{a^*}$ maximizes total inlier similarity score within a scale-adaptive threshold $\tau_{\text{inl}} = \mathrm{clip}(2.4 / r,\, 5.5,\, 7.5)\,\text{px}$.

\subsection{Preemptive RANSAC and IRLS-Weighted 2D-SVD Refinement}
Around $\theta_{a^*}$, we generate fine yaw hypotheses $\theta_{a^*} + \delta\theta$ ($\delta\theta \in [-0.05, +0.05]\,\text{rad}$, $5$ steps) and prune them via a 3-stage \texttt{PreemptiveRANSACPruner} with keep counts $(48, 12, 1)$. To suppress longitudinal corridor drift on wide-span maps ($S_{\text{span}} \approx 400\,\text{m}$), we refine $(\Delta x, \Delta y, \theta)$ over $5$ iterations of IRLS-weighted 2D-SVD alignment. Inliers ($e_i = \Vert{}T(\mathbf{p}_i^Q) - \mathbf{p}_i^{\text{DB}}\Vert{}_2 \le \tau_{\text{inl}}$) receive Cauchy-damped weights:
\begin{equation}
  \tilde{w}_i = \frac{s_i}{1 + (e_i / (0.5\,\tau_{\text{inl}}))^2}, \qquad
  w_i = \frac{\tilde{w}_i}{\sum_{j \in \mathcal{V}} \tilde{w}_j},
\end{equation}
and $(\mathbf{R}, \mathbf{t})$ is updated in closed form via weighted cross-covariance SVD $\mathbf{H} = \sum_{i \in \mathcal{V}} w_i (\tilde{\mathbf{p}}_i^Q - \bar{\mathbf{p}}^Q)(\tilde{\mathbf{p}}_i^{\text{DB}} - \bar{\mathbf{p}}^{\text{DB}})^\top = \mathbf{U}\boldsymbol{\Sigma}\mathbf{V}^\top$, $\mathbf{R} = \mathbf{V}\mathrm{diag}(1, \det(\mathbf{V}\mathbf{U}^\top))\mathbf{U}^\top$.

%% file: suppc2.tex
\section{Experimental Evaluation}
\label{sec:experiments}

To validate our continuous foundation-feature integral extension of PartSLAM under relaxed orientation and observability assumptions, we evaluate our single-pass DINOv2-SAT and IRLS-2D-SVD 3-DoF alignment pipeline on 12 real-world 2D LiDAR sequences from the RADISH dataset~\cite{15} spanning physical extents $S_{\text{span}}$ from $80.0\,\text{m}$ to $404.7\,\text{m}$ (\texttt{MIT\_Infinite\_Corridor}, \texttt{aces\_publicb}, \texttt{belgioioso}, \texttt{edmonton}, \texttt{fhw-rec-001}, \texttt{fr-campus}, \texttt{fr079}, \texttt{fr101}, \texttt{intel}, \texttt{mexico}, \texttt{orebro}, and \texttt{seattle-r}). Unlike the main paper's pre-aligned evaluation, query submaps here undergo an uncompensated planar rigid perturbation of $+10.0^\circ$ yaw rotation and $(+8.0, -6.0)\,\text{px}$ translation relative to the reference frame without entropy-based axis pre-alignment, and all online latencies are measured end-to-end on CPU.

\subsection{Qualitative Valid-Region BB Overlay and 3-DoF Alignment}
\begin{figure}[t]
  \centering
  \includegraphics[width=\linewidth]{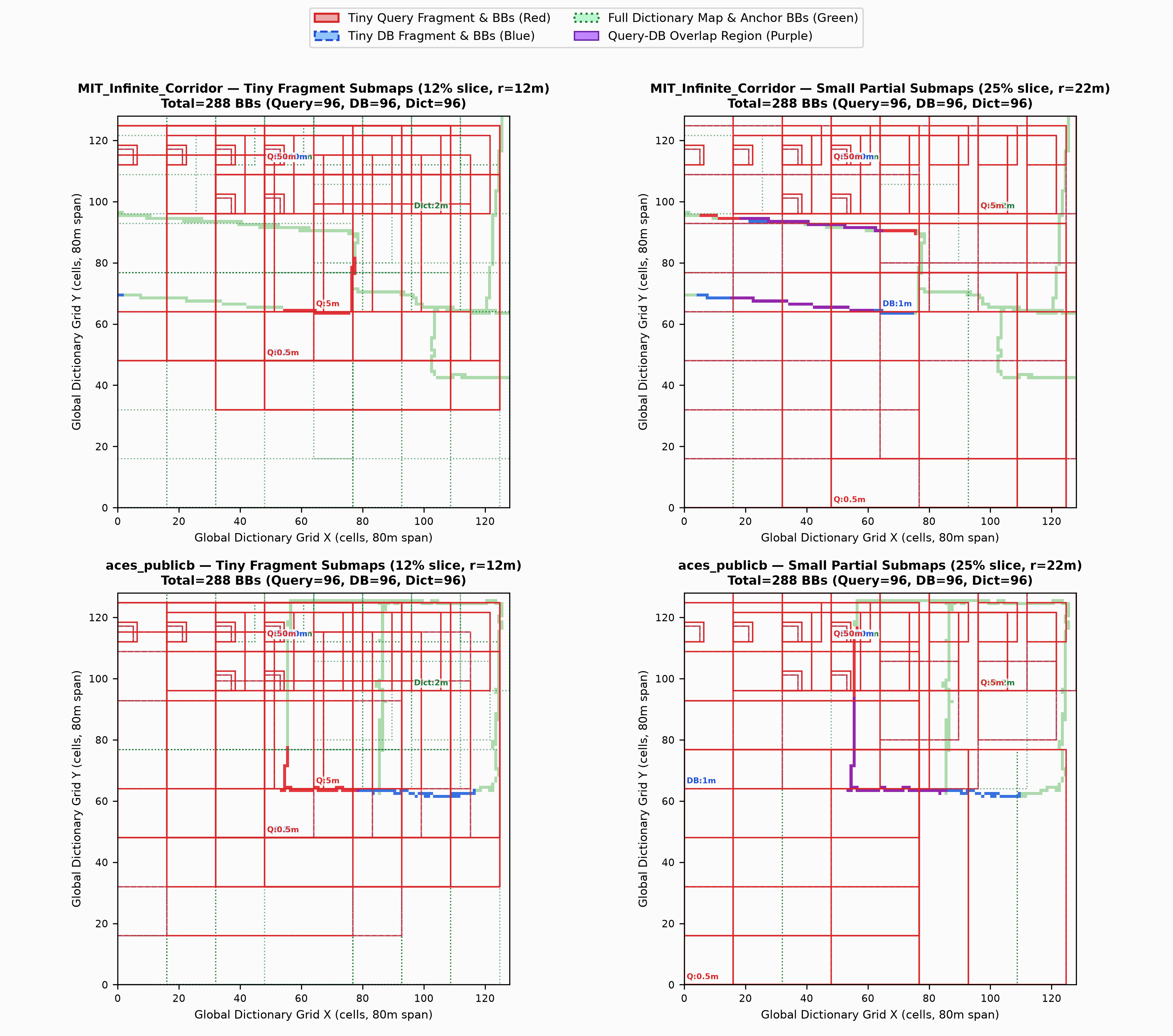}
  \caption{Qualitative 3-DoF alignment and valid-region continuous BB overlay on RADISH maps (\texttt{MIT\_Infinite\_Corridor} and \texttt{aces\_publicb}) comparing aligned Query fragments (red, solid), partial DB fragments (blue, dashed), and the full $100\%$ Dictionary map (green). Purple cells indicate overlapping Query--DB structural walls.}
  \label{fig:bb_overlay}
\end{figure}

Figure~\ref{fig:bb_overlay} shows qualitative 3-DoF alignment and continuous valid-region bounding-box extraction on \texttt{MIT\_Infinite\_Corridor} and \texttt{aces\_publicb} under severe partial trajectories ($Q=12\% / \text{DB}=25\%$ and $Q=25\% / \text{DB}=50\%$) against the full $100\%$ global Dictionary map. Unlike uniform sliding grids that place boxes over unobserved regions, validity-filtered sampling ($\bar{W} > 0.12$) concentrates continuous part BBs along observed walls and corridor junctions, enabling accurate alignment even under severe partial observability ($12\%$ query coverage).

\subsection{Partial-to-Full Trajectory Coverage Benchmark}
To simulate incremental SLAM relocalization and multi-session map merging where incoming submaps cover only a fraction of the environment, we evaluate four trajectory coverage ratios across all 12 RADISH sequences: $(Q=12\%, \text{DB}=25\%)$, $(Q=25\%, \text{DB}=50\%)$, $(Q=50\%, \text{DB}=100\%)$, and $(Q=100\%, \text{DB}=100\%)$ against a $100\%$ reference dictionary. Table~\ref{tab:coverage_summary} summarizes the mean translation error, mean rotation error, and mean online inference latency across all 12 sequences.

\begin{table*}[t]
  \centering
  \caption{Mean 3-DoF pose accuracy and end-to-end online latency across all 12 RADISH sequences under varying trajectory coverages.}
  \label{tab:coverage_summary}
  \scriptsize
  \begin{tabular}{cccc}
    \toprule
    \textbf{Coverage (Q / DB / Dict)} & \textbf{Trans.\ Err.\ (m)} & \textbf{Rot.\ Err.\ ($^\circ$)} & \textbf{Online Time (ms)} \\
    \midrule
    $12\% / 25\% / 100\%$  & $2.51$ & $4.82$ & $178.8$ \\
    $25\% / 50\% / 100\%$  & $2.49$ & $3.82$ & $171.8$ \\
    $50\% / 100\% / 100\%$ & $1.81$ & $3.26$ & $167.0$ \\
    $\mathbf{100\% / 100\% / 100\%}$ & $\mathbf{1.51}$ & $\mathbf{2.11}$ & $\mathbf{168.6}$ \\
    \bottomrule
  \end{tabular}
\end{table*}
\begin{table}[t]
  \centering
  \caption{Per-sequence 3-DoF alignment error and online latency across 12 RADISH sequences ($Q=50\%/\text{DB}=100\%$ and $Q=100\%/\text{DB}=100\%$).}
  \label{tab:per_sequence}
  \scriptsize
  \setlength{\tabcolsep}{3.5pt}
  \begin{tabular}{lcccccc}
    \toprule
    & & \multicolumn{2}{c}{\textbf{Q=50\% / DB=100\%}} & \multicolumn{3}{c}{\textbf{Q=100\% / DB=100\%}} \\
    \cmidrule(lr){3-4} \cmidrule(lr){5-7}
    \textbf{Sequence} & \textbf{Span} & \textbf{Trans.} & \textbf{Rot.} & \textbf{Trans.} & \textbf{Rot.} & \textbf{Time} \\
    \midrule
    \texttt{MIT\_Corridor} & $395.4\,\text{m}$ & $10.33\,\text{m}$ & $8.19^\circ$ & $3.50\,\text{m}$ & $1.79^\circ$ & $169.9\,\text{ms}$ \\
    \texttt{aces\_publicb} & $80.0\,\text{m}$  & $0.60\,\text{m}$  & $2.94^\circ$ & $0.90\,\text{m}$ & $1.42^\circ$ & $167.9\,\text{ms}$ \\
    \texttt{belgioioso}    & $97.2\,\text{m}$  & $0.38\,\text{m}$  & $1.49^\circ$ & $1.89\,\text{m}$ & $1.00^\circ$ & $170.0\,\text{ms}$ \\
    \texttt{edmonton}      & $115.4\,\text{m}$ & $0.24\,\text{m}$  & $3.56^\circ$ & $1.23\,\text{m}$ & $1.46^\circ$ & $168.8\,\text{ms}$ \\
    \texttt{fhw-rec-001}   & $80.0\,\text{m}$  & $0.46\,\text{m}$  & $3.82^\circ$ & $0.75\,\text{m}$ & $1.94^\circ$ & $166.9\,\text{ms}$ \\
    \texttt{fr-campus}     & $404.7\,\text{m}$ & $4.81\,\text{m}$  & $0.10^\circ$ & $3.92\,\text{m}$ & $2.14^\circ$ & $167.4\,\text{ms}$ \\
    \texttt{fr079}         & $80.0\,\text{m}$  & $0.24\,\text{m}$  & $6.42^\circ$ & $0.49\,\text{m}$ & $7.96^\circ$ & $171.8\,\text{ms}$ \\
    \texttt{fr101}         & $80.0\,\text{m}$  & $0.29\,\text{m}$  & $3.19^\circ$ & $0.76\,\text{m}$ & $0.51^\circ$ & $167.9\,\text{ms}$ \\
    \texttt{intel}         & $80.0\,\text{m}$  & $1.28\,\text{m}$  & $3.36^\circ$ & $1.60\,\text{m}$ & $1.88^\circ$ & $167.0\,\text{ms}$ \\
    \texttt{mexico}        & $120.1\,\text{m}$ & $1.54\,\text{m}$  & $2.06^\circ$ & $1.31\,\text{m}$ & $2.08^\circ$ & $168.3\,\text{ms}$ \\
    \texttt{orebro}        & $80.9\,\text{m}$  & $0.27\,\text{m}$  & $2.39^\circ$ & $0.82\,\text{m}$ & $1.33^\circ$ & $167.0\,\text{ms}$ \\
    \texttt{seattle-r}     & $80.0\,\text{m}$  & $1.23\,\text{m}$  & $1.60^\circ$ & $0.94\,\text{m}$ & $1.84^\circ$ & $170.8\,\text{ms}$ \\
    \midrule
    \textbf{Mean (12 logs)} & --- & $\mathbf{1.81\,\text{m}}$ & $\mathbf{3.26^\circ}$ & $\mathbf{1.51\,\text{m}}$ & $\mathbf{2.11^\circ}$ & $\mathbf{168.6\,\text{ms}}$ \\
    \bottomrule
  \end{tabular}
\end{table}
\begin{table}[t]
  \centering
  \caption{Key Architecture, Training, and Inference Hyperparameters for Supplementary Material~C.}
  \label{tab:hyperparams}
  \scriptsize
  \setlength{\tabcolsep}{4pt}
\resizebox{\columnwidth}{!}{%
  \begin{tabular}{lll}
    \toprule
    \textbf{Stage / Module} & \textbf{Parameter} & \textbf{Value / Specification} \\
    \midrule
    \multirow{3}{*}{\textbf{Input Rasterization}}
      & Occupancy grid size & $256 \times 256$ ($\{0, 127, 255\}$) \\
      & Spatial resolution ($r$) & $S_{\text{span}} / 256\,\text{m/px}$ ($80.0\text{--}404.7\,\text{m}$) \\
      & Input tensor ($\mathbf{X}$) & $3 \times 448 \times 448$ (Occ., EDT, Vis.) \\
    \midrule
    \multirow{4}{*}{\textbf{Backbone \& Adapter}}
      & Vision backbone & DINOv2 ViT-S/14 (last 2 blocks tuned) \\
      & Feature lattice ($G \times G \times D$) & $32 \times 32 \times 128$ ($\ell_2$-normalized) \\
      & Adapter head & 2-layer residual MLP (LN + GELU) \\
      & Training losses & Spatial margin ($\mathcal{L}_{\text{margin}}$) + $\mathcal{L}_{\text{flat2D}}$ \\
    \midrule
    \multirow{4}{*}{\textbf{Homogeneous SAT}}
      & Validity mask ($W$) & $[0, 1]^{32 \times 32}$, box threshold $\bar{W} > 0.12$ \\
      & Part descriptor ($\mathbf{v}(B)$) & $640$-D ($\text{Full} + 2\times 2\text{ SPM}$, $\ell_2$-norm) \\
      & Database / Query boxes & $M_{\text{DB}} = 512$ (offline), $M_Q = 112$ (online) \\
      & Box query complexity & $O(1)$ via bilinear-interpolated 2D SAT \\
    \midrule
    \multirow{5}{*}{\textbf{3-DoF Alignment}}
      & Coarse yaw search ($\Theta$) & $\{-15^\circ, -10^\circ, \dots, +15^\circ\}$ ($|\Theta| = 7$) \\
      & Top correspondences & Top-$14$ pairs per yaw hypothesis \\
      & Inlier threshold ($\tau_{\text{inl}}$) & $\mathrm{clip}(2.4 / r,\, 5.5,\, 7.5)\,\text{px}$ \\
      & Fine yaw / Preemptive RANSAC & $\pm 0.05\,\text{rad}$ ($5$ steps), keep $(48, 12, 1)$ \\
      & IRLS 2D-SVD refinement & $5$ iters, Cauchy scale $0.5\,\tau_{\text{inl}}$ \\
    \bottomrule
  \end{tabular}
}
\end{table}

As query and database coverages increase from $12\% / 25\%$ to $100\% / 100\%$, mean translation error monotonically decreases from $2.51\,\text{m}$ to $1.51\,\text{m}$ and mean rotation error drops from $4.82^\circ$ to $2.11^\circ$. Crucially, because DINOv2 inference is executed only once per query and multi-angle search operates on the compact $32 \times 32$ feature lattice via $O(1)$ SAT lookups—without retaining raw point clouds for \texttt{dMM}-style ICP/RANSAC refinement—end-to-end online latency remains invariant at $167.0\text{--}178.8\,\text{ms}$ ($\le 200\,\text{ms}$ real-time budget) across all coverage regimes.

\subsection{Per-Sequence Breakdown Across Varied Map Scales}
Table~\ref{tab:per_sequence} details the per-sequence performance across all 12 RADISH environments under partial ($Q=50\%, \text{DB}=100\%$) and full ($Q=100\%, \text{DB}=100\%$) observations. On standard $80\text{--}120\,\text{m}$ indoor facilities, our pipeline consistently achieves sub-meter translation accuracy (e.g., $0.24\,\text{m}$ on \texttt{edmonton} and \texttt{fr079}, $0.27\,\text{m}$ on \texttt{orebro}, $0.29\,\text{m}$ on \texttt{fr101}, and $0.38\,\text{m}$ on \texttt{belgioioso}). Even on wide-span $400\,\text{m}$-class campus and corridor benchmarks (\texttt{MIT\_Infinite\_Corridor} at $395.4\,\text{m}$ and \texttt{fr-campus} at $404.7\,\text{m}$, where $1\,\text{px} \approx 1.56\,\text{m}$), IRLS-weighted 2D-SVD refinement constrains full-map translation error to $3.50\,\text{m}$ ($2.2\,\text{px}$) and $3.92\,\text{m}$ ($2.5\,\text{px}$) with $1.79^\circ$ and $2.14^\circ$ rotation error.

\subsection{Implementation and Hyperparameter Summary}
Table~\ref{tab:hyperparams} summarizes the complete set of rasterization, network architecture, homogeneous SAT, and 3-DoF geometric alignment hyperparameters used across all experiments in this appendix.

%% file: ieee.bbl
\begin{thebibliography}{10}

\bibitem{1}
Brian Yamauchi and Randall Beer.
\newblock Spatial learning for navigation in dynamic environments.
\newblock {\em IEEE Trans. Systems, Man, and Cybernetics, Part B},
  26(3):496--505, 1996.

\bibitem{2}
Shoudong Huang, Zhan Wang, and Gamini Dissanayake.
\newblock Sparse local submap joining filter for building large-scale maps.
\newblock {\em IEEE Trans. Robotics (TRO)}, 24(5):1121--1130, 2008.

\bibitem{3}
Andreas Wendel, Michael Maurer, Gottfried Graber, Thomas Pock, and Horst
  Bischof.
\newblock Dense reconstruction on-the-fly.
\newblock In {\em IEEE Int. Conf. Computer Vision and Pattern Recognition
  (CVPR)}, pages 1450--1457, 2012.

\bibitem{4}
Stephen Se, David~G. Lowe, and James~J. Little.
\newblock Mobile robot localization and mapping with uncertainty using
  scale-invariant visual landmarks.
\newblock {\em I. J. Robotic Res.}, 21(8):735--760, 2002.

\bibitem{5}
Mark Cummins and Paul Newman.
\newblock Highly scalable appearance-only slam - fab-map 2.0.
\newblock In {\em Robotics: Science and Systems}, 2009.

\bibitem{16}
Davide Scaramuzza, Friedrich Fraundorfer, and Marc Pollefeys.
\newblock Closing the loop in appearance-guided omnidirectional visual odometry
  by using vocabulary trees.
\newblock {\em Robot. Auton. Syst.}, 58(6):820--827, 2010.

\bibitem{11}
M.A. Fischler and R.A. Elschlager.
\newblock The representation and matching of pictorial structures.
\newblock {\em IEEE Trans. Computers}, C-22(1):67 -- 92, 1973.

\bibitem{12}
Bastian Leibe, Ales Leonardis, and Bernt Schiele.
\newblock Combined object categorization and segmentation with an implicit
  shape model.
\newblock In {\em Euro. Conf. Compuer Vision (ECCV) workshop on statistical
  learning in computer vision}, pages 17--32, 2004.

\bibitem{10}
Pablo Arbelaez, Bharath Hariharan, Chunhui Gu, Saurabh Gupta, Lubomir~D.
  Bourdev, and Jitendra Malik.
\newblock Semantic segmentation using regions and parts.
\newblock In {\em IEEE Int. Conf. Computer Vision and Pattern Recognition
  (CVPR)}, pages 3378--3385, 2012.

\bibitem{13}
Hung-Khoon Tan and Chong-Wah Ngo.
\newblock Common pattern discovery using earth mover s distance and local flow
  maximization.
\newblock In {\em IEEE Int. Conf. Computer Vision (ICCV)}, pages 1222--1229,
  2005.

\bibitem{14}
Yuning Jiang, Jingjing Meng, and Junsong Yuan.
\newblock Randomized visual phrases for object search.
\newblock In {\em IEEE Int. Conf. Computer Vision and Pattern Recognition
  (CVPR)}, pages 3100--3107, 2012.

\bibitem{21}
Yuuto Chokushi, Kanji Tanaka, and Masatoshi Ando.
\newblock Common landmark discovery in urban scenes.
\newblock {\em IAPR Int. Conf. Machine Vision Applications}, 2013.

\bibitem{15}
Andrew Howard and Nicholas Roy.
\newblock The robotics data set repository (radish), 2003.

\bibitem{19}
Pedro~F. Felzenszwalb, Ross~B. Girshick, David~A. McAllester, and Deva Ramanan.
\newblock Object detection with discriminatively trained part-based models.
\newblock {\em IEEE Trans. Pattern Analysis and Machine Intelligence (PAMI)},
  32(9):1627--1645, 2010.

\bibitem{33ism}
Megha Pandey and Svetlana Lazebnik.
\newblock Scene recognition and weakly supervised object localization with
  deformable part-based models.
\newblock In {\em ICCV}, pages 1307--1314, 2011.

\bibitem{35ism}
S.N. Parizi, J.G. Oberlin, and P.F. Felzenszwalb.
\newblock Reconfigurable models for scene recognition.
\newblock In {\em Computer Vision and Pattern Recognition (CVPR), 2012 IEEE
  Conference on}, pages 2775--2782, 2012.

\bibitem{20}
Kenichi Saeki, Kanji Tanaka, and Takeshi Ueda.
\newblock Lsh-ransac: An incremental scheme for scalable localization.
\newblock In {\em IEEE Int. Conf. Robotics and Automation (ICRA)}, pages
  3523--3530, 2009.
\newblock \\\url{http://rc.his.u-fukui.ac.jp/LR.pdf}.

\bibitem{22}
Tomomi Nagasaka and Kanji Tanaka.
\newblock An incremental scheme for dictionary-based compressive slam.
\newblock In {\em IEEE/RSJ Int. Conf. Intelligent Robots and Systems (IROS)},
  pages 872--879, 2011.
\newblock \\\url{http://rc.his.u-fukui.ac.jp/ICDCS.pdf}.

\bibitem{23}
Kanji Tanaka and Tomomi Nagasaka.
\newblock Dictionary-based compressive slam.
\newblock {\em SICE JCMSI Journal of SICE}, 6(1):54--64, 2013.
\newblock \\\url{http://rc.his.u-fukui.ac.jp/DCS.pdf}.

\bibitem{33}
Tinne Tuytelaars, Christoph~H. Lampert, Matthew~B. Blaschko, and Wray~L.
  Buntine.
\newblock Unsupervised object discovery: A comparison.
\newblock {\em International Journal of Computer Vision}, 88(2):284--302, 2010.

\bibitem{17}
Josef Sivic and Andrew Zisserman.
\newblock Video google: A text retrieval approach to object matching in videos.
\newblock In {\em IEEE Int. Conf. Computer Vision (ICCV)}, pages 1470--1477,
  2003.

\bibitem{18}
Sven Olufs and Markus Vincze.
\newblock Robust single view room structure segmentation in manhattan-like
  environments from stereo vision.
\newblock In {\em IEEE Int. Conf. Robotics and Automation (ICRA)}, pages
  5315--5322, 2011.

\bibitem{32}
Masatoshi Ando, Kanji Tanaka, and Yousuke Inagaki.
\newblock A bag-of-bounding-boxes approach to object-level view image
  retrieval.
\newblock In {\em Proc. SICE Annual Conference}, 2013.
\newblock \\\url{http://rc.his.u-fukui.ac.jp/BOBB.pdf}.

\bibitem{6}
Kai Ni, Hailin Jin, and Frank Dellaert.
\newblock Groupsac: Efficient consensus in the presence of groupings.
\newblock In {\em IEEE Int. Conf. Computer Vision (ICCV)}, pages 2193--2200,
  2009.

\bibitem{24}
S.~Se, D.G. Lowe, and J.J. Little.
\newblock Vision-based global localization and mapping for mobile robots.
\newblock {\em Robotics, IEEE Transactions on}, 21(3):364--375, 2005.

\bibitem{25}
J.~Neira, J.D. Tardos, and J.A. Castellanos.
\newblock Linear time vehicle relocation in slam.
\newblock In {\em Proc. IEEE Int. Conf. Robotics and Automation}, volume~1,
  pages 427--433, 2003.

\end{thebibliography}
